\documentclass[journal]{IEEEtran}

\ifCLASSINFOpdf

\else

\fi

\usepackage{algorithm}
\usepackage{algorithmic}
\usepackage{array}
\usepackage{tabularx}
\usepackage{multirow}
\usepackage{makecell}
\usepackage{diagbox}
\usepackage{booktabs}
\usepackage{amssymb}
\usepackage{bbding}
\usepackage{threeparttable}

\usepackage{graphicx}
\usepackage{amsmath} 
\usepackage{stfloats}
\usepackage{xcolor}

\begin{document}

\title{FBSDiff: Plug-and-Play Frequency Band Substitution of Diffusion Features for Highly Controllable Text-Driven Image Translation}

\author{Xiang Gao,
        Jiaying Liu*,~\IEEEmembership{Senior Member,~IEEE}
\thanks{\textbf{Accepted by ACM MM 2024}. Xiang Gao is with Wangxuan Institute of Computer Technology, Peking University, Beijing, 100871 China, e-mail: (gaoxiang1102@pku.edu.cn). Jiaying Liu is with Wangxuan Institute of Computer Technology, Peking University, Beijing, 100871 China, e-mail: (liujiaying@pku.edu.cn).
}}

\markboth{}{}

\maketitle

\begin{abstract}
Large-scale text-to-image diffusion models have been a revolutionary milestone in the evolution of generative AI and multimodal technology, allowing wonderful image generation with natural-language text prompt. However, the issue of lacking controllability of such models restricts their practical applicability for real-life content creation. Thus, attention has been focused on leveraging a reference image to control text-to-image synthesis, which is also regarded as manipulating (or editing) a reference image as per a text prompt, namely, text-driven image-to-image translation. This paper contributes a novel, concise, and efficient approach that adapts pre-trained large-scale text-to-image (T2I) diffusion model to the image-to-image (I2I) paradigm in a plug-and-play manner, realizing high-quality and versatile text-driven I2I translation without any model training, model fine-tuning, or online optimization process. To guide T2I generation with a reference image, we propose to decompose diverse guiding factors with different frequency bands of diffusion features in the DCT spectral space, and accordingly devise a novel frequency band substitution layer which realizes dynamic control of the reference image to the T2I generation result in a plug-and-play manner. We demonstrate that our method allows flexible control over both guiding factor and guiding intensity of the reference image simply by tuning the type and bandwidth of the substituted frequency band, respectively. Extensive qualitative and quantitative experiments verify superiority of our approach over related methods in I2I translation visual quality, versatility, and controllability. The code is publicly available at: https://github.com/XiangGao1102/FBSDiff.
\end{abstract}


\IEEEpeerreviewmaketitle

\begin{figure}[t]
    \centering
    \includegraphics[width=3.5in]{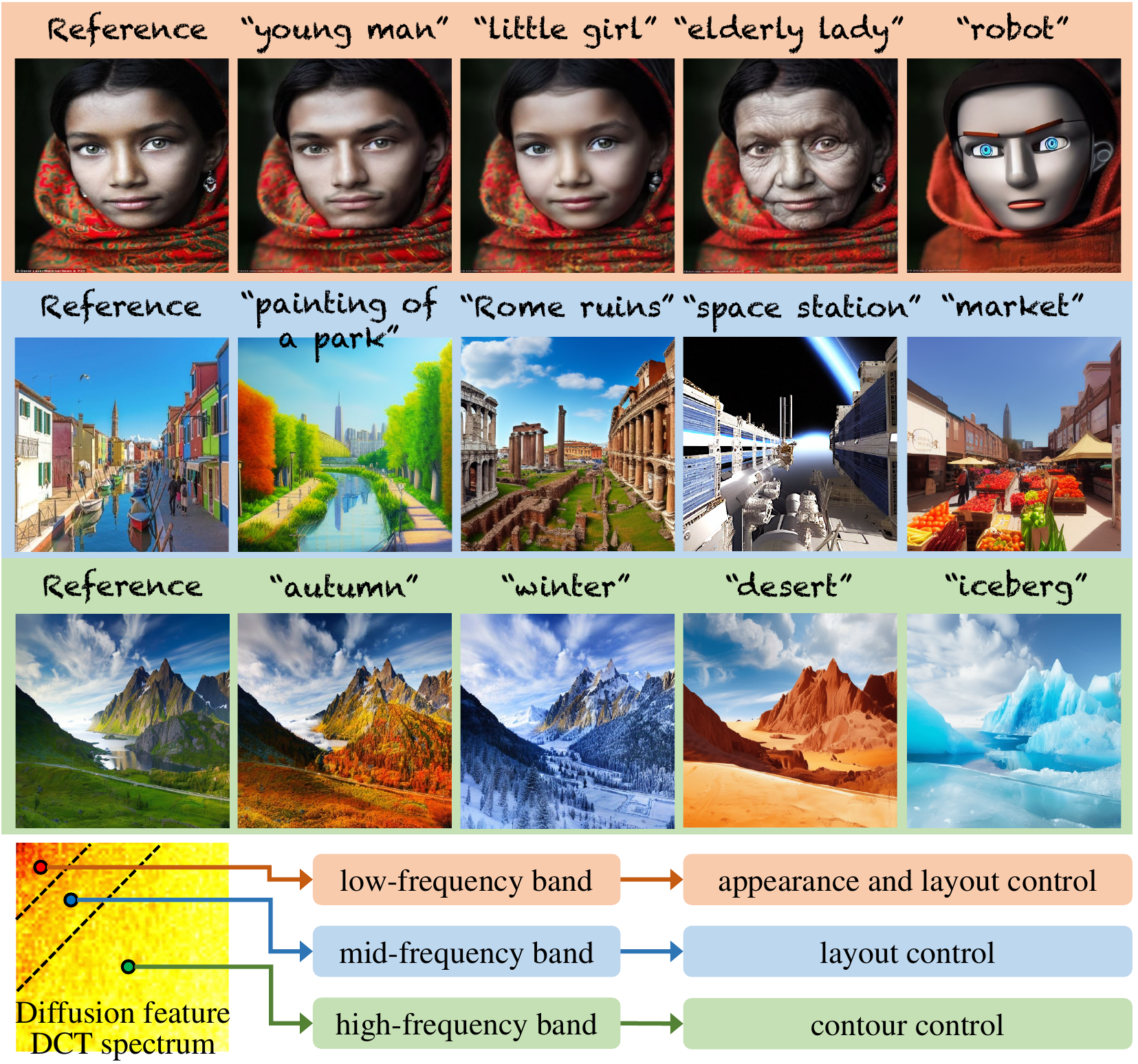}
    \caption{Based on the pre-trained text-to-image diffusion model, FBSDiff enables efficient text-driven image-to-image translation by proposing a plug-and-play reference image guidance mechanism. It allows flexible control over different guiding factors (e.g., image appearance, image layout, image contours) of the reference image to the T2I generated image, simply by dynamically substituting different types of DCT frequency bands during the reverse sampling process of the diffusion model. Better viewed with zoom-in.}
    \label{fig:teaser}
\end{figure}

\section{Introduction}
As a typical application of the booming multimodal technology, text-driven I2I translation is an appealing computer vision problem that aims to translate a reference image as per a text prompt. It extends text-to-image (T2I) synthesis to more controllability by controlling T2I generation result with a reference image. Since the advent of CLIP \cite{radford2021learning} bridging vision and language through large-scale contrastive pre-training, attempts have been made to instruct image manipulation with text by combining CLIP with generative models. VQGAN-CLIP \cite{crowson2022vqgan} pioneers text-driven image translation by optimizing VQGAN \cite{esser2021taming} latent image embedding with CLIP text-image similarity loss. DiffusionCLIP \cite{kim2022diffusionclip} fine-tunes diffusion model \cite{ho2020denoising} under the same CLIP loss to manipulate an image as per a text. DiffuseIT \cite{kwon2022diffusion} combines VIT-based structure loss \cite{tumanyan2022splicing} and CLIP-based semantic loss to guide diffusion model's reverse sampling process via manifold constrained gradient \cite{chung2022improving}, synthesizing translated image that complies with the target text while maintaining the structure of the reference image. However, these methods are not competitive in generation visual quality due to the limited model capacity of backbone generative model as well as the inherent unstability caused by online fine-tuning or optimization process. 

To promote image translation visual quality, efforts have been made to train large models on massive data. InstructPix2Pix \cite{brooks2023instructpix2pix} employs GPT-3 \cite{brown2020language} and Stable Diffusion \cite{rombach2022high} to synthesize huge amounts of paired training data, based on which trains a supervised text-driven I2I mapping for general image manipulation task. Design Booster \cite{sun2023design} trains a latent diffusion model \cite{rombach2022high} conditioned on both text embedding and image embedding, realizing layout-preserved text-driven I2I translation. Nevertheless, these methods are computationally intensive in training large models from scratch and less efficient in collecting immense training data.

To circumvent formidable training costs, research has been focused on leveraging off-the-shelf large-scale T2I diffusion models for text-driven I2I translation. This type of methods further divide into fine-tuning-based methods and inversion-based methods. 

The former type of fine-tuning-based methods represented by SINE \cite{zhang2023sine} and Imagic \cite{kawar2023imagic} fine-tune the pre-trained T2I diffusion model to reconstruct an input reference image before manipulating it with a target text. These methods require separate fine-tuning of the entire diffusion model for each time of image manipulation, which is less efficient and prone to underfitting or overfitting to the reference image.

The latter type of inversion-based methods invert reference image into diffusion model's Gaussian noise space and then generate the translated image via the reverse sampling process guided by the target text. A pivotal challenge of this pipeline is that the sampling trajectory may severely deviate from the inversion trajectory due to the error accumulation caused by the classifier-free guidance technique \cite{ho2022classifier}, which severely impairs the correlation between the reference image and the translated image. To remedy this issue, Null-text Inversion \cite{mokady2023null} optimizes the unconditional null-text embedding to calibrate the sampling trajectory step by step. Prompt Tuning Inversion \cite{dong2023prompt} proposes to minimize trajectory divergence with an optimization to encode the reference image into a learnable prompt embedding. Similarly, StyleDiffusion \cite{li2023stylediffusion} opts to optimize the ``value" embedding of the cross-attention layer as the visual encoding of the reference image. Pix2Pix-zero \cite{parmar2023zero} penalizes trajectory deviation by matching cross-attention maps between the two trajectories with least-square loss. These methods apply per-step online optimization to calibrate the whole sampling trajectory, introducing additional computational cost and time overhead. Moreover, most of these methods adopt the cross-attention control technique introduced in Prompt-to-Prompt \cite{hertz2022prompt} for image structure preservation. This makes them rely on a paired source text of the reference image, which is not flexible or even available in most cases. Plug-and-Play (PAP) \cite{tumanyan2023plug} proposes to leverage feature maps and self-attention maps extracted from internal layers of the denoising U-Net to maintain image structure, realizing optimization-free text-driven I2I translation. However, the algorithm is sensitive to specific layer selection, and the feature extraction process is also time-consuming.

In this paper, we propose a concise and efficient approach termed FBSDiff, realizing plug-and-play and highly controllable text-driven I2I translation from a frequency-domain perspective. To guide T2I generation with a reference image, a key missing ingredient of existing methods is the mechanism to control the guiding factor (e.g., image appearance, layout, contours) and guiding intensity of the reference image. Since different image guiding factors are difficult to isolate in the spatial domain, we consider decomposing them in the frequency domain by modeling them with different frequency bands of diffusion features in the Discrete Cosine Transform (DCT) spectral space. Based on this motivation, we propose an inversion-based text-driven I2I translation framework featured with a novel frequency band substitution mechanism, which efficiently enables reference image guidance of the T2I generation by dynamically substituting a certain DCT frequency band of diffusion features with the corresponding counterpart of the reference image along the reverse sampling process. As displayed in Fig. \ref{fig:teaser}, T2I generation with appearance and layout control, pure layout control, and contour control of the reference image can be respectively realized by transplanting low-frequency band, mid-frequency band, and high-frequency band between diffusion features, allowing versatile and highly controllable text-driven I2I translation.

The strengths of our method are fourfold: (\uppercase\expandafter{\romannumeral1}) plug-and-play efficiency: our method extends pre-trained T2I diffusion model to the realm of I2I in a plug-and-play manner; (\uppercase\expandafter{\romannumeral2}) conciseness: our method dispenses with the need for the paired source text of the reference image as well as cumbersome attention modulation process as compared with existing advanced methods, all while achieving leading I2I translation performance; (\uppercase\expandafter{\romannumeral3}) model generalizability: our method transplants frequency band of diffusion features along the reverse sampling trajectory, requiring no access to any internal features of the denoising network, and thus decouples with the specific diffusion model backbone architecture as compared with existing methods; (\uppercase\expandafter{\romannumeral4}) controllability: our method allows flexible control over the guiding factor and guiding intensity of the reference image simply by tuning the type and bandwidth of the substituted frequency band. The effectiveness of our method is fully demonstrated both qualitatively and quantitatively. To summarize, we make the following key contributions:
\begin{itemize}
    \item We provide new insights about controllable diffusion process from a novel frequency-domain perspective.
    \item We propose a novel frequency band substitution technique, realizing plug-and-play text-driven I2I translation without any model training, model fine-tuning, and online optimization process.
    \item We contribute a concise and efficient text-driven I2I framework that is free from source text and cumbersome attention modulations, highly controllable in both guiding factor and guiding intensity of the reference image, and invariant to the architecture of the used diffusion model backbone, all while achieving superior I2I translation performance compared with existing advanced methods.
\end{itemize}

\section{Related Work}
\subsection{Diffusion Model}
Since the advent of DDPM \cite{ho2020denoising}, diffusion model has soon dominated the family of generative models \cite{dhariwal2021diffusion}. Afterwards, much progress have been made to improve diffusion model in both methodology and application. DDIM \cite{song2020denoising} and its variants \cite{lu2022dpm,lu2022dpm} accelerate diffusion model sampling process to tens of times with only marginal drop in generation quality, promoting its practicability dramatically. Palette \cite{saharia2022palette} extends diffusion model from unconditional image generation to the realm of conditional image synthesis, opening the door of diffusion-based image-to-image translation. With the advancement of multimodal technology, large-scale T2I diffusion models \cite{nichol2022glide, ramesh2022hierarchical, saharia2022photorealistic} are proposed to generate high-resolution images with open-domain text prompts, bringing content creation to an unprecedented level. To lower computational overhead of large-scale T2I model, Latent Diffusion Model (LDM) \cite{rombach2022high} proposes to transfer diffusion model from high-dimension pixel space to low-dimensional feature space, contributing the most widely used architecture in AIGC industry. To introduce more controllability to T2I synthesis, ControlNet \cite{zhang2023adding} and T2i-adapter \cite{mou2024t2i} add spatial control to T2I diffusion models by training a control module of the denoising U-Net conditioned on certain image priors (e.g., canny edges, depth maps, human key points, etc.). SDXL \cite{podell2023sdxl} and DiTs \cite{peebles2023scalable} propose Transformer \cite{vaswani2017attention} based backbone denoising network, improving T2I diffusion model to larger capacity. Up to now, diffusion model has been applied to a wide variety of vision fields such as image super-resolution \cite{saharia2022image}, image inpainting \cite{lugmayrinpainting}, image colorization \cite{liang2024control}, semantic segmentation \cite{tan2023diffss}, point cloud generation \cite{luo2021diffusion}, video synthesis \cite{yu2023video}, 3D reconstruction \cite{anciukevivcius2023renderdiffusion}, etc, and is still making rapid progress in theory and potential applications.

\begin{figure*}[t]
    \centering
    \includegraphics[width=\textwidth]{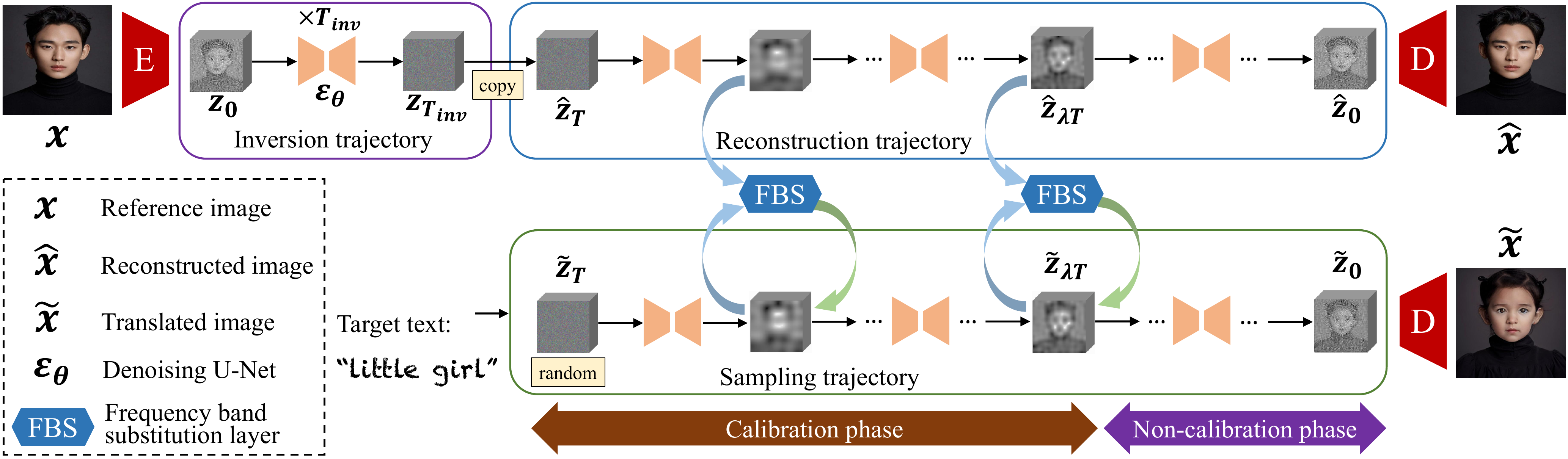}
    \caption{Overview of FBSDiff. Based on the pre-trained latent diffusion model (LDM), FBSDiff starts with an inversion trajectory that inverts reference image into the LDM Gaussian noise space, then a reconstruction trajectory is applied to reconstruct the reference image from the inverted Gaussian noise, providing intermediate denoising results as pivotal guidance features. The guidance features are leveraged to guide the text-driven sampling trajectory of the LDM to exert reference image control, which is realized by dynamically transplanting certain DCT frequency bands from diffusion features along the reconstruction trajectory into the corresponding features along the sampling trajectory. The dynamic DCT frequency band transplantation is implemented in a plug-and-play manner with our proposed frequency band substitution layer (FBS layer).}
    \label{fig:method_overview}
\end{figure*}

\begin{figure}[t]
    \centering
    \includegraphics[width=3.5in]{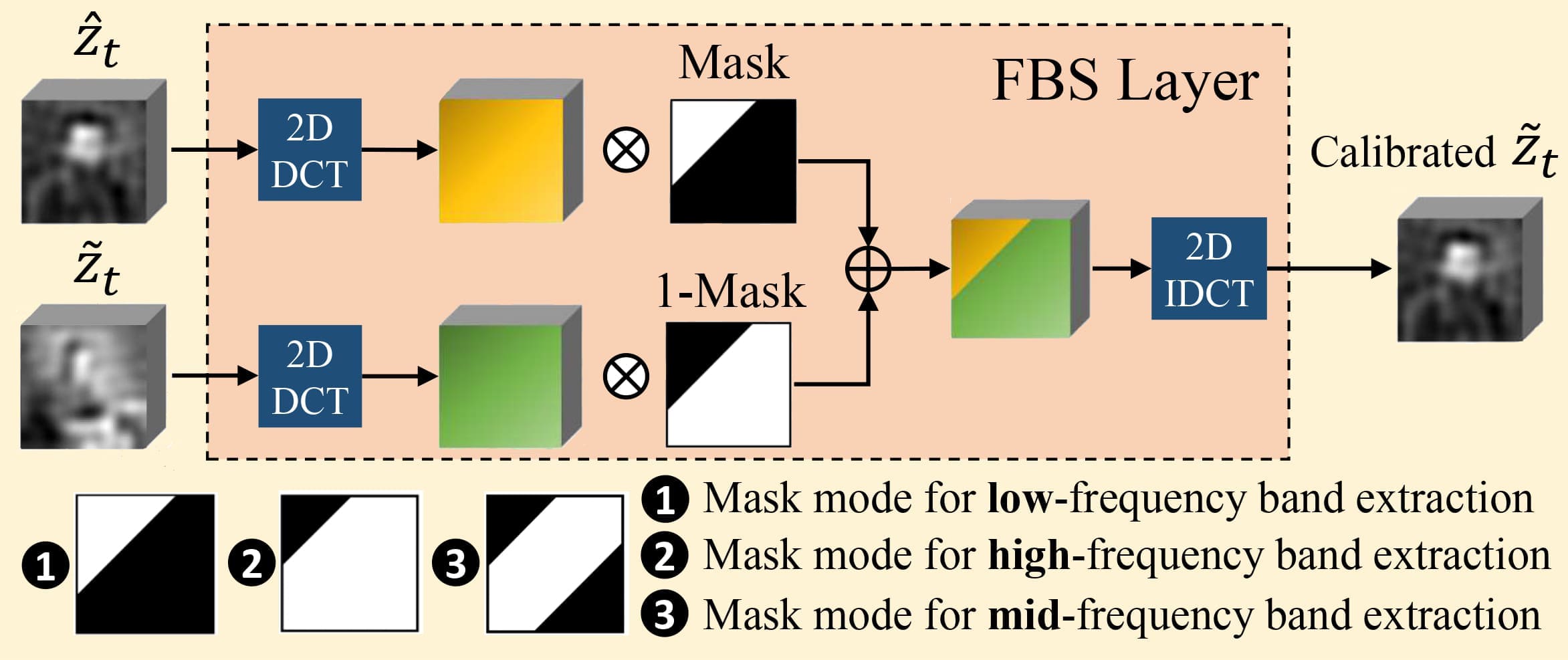}
    \caption{Illustration of the proposed frequency band substitution (FBS) layer. The FBS layer takes in two diffusion features and substitutes a certain frequency band of one feature with the corresponding frequency band of the other feature. This is realized by converting the two diffusion features into the frequency domain via 2D DCT, extracting and transplanting a certain DCT frequency band, and converting the fused DCT features back to spatial domain via 2D IDCT. The frequency band extraction and transplantation are implemented with binary masking.}
    \label{fig:FBS}
\end{figure}

\subsection{Computer Vision in Frequency Perspective}
Deep neural network models are mostly applied to tackle vision tasks in spatial or temporal domain, some research reveals that model performance can also be boosted from frequency domain. For example, Ghosh et al. \cite{ghosh2016deep} introduce DCT to convolutional neural network  for image classification, accelerating network convergence speed. Xie et al. \cite{xie2021learning} propose a frequency-aware dynamic network for lightweight image super-resolution. Cai et al. \cite{cai2021frequency} impose Fourier frequency spectrum consistency to image translation tasks, achieving better identity preservation ability. FreeU \cite{si2023freeu} improves T2I generation quality by selectively enhancing or depressing different frequency components of diffusion features inside the denoising U-Net model. ILVR \cite{choi2021ilvr} proposes to fuse low-frequency information of the forward diffusion process into the reverse sampling process for conditioned image synthesis. Our method differs with ILVR in that ILVR simulates low-pass filtering with simple feature downsampling and upsampling and performs information fusion in the spatial domain, while our method explicitly extracts and transplants frequency bands of diffusion features in the DCT domain. FCDiffusion \cite{gao2024frequency} shares similar spirit of frequency-based control of T2I diffusion model with our method. However, FCDiffusion relies on training multiple frequency control branches to realize versatile control effects, while our method achieves versatility and high controllability in both guiding factor and guiding intensity of the reference image in a training-free and plug-and-play manner.

\section{Method}
In this section, we first describe the overall model architecture of our FBSDiff, then elaborate on our proposed frequency band substitution mechanism, and finally summarize our algorithm and describe implementation details. For the diffusion model background, please refer to the Appendix.

\subsection{Overall Architecture}
Established on the pre-trained Latent Diffusion Model (LDM), FBSDiff adapts it from T2I generation to the realm of text-driven I2I translation with our proposed plug-and-play reference image guidance mechanism: dynamic frequency band substitution, which efficiently realizes flexible control over both guiding factor and guiding intensity of the reference image to the T2I generated image. 

As Fig. \ref{fig:method_overview} shows, FBSDiff comprises three diffusion trajectories: (\romannumeral1) inversion trajectory ($z_{0}\rightarrow z_{T_{inv}}$); (\romannumeral2) reconstruction trajectory ($z_{T_{inv}}=\hat{z}_{T}\rightarrow \hat{z}_{0}\approx z_{0}$); (\romannumeral3) sampling trajectory (${\tilde{z}}_{T}\rightarrow {\tilde{z}}_{0}$). Starting from the initial feature $z_{0}=E(x)$ extracted from the reference image $x$ by the LDM encoder $E$, a $T_{inv}$-step DDIM inversion is employed to project $z_{0}$ into the Gaussian noise latent space conditioned on the null-text embedding $v_{\emptyset}$, based on the assumption that the ODE process can be reversed in the limit of small steps:
\begin{equation}
    z_{t+1}=\sqrt{\bar{\alpha}_{t+1}}f_{\theta}(z_{t}, t, v_{\emptyset})+\sqrt{1-\bar{\alpha}_{t+1}}\epsilon_{\theta}(z_{t}, t, v_{\emptyset}),
    \label{eq:dim_inversion}
\end{equation}
\begin{equation}
    f_{\theta}(z_{t}, t, v_{\emptyset})=\frac{z_{t}-\sqrt{1-\bar{\alpha}_{t}}\epsilon_{\theta}(z_{t}, t, v_{\emptyset})}{\sqrt{\bar{\alpha}_{t}}},
    \label{eq:back_to_z0}
\end{equation}
where \{$\bar{\alpha}_{t}$\} are schedule parameters that follows the same setting as DDPM \cite{ho2020denoising}, $\epsilon_{\theta}$ is the denoising U-Net of the pre-trained LDM. The Gaussian noise $z_{T_{inv}}$ obtained after the $T_{inv}$-step DDIM inversion is directly used as the initial noise feature of the subsequent reconstruction trajectory, which is a $T$-step DDIM sampling process that reconstructs $\hat{z}_{0}\approx z_{0}$ from the inverted noise feature $\hat{z}_{T}=z_{T_{inv}}$:
\begin{equation}
    {\hat{z}}_{t-1}=\sqrt{{\bar{\alpha}}_{t-1}}f_{\theta}({\hat{z}}_{t}, t, v_{\emptyset}) + \sqrt{1-{\bar{\alpha}}_{t-1}}\epsilon_{\theta}({\hat{z}}_{t}, t, v_{\emptyset}),
    \label{eq:recon}
\end{equation}
in which $f_{\theta}({\hat{z}}_{t}, t, v_{\emptyset})$ follows the same form as Eq. \ref{eq:back_to_z0}. The length of the reconstruction trajectory could be much smaller than that of the inversion trajectory (i.e., $T \ll T_{inv}$) to save inference time. The reconstruction trajectory is conditioned on the same null-text embedding $v_{\emptyset}$ as the inversion trajectory to ensure feature reconstructability (i.e., $\hat{z}_{0}\approx z_{0}$). 

Meanwhile, an equal-length sampling trajectory is applied in parallel with the reconstruction trajectory for T2I synthesis. The sampling trajectory is also a $T$-step DDIM sampling process that progressively denoises a randomly initialized Gaussian noise feature ${\tilde{z}}_{T} \sim \mathcal{N}(0, I)$ into ${\tilde{z}}_{0}$ conditioned on the text embedding $v$ of the target text prompt. To amplify the effect of text guidance, we employ classifier-free guidance technique \cite{ho2022classifier} by interpolating conditional (target text) and unconditional (null text) noise prediction at each time step with a guidance scale $\omega$ along the sampling process:
\begin{equation}
    {\tilde{z}}_{t-1}=\sqrt{{\bar{\alpha}}_{t-1}}f_{\theta}({\tilde{z}}_{t}, t, v, v_{\emptyset}) + \sqrt{1-{\bar{\alpha}}_{t-1}}\epsilon_{\theta}({\tilde{z}}_{t}, t, v, v_{\emptyset}),
    \label{eq:sampling}
\end{equation}
\begin{equation}
    f_{\theta}({\tilde{z}}_{t}, t, v, v_{\emptyset})=\frac{{\tilde{z}}_{t}-\sqrt{1-{\bar{\alpha}}_{t}}\epsilon_{\theta}({\tilde{z}}_{t}, t, v, v_{\emptyset})}{\sqrt{{\bar{\alpha}}_{t}}},
\end{equation}
\begin{equation}
    \epsilon_{\theta}({\tilde{z}}_{t}, t, v, v_{\emptyset})=\omega\cdot \epsilon_{\theta}({\tilde{z}}_{t}, t, v)+(1-\omega)\cdot\epsilon_{\theta}({\tilde{z}}_{t}, t, v_{\emptyset}).
\end{equation}

Due to the inherent property of DDIM inversion and DDIM sampling, the reconstruction trajectory forms a deterministic denoising mapping towards the reference image, during which the intermediate denoising results $\{{\hat{z}}_{t}\}$ can function as pivotal guidance features to calibrate the corresponding counterparts $\{{\tilde{z}}_{t}\}$ along the sampling trajectory. Thus, correlation between the reference image and the generated image can be established to allow for text-driven I2I translation. Specifically, we implement feature calibration by inserting a plug-and-play frequency band substitution (FBS) layer in between the reconstruction trajectory and the sampling trajectory. FBS layer substitutes a certain frequency band of $\tilde{z}_{t}$ in the sampling trajectory with the corresponding frequency band of $\hat{z}_{t}$ in the reconstruction trajectory along the reverse sampling process. The frequency band substitution effectively and efficiently imposes guidance of the reference image to the T2I synthesis process. Both the guiding factor (e.g., image appearance, image layout, image contours) and guiding intensity of the reference image can be flexibly controlled simply by tuning the type and bandwidth of the substituted frequency band, respectively. 

To improve I2I translation visual quality, we partition the sampling process into a calibration phase and a non-calibration phase, separated by the time step $\lambda T$. In the former calibration phase (${\tilde{z}}_{T} \rightarrow {\tilde{z}}_{\lambda T}$), dynamic frequency band substitution is applied at each time step for smooth calibration of the sampling trajectory; in the latter non-calibration phase (${\tilde{z}}_{\lambda T-1} \rightarrow {\tilde{z}}_{0}$), we remove FBS layer to avoid over-constrained sampling result, fully unleashing the generative power of the pre-trained T2I model to improve image generation quality. Here $\lambda$ denotes the ratio of the length of the non-calibration phase to that of the entire sampling trajectory. 

At last, the final result ${\tilde{z}}_{0}$ of the sampling trajectory is decoded back to the image space via the LDM decoder $D$, producing the final translated image $\tilde{x}$, i.e., $\tilde{x}=D({\tilde{z}}_{0})$. 

\subsection{Frequency Band Substitution Layer}
As Fig. \ref{fig:FBS} illustrates, the FBS layer takes in a pair of diffusion features ${\hat{z}}_{t}$ and ${\tilde{z}}_{t}$, converts them from the spatial domain into the frequency domain via 2D-DCT, then transplants a certain frequency band in the DCT spectrum of ${\hat{z}}_{t}$ to the same position in the DCT spectrum of ${\tilde{z}}_{t}$. Finally, 2D-IDCT is applied to transform the manipulated DCT spectrum of $\tilde{z}_{t}$ back into the spatial domain as the final calibrated feature.

In 2D DCT spectrum, elements with smaller coordinates (nearer to the top-left origin) encode lower-frequency information while larger-coordinate elements (nearer to the bottom-right corner) correspond to higher-frequency components. Most of the DCT spectral energy is occupied by a small proportion of low-frequency elements near the top-left origin. 

In FBS layer, the sum of 2D coordinates is used as the threshold to extract DCT frequency bands of different types and bandwidths through binary masking. Specifically, we design three types of binary masks which are respectively termed the low-pass mask ($Mask_{lp}$), high-pass mask ($Mask_{hp}$), and mid-pass mask ($Mask_{mp}$):
$$\left\{ 
        \begin{array}{lr}
             Mask_{lp}(x,y)=1\ \ if\ \ x+y \leq th_{lp}\ \ else\ \ 0, &  \\
             Mask_{hp}(x,y)=1\ \ if\ \ x+y>th_{hp}\ \ else\ \ 0, &  \\
             Mask_{mp}(x,y)=1\ \ if\ \ th_{mp1}<x+y \leq th_{mp2}\ \ else\ \ 0,
        \end{array}
\right.
$$
where $th_{lp}$ is the threshold of the low-pass filtering; $th_{hp}$ is the threshold of the high-pass filtering; $th_{mp1}$ and $th_{mp2}$ are respectively the lower bound and upper bound of the mid-pass filtering. Given a binary mask $Mask_{*} \in \{Mask_{lp}, Mask_{hp}, Mask_{mp}\}$, the frequency band substitution operation in the FBS layer can be formulated as:
\begin{equation}
    \begin{aligned}
    {\tilde{z}}_{t}=&IDCT(DCT({\hat{z}}_{t})\cdot Mask_{*} + \\
    &DCT({\tilde{z}}_{t})\cdot (1-Mask_{*})),
    \label{eq:fbs}
\end{aligned}
\end{equation}
where $DCT$ and $IDCT$ refer to the 2D-DCT and 2D-IDCT transformations respectively, which are introduced in detail in the Appendix section. The use of the low-pass mask $Mask_{lp}$, high-pass mask $Mask_{hp}$, and mid-pass mask $Mask_{mp}$ respectively corresponds to the extraction and substitution of the low-frequency band, high-frequency band, and mid-frequency band, which controls different guiding factors of the reference image to the T2I generated result:

\begin{itemize}
    \item \textbf{Low-frequency band} substitution enables low-frequency information guidance of the reference image $x$, realizing image appearance (e.g., color, luminance) and layout control over the generated image $\tilde{x}$;
    \item \textbf{High-frequency band} substitution enables high-frequency information guidance of $x$, realizing image contour control over the generated image $\tilde{x}$;
    \item \textbf{Mid-frequency band} substitution enables mid-frequency information guidance of the reference image $x$. By filtering out higher-frequency contour information and lower-frequency appearance information in the DCT spectrum, the substitution of the mid-frequency band realizes only image layout control over the generated image $\tilde{x}$.
\end{itemize}

The DCT masking type and the corresponding thresholds used in the FBS layer are hyper-parameters of our method, which could be flexibly modulated to enable control over diverse guiding factors and continuous guiding intensity of the reference image $x$ to the T2I generated image $\tilde{x}$.

\subsection{Implementation Details}
\begin{algorithm}[t]
    \caption{Complete algorithm of FBSDiff}
    \label{algorithm}
    \begin{algorithmic}[1]
    \renewcommand{\algorithmicrequire}{\textbf{Input:}}
    \renewcommand{\algorithmicensure}{\textbf{Output:}}
    \REQUIRE{the reference image $x$ and the target text.}
    \ENSURE{the translated image $\tilde{x}$.}
        \STATE Extract the initial latent feature $z_{0}=E(x)$.
        \FOR{$t=0$ to $T_{inv}-1$}
            \STATE compute $z_{t+1}$ from $z_{t}$ via Eq. \ref{eq:dim_inversion};
        \ENDFOR 
        \COMMENT{DDIM inversion}
        \STATE Initialize $\hat{z}_{T}=z_{T_{inv}}$, $\tilde{z}_{T} \sim \mathcal{N}(0, I)$.
        \FOR{$t=T$ to $\lambda T+1$}
            \STATE compute $\hat{z}_{t-1}$ from $\hat{z}_{t}$ via Eq. \ref{eq:recon};
            \STATE compute $\tilde{z}_{t-1}$ from $\tilde{z}_{t}$ via Eq. \ref{eq:sampling};
            \STATE substitute a certain frequency band of $\tilde{z}_{t-1}$ with the corresponding counterpart of $\hat{z}_{t-1}$ via Eq. \ref{eq:fbs};
        \ENDFOR\COMMENT{DDIM sampling in the calibration phase}
        \FOR{$t=\lambda T$ to $1$}
            \STATE compute $\tilde{z}_{t-1}$ from $\tilde{z}_{t}$ via Eq. \ref{eq:sampling};
        \ENDFOR\COMMENT{DDIM sampling in the non-calibration phase}
        \STATE Obtain $\tilde{z}_{0}$ and the final translated image $\tilde{x}=D(\tilde{z}_{0})$.
    \end{algorithmic}
\end{algorithm}

We use the pre-trained Stable Diffusion v1.5 as the backbone diffusion model and set the classifier-free guidance scale $\omega=7.5$. We use 1000-step DDIM inversion to ensure high-quality reconstruction, i.e., $T_{inv}$=1000, and use 50-step DDIM sampling for both the reconstruction and sampling trajectory, i.e., $T$=50. Along the sampling trajectory, we allocate 55\% time steps to the calibration phase and the remaining 45\% steps for the non-calibration phase, i.e., $\lambda$=0.45. For the default DCT masking thresholds used in the FBS layer, we set $th_{lp}$=80 for low-frequency band substitution (low-FBS); $th_{hp}$=5 for high-frequency band substitution (high-FBS); $th_{mp1}$=5, $th_{mp2}$=80 for mid-frequency band substitution (mid-FBS). The complete algorithm of FBSDiff is presented in Alg. \ref{algorithm}.

\section{Experiments}
In this section, we first present and analyze the qualitative results of our method as well as qualitative comparison with related advanced methods; then we delve into the frequency band substitution mechanism in detail with ablation studies; finally, we show quantitative evaluations of our method and related approaches. 

\begin{figure*}[htbp]
    \centering
    \includegraphics[width=\textwidth]{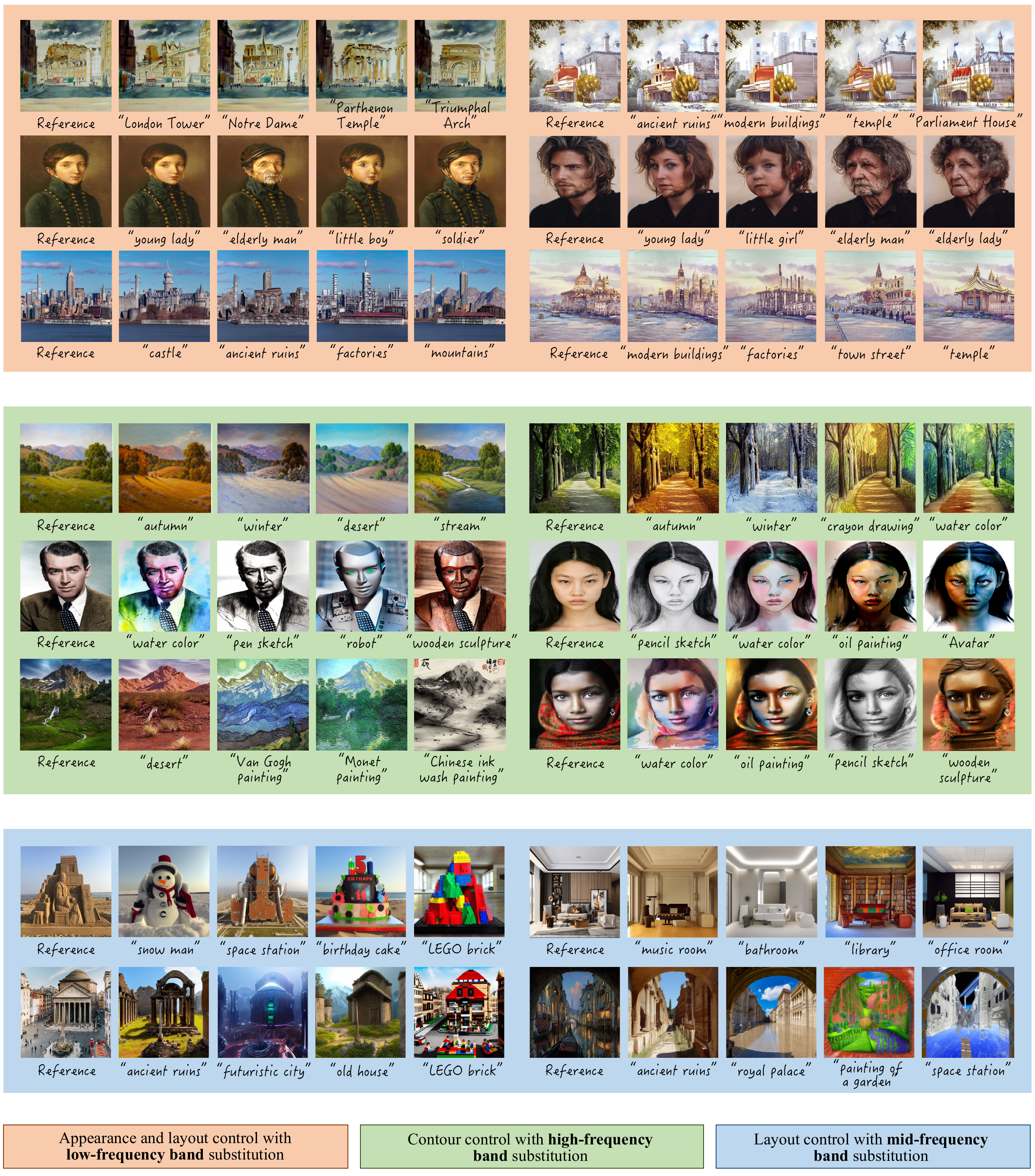}
    \caption{Qualitative results of our method with different types of frequency band substitution. For low-frequency band substitution (low-FBS), the generated image is controlled by the reference image in terms of image appearance and layout; for high-frequency band substitution (high-FBS), the reference image controls image contours of the generated image; as for mid-frequency band substitution (mid-FBS), only image layout of the generated image is controlled by the reference image. \textbf{Better viewed with zoom-in}.}
    \label{fig:qualitative_results}
\end{figure*}

\begin{figure}[htbp]
    \centering
    \includegraphics[width=3.5in]{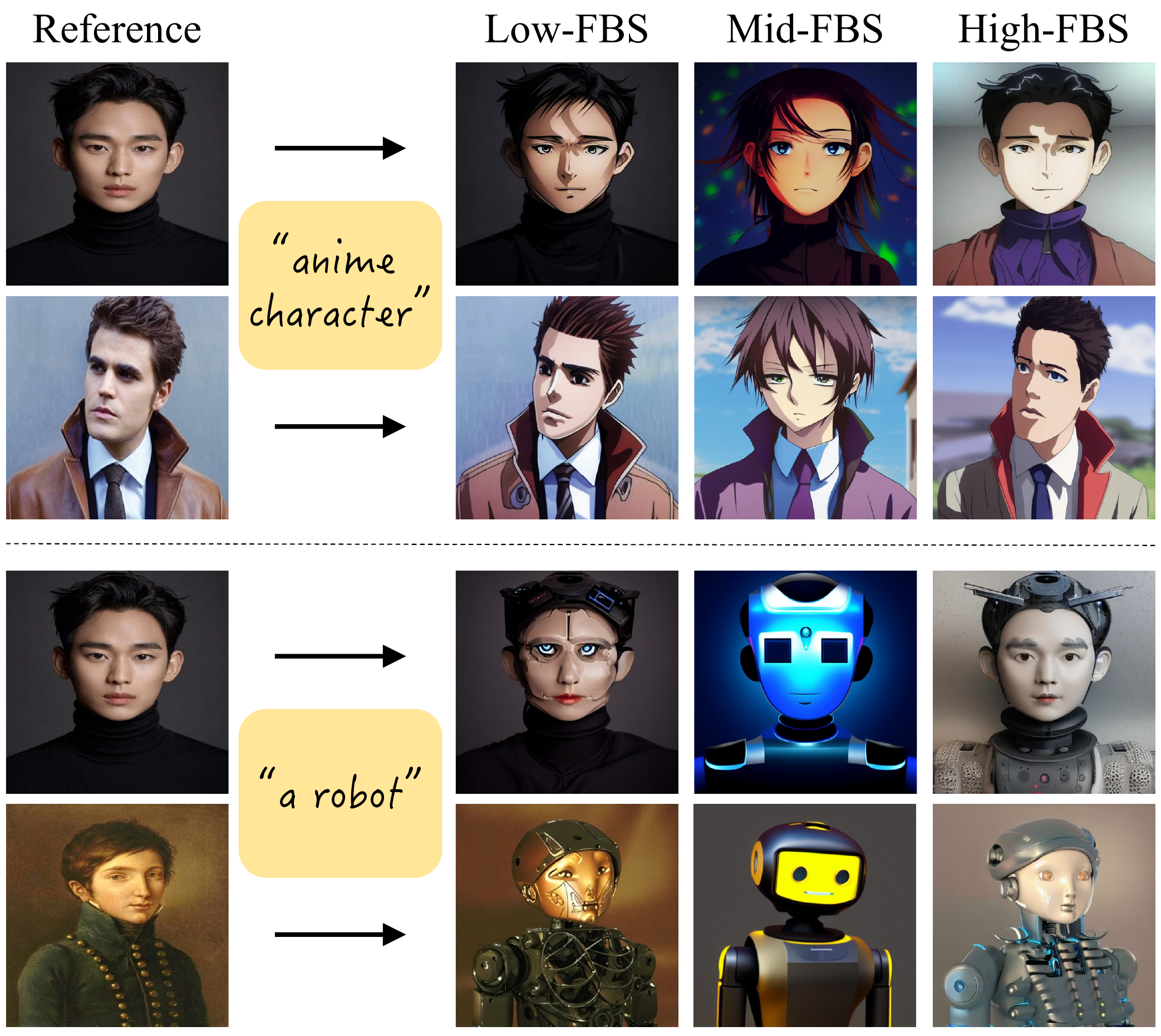}
    \caption{Comparison among different reference image control effects achieved by low-FBS, mid-FBS, and high-FBS. Low-FBS controls image appearance and layout, mid-FBS controls only image layout, and high-FBS controls image contours.}
    \label{fig:high_mid_low}
\end{figure}

\subsection{Qualitative Results}
Example text-driven I2I translation results of our method are shown in Fig. \ref{fig:qualitative_results}. Our method effectively decomposes different guiding factors of the reference image by dynamically transplanting different types of DCT frequency bands of diffusion features. The low-FBS transfers low-frequency information of the reference image into the sampling trajectory, producing translated image that inherits the original image appearance and layout. In the mode of high-FBS that dynamically transplants high-frequency components of diffusion features, the generated image is aligned with the reference image in high-frequency contours while the low-frequency appearance is not restricted. Results of mid-FBS maintain only overall image layout of the reference image, since the lower-frequency appearance information and higher-frequency contour information of the reference image are filtered out in the DCT domain. For all three modes of frequency band substitution, the image translation results exhibit high visual quality and high text fidelity for both real-world and artistic-style reference images. 

The control of our method over different guiding factors of the reference image is more clearly demonstrated in Fig. \ref{fig:high_mid_low}. The T2I generated image maintains the appearance and layout of the reference image with low-FBS; preserves detailed image contours of the reference image with high-FBS; and inherits pure image layout with mid-FBS. 

\begin{figure}[t]
    \centering
    \includegraphics[width=3.5in]{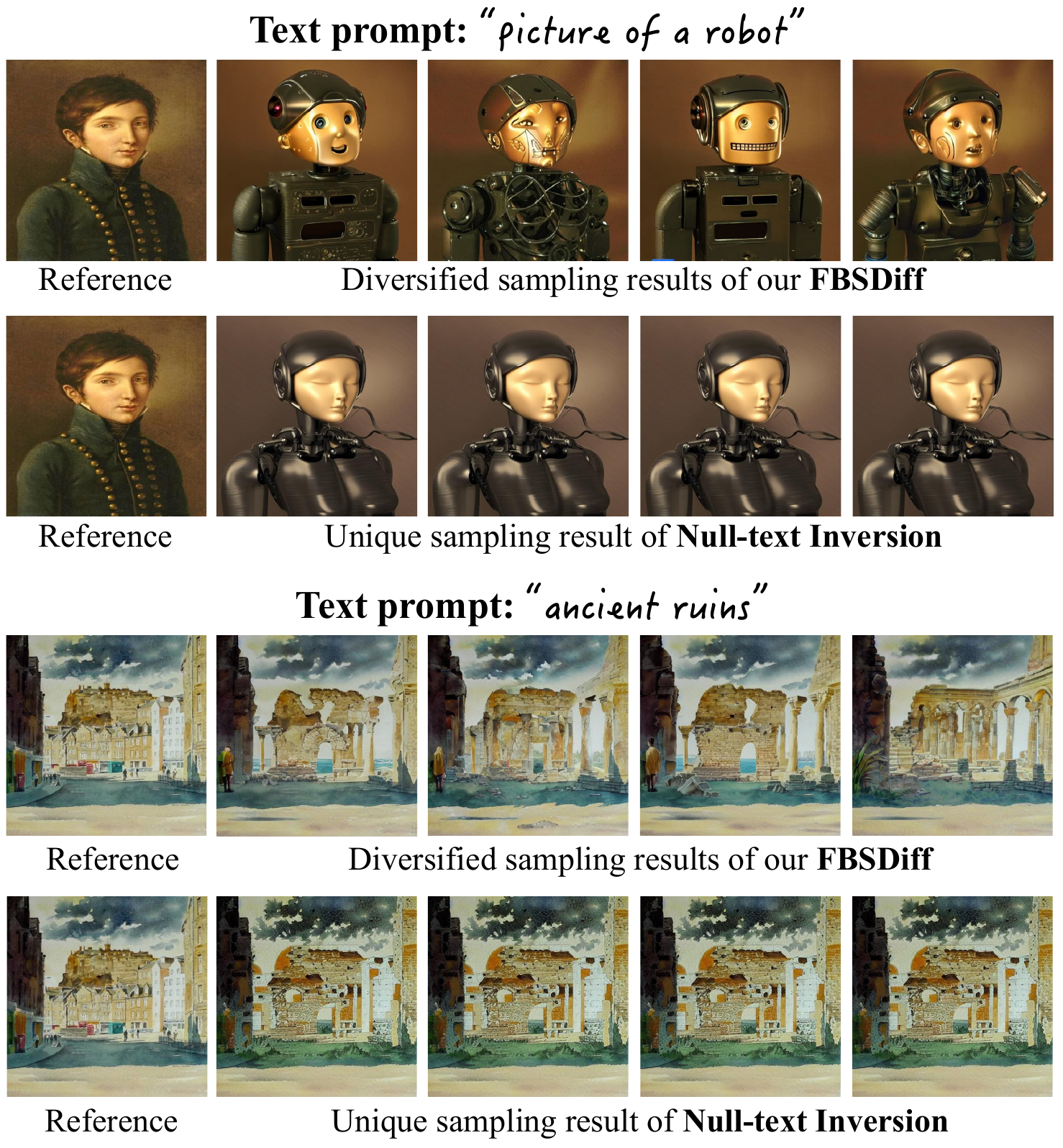}
    \caption{Our method enables diverse sampling results for fixed reference image and text prompt, as contrasted with Null-text Inversion that produces unique text-driven I2I result. Our method also produces results with better visual quality than Null-text Inversion.}
    \label{fig:diversified_sampling}
\end{figure}

\begin{figure*}[t]
    \centering
    \includegraphics[width=\textwidth]{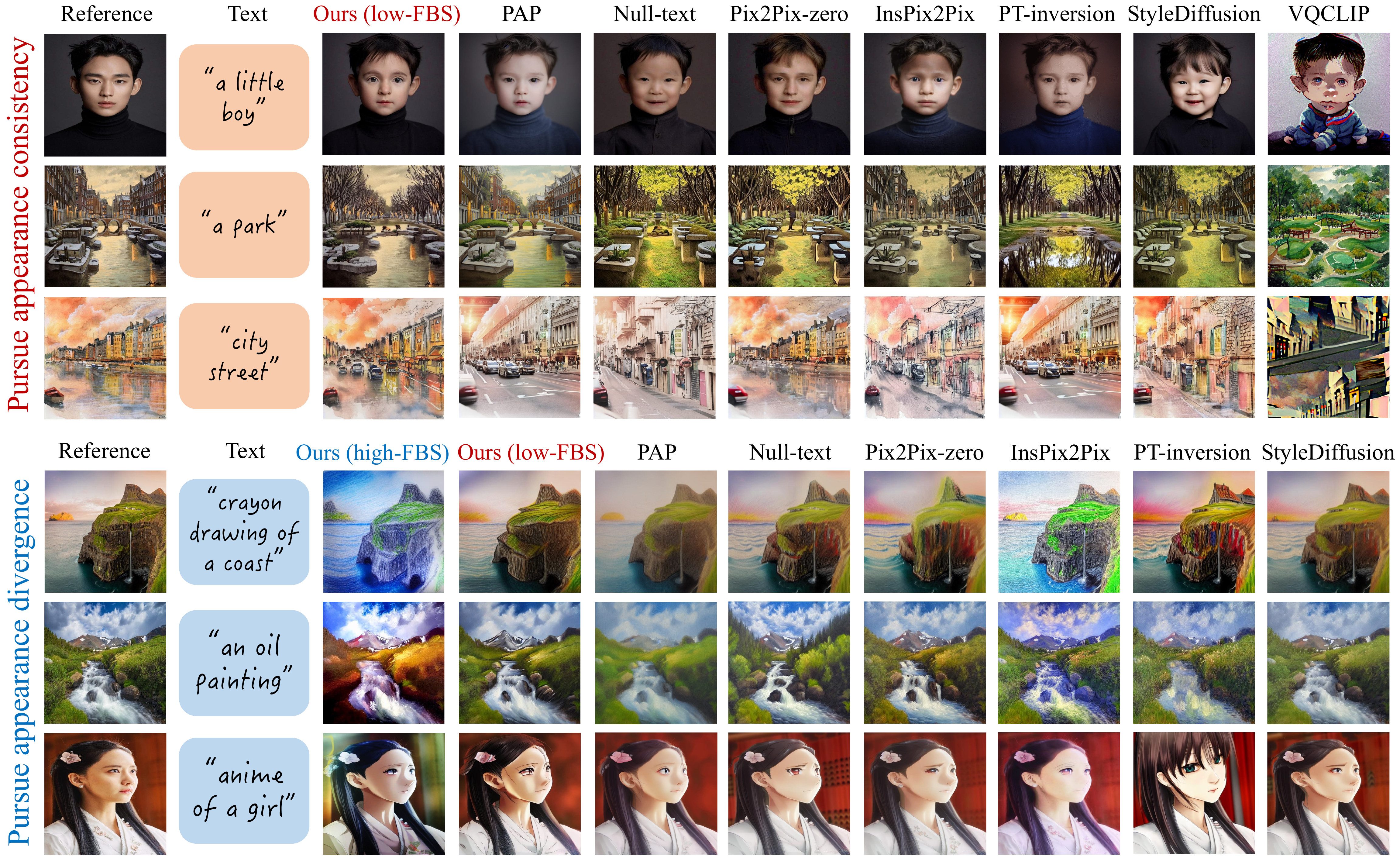}
    \caption{Qualitative method comparisons. Our FBSDiff with low-FBS is more adept at appearance preservation than related methods, which better suits to I2I task pursuing appearance consistency between the reference image and the generated image (top panel). Conversely, our method with high-FBS remarkably facilitates I2I appearance change compared with related methods, which better suits to I2I task pursuing appearance divergence (bottom panel).}
    \label{fig:method_compare}
\end{figure*}

We qualitatively compare our method with SOTA text-driven I2I translation methods including Plug-and-Play (PAP) \cite{tumanyan2023plug}, Null-text Inversion (Null-text) \cite{mokady2023null}, Pix2Pix-zero \cite{parmar2023zero}, InstructPix2Pix (InsPix2Pix) \cite{brooks2023instructpix2pix}, Prompt Tuning Inversion (PT-inversion) \cite{dong2023prompt}, StyleDiffusion \cite{li2023stylediffusion}, and VQGAN-CLIP (VQCLIP) \cite{crowson2022vqgan}, results are displayed in Fig. \ref{fig:method_compare}. The top panel of Fig. \ref{fig:method_compare} shows that our method with low-FBS achieves better appearance consistency between the reference image and the translated result than related approaches, and is thus better suited to image creation scenario which favors inheriting the appearance and style from an existing image. The bottom panel of Fig. \ref{fig:method_compare} shows that existing SOTA text-driven I2I methods struggle at producing I2I results with large appearance change from the reference images, while our method with high-FBS excels in generating images with significantly different appearance, and is thus more suitable to image creation scenario where appearance divergence is pursued. Among the compared approaches, our method is the only one that enables flexible control over different guiding factors of the reference image, and is also the only approach that simultaneously dispenses with model training, fine-tuning, online optimization, and attention modulations.

An advantage of our approach over related methods is sampling diversity. As displayed in Fig. \ref{fig:diversified_sampling}, our FBSDiff can produce diverse text-guided I2I results by randomly sampling $\tilde{z}_{T}$ from isotropic Gaussian distribution, while other inversion-based methods \cite{mokady2023null,tumanyan2023plug,dong2023prompt,parmar2023zero,li2023stylediffusion} lack such sampling diversity due to directly initializing $\tilde{z}_{T}$ with the inverted feature embedding of the reference image. 

\begin{figure*}[htbp]
    \centering
    \includegraphics[width=\textwidth]{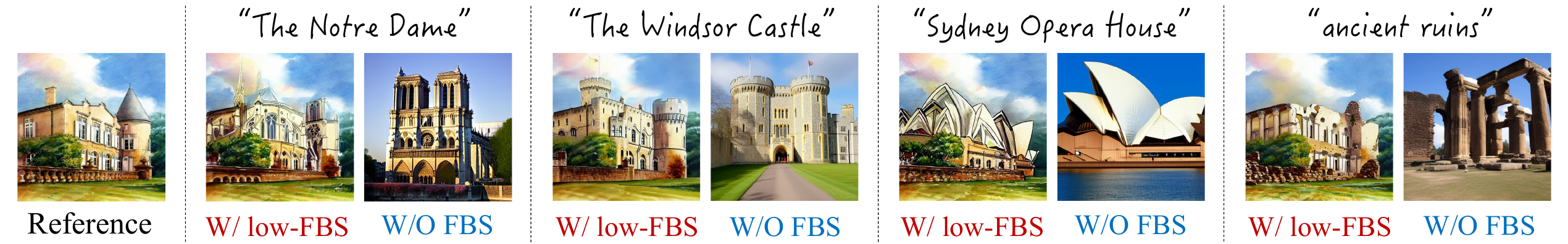}
    \caption{Comparison between results of our method with low-FBS and without frequency band substitution.}
    \label{fig:FBS_ablation}
\end{figure*}

\begin{figure*}[htbp]
    \centering
    \includegraphics[width=\textwidth]{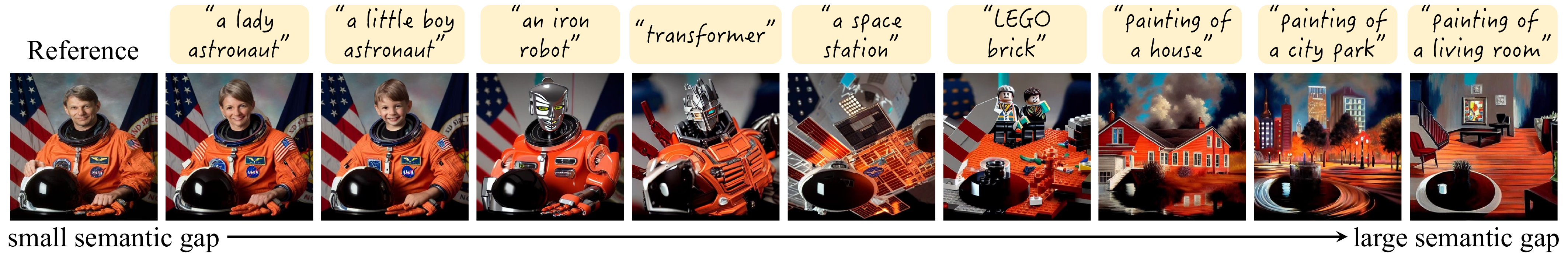}
    \caption{Our method well adapts to varying degrees of semantic gap between the reference image and the target text prompt.}
    \label{fig:semantic_adaptation}
\end{figure*}

The importance of frequency band substitution (FBS) for reference image control is clearly shown in Fig. \ref{fig:FBS_ablation}, from which we see that low-FBS establishes appearance and layout correlations between the reference and the generated images, while removing frequency band substitution leads to results without any correlation to the reference images. Moreover, as Fig. \ref{fig:semantic_adaptation} displays, our method robustly adapts to varying degrees of semantic gap between the reference image and the target text prompt. The translated image of our method can still comply with the target text accurately with satisfying visual quality even in the case of very large image-text semantic discrepancy.

\begin{figure*}[htbp]
    \centering
    \includegraphics[width=\textwidth]{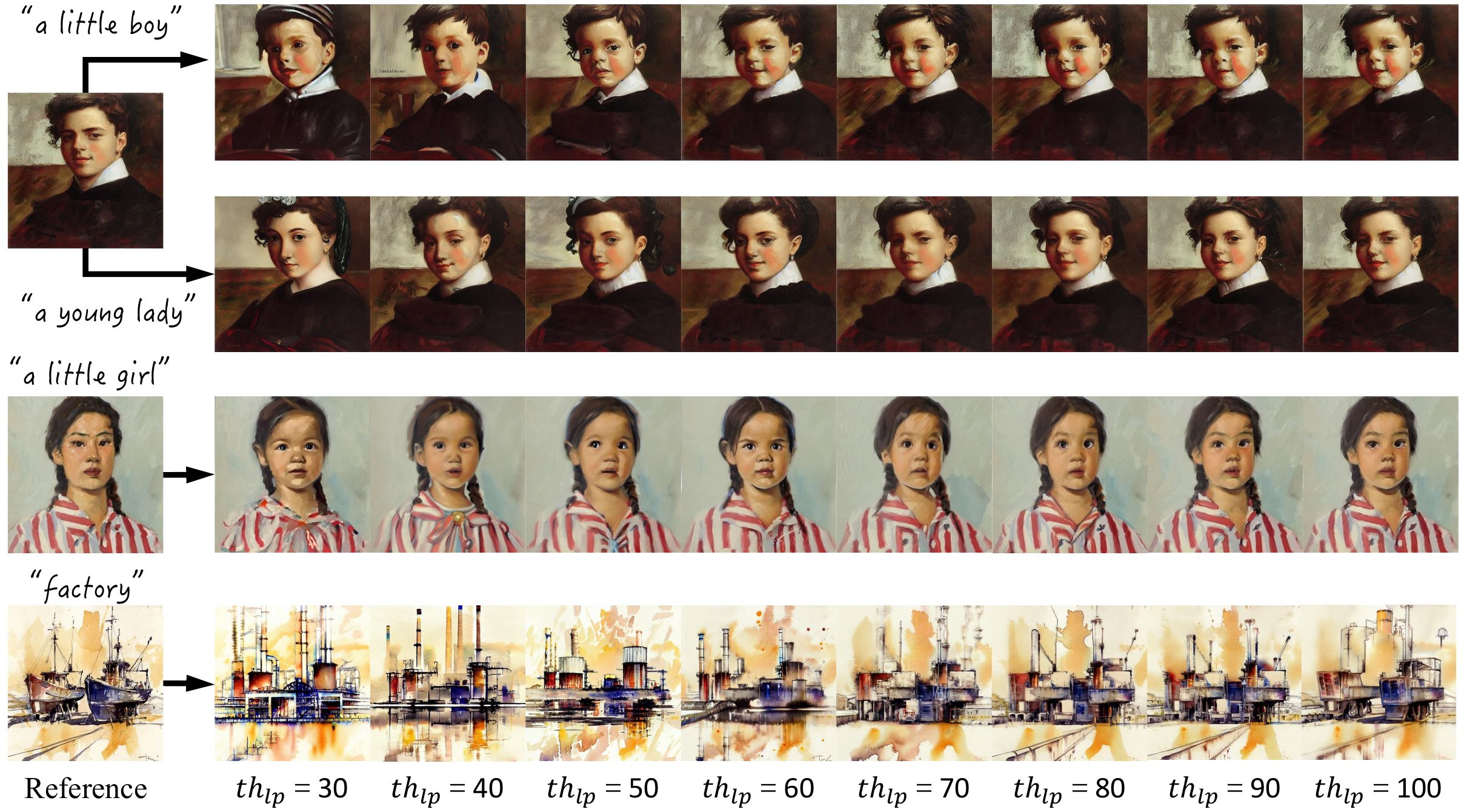}
    \caption{Demonstration of our method in controlling the appearance and layout guiding intensity of the reference image by varying the $th_{lp}$ in low-FBS.}
    \label{fig:low_control}
\end{figure*}

Besides the controllability in the guiding factors of the reference image, our method also allows continuous control over the guiding intensity simply by modulating the bandwidth of the substituted frequency band. Results displayed in Fig. \ref{fig:low_control} demonstrate the image appearance and layout guiding intensity control of our method by adjusting the low-pass filtering threshold $th_{lp}$ in the mode of low-FBS. Enlarging the value of $th_{lp}$ widens the bandwidth of the transplanted low-frequency band and thus increases the amount of guiding information of the reference image, leading to the translated image with more resemblance to the reference image. Conversely, lowering the value of $th_{lp}$ narrows the bandwidth of the substituted frequency band, which reduces the amount of guiding information and thus brings more variations to the translated result as compared with the reference image. 

\begin{figure}[t]
    \centering
    \includegraphics[width=3.5in]{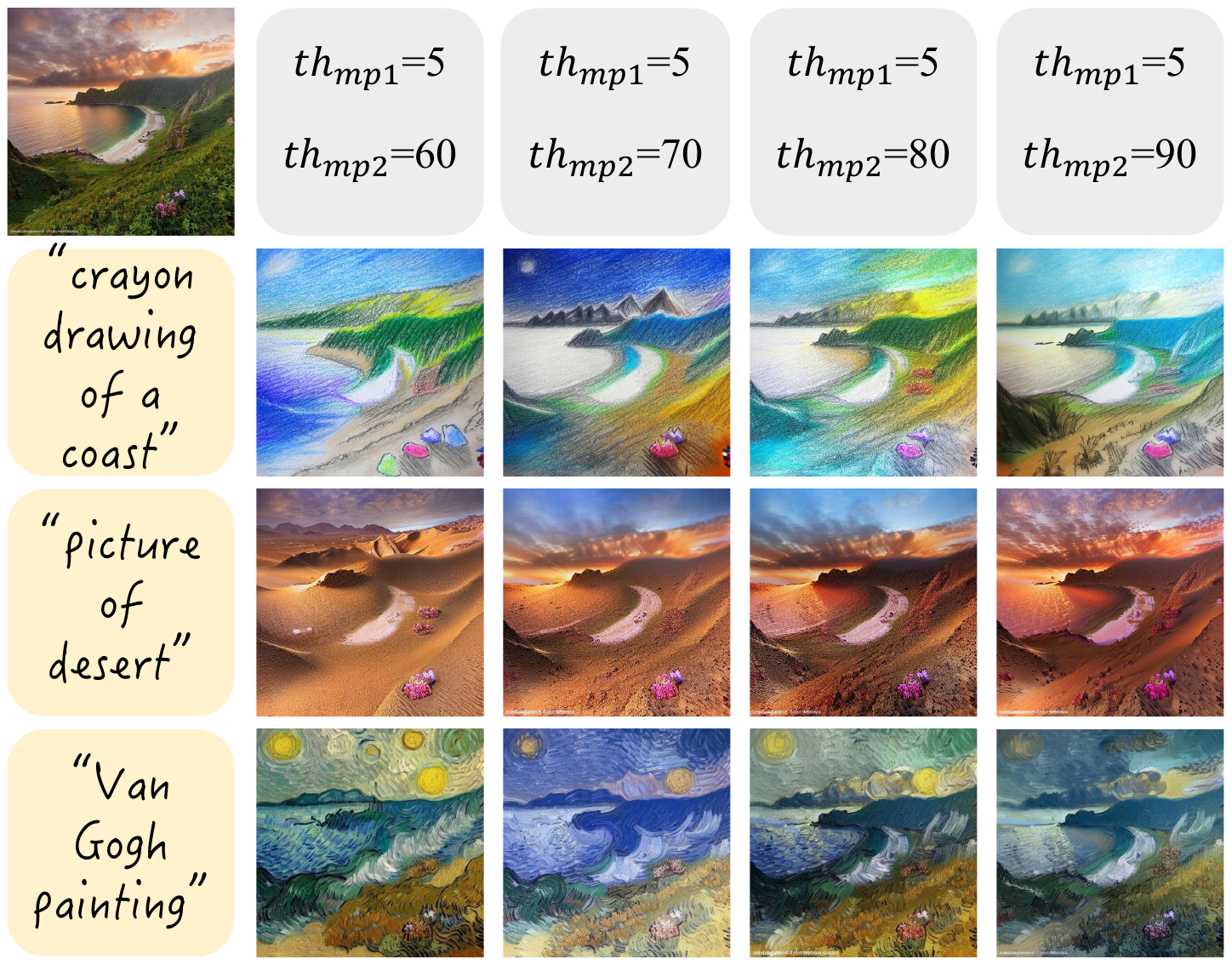}
    \caption{Demonstration of our method in controlling the contour guiding intensity of the reference image by varying the $th_{mp2}$ in mid-FBS.}
    \label{fig:high_control}
\end{figure}

Likewise, results in Fig. \ref{fig:high_control} demonstrate the image contour guiding intensity control of our method by adjusting the mid-pass filtering upper bound threshold $th_{mp2}$ in the mode of mid-FBS. When increasing the value of $th_{mp2}$, more high-frequency components of the reference image (high-frequency guiding information) are included into the transplanted frequency band and transferred to the sampling trajectory, which results in more consistent image contours between the reference image and the translated image. On the contrary, decreasing the value of $th_{mp2}$ shrinks the transplanted high-frequency guiding information and thus leads to weaker image contour consistency. 

\begin{table*}[t]\large
\begin{center}
\caption{Quantitative evaluations of text-driven I2I translation methods.}
\label{tab:metrics}
\resizebox{\textwidth}{!}{
\begin{threeparttable}
\begin{tabular}{|c|ccccc|cccc|}
\hline
{\textbf{Emphasis}} & \multicolumn{5}{|c|}{\textbf{Pursuing image appearance consistency}} & \multicolumn{4}{|c|}{\textbf{Pursuing image appearance divergence}} \\
\hline
\diagbox[]{\textbf{Methods}}{\textbf{Metrics}} & \makecell[c]{Structure\\Similarity($\uparrow$)} & \makecell[c]{LPIPS($\downarrow$)} & \makecell[c]{AdaIN Style\\ Loss($\downarrow$)} & \makecell[c]{CLIP\\Similarity($\uparrow$)} & \makecell[c]{Aesthetic\\Score($\uparrow$)} &  \makecell[c]{Structure\\Similarity($\uparrow$)} & \makecell[c]{AdaIN Style\\ Loss($\uparrow$)} & \makecell[c]{CLIP\\Similarity($\uparrow$)} & \makecell[c]{Aesthetic\\Score($\uparrow$)}\\

\hline
{PAP \cite{tumanyan2023plug}} & 0.954 & 0.272 & 20.440 & \textcolor{red}{0.287} & \textcolor{red}{6.590} & 0.956 & 28.337 & \textcolor{blue}{0.279} & \textcolor{blue}{6.458} \\

{Null-text \cite{mokady2023null}} & 0.948 & 0.247 & 17.546 & 0.276 & 6.505 & 0.952 & 22.545 & 0.270 & 6.402 \\

{Pix2Pix-zero \cite{parmar2023zero}} & 0.951 & \textcolor{blue}{0.243} & \textcolor{blue}{16.875} & 0.262 & 6.484 & 0.953 & 21.240 & 0.258 & 6.344 \\

{InsPix2Pix \cite{brooks2023instructpix2pix}} & \textcolor{blue}{0.958} & 0.266 & 23.373 & 0.258 & 6.269 & \textcolor{red}{0.965} & \textcolor{blue}{30.804} & 0.264 & 6.196 \\

{PT-inversion \cite{dong2023prompt}} & 0.947 & 0.248 & 21.667 & 0.271 & 6.481 & 0.948 & 24.367 & 0.267 & 6.285 \\

{StyleDiffusion \cite{li2023stylediffusion}} & 0.944 & 0.251 & 22.484 & 0.267 & 6.477 & 0.947 & 25.166 & 0.260 & 6.267 \\

{\textbf{FBSDiff (ours)}} & \textcolor{red}{0.962} & \textcolor{red}{0.241} & \textcolor{red}{15.452} & \textcolor{blue}{0.285} & \textcolor{blue}{6.583} & \textcolor{blue}{0.964} & \textcolor{red}{33.875} & \textcolor{red}{0.281} & \textcolor{red}{6.463} \\

\hline
\end{tabular}
\begin{tablenotes}
        \footnotesize
        \item The \textcolor{red}{red font} indicates the top-ranked value and the \textcolor{blue}{blue font} indicates the second-ranked value.
      \end{tablenotes}
\end{threeparttable}
}
\end{center}
\end{table*}

\subsection{Ablation Study}
To verify the rationality and effectiveness of our proposed method, we also explore and compare with other designs of frequency band substitution, including substituting the frequency band only once at $\lambda T$ time step rather than along the whole calibration phase (which we denote \textbf{Once Substitution}), and substituting the full DCT spectrum rather than a partial sub-band of it (which we refer to as \textbf{Full Substitution}). 

\begin{figure}[t]
    \centering
    \includegraphics[width=3.5in]{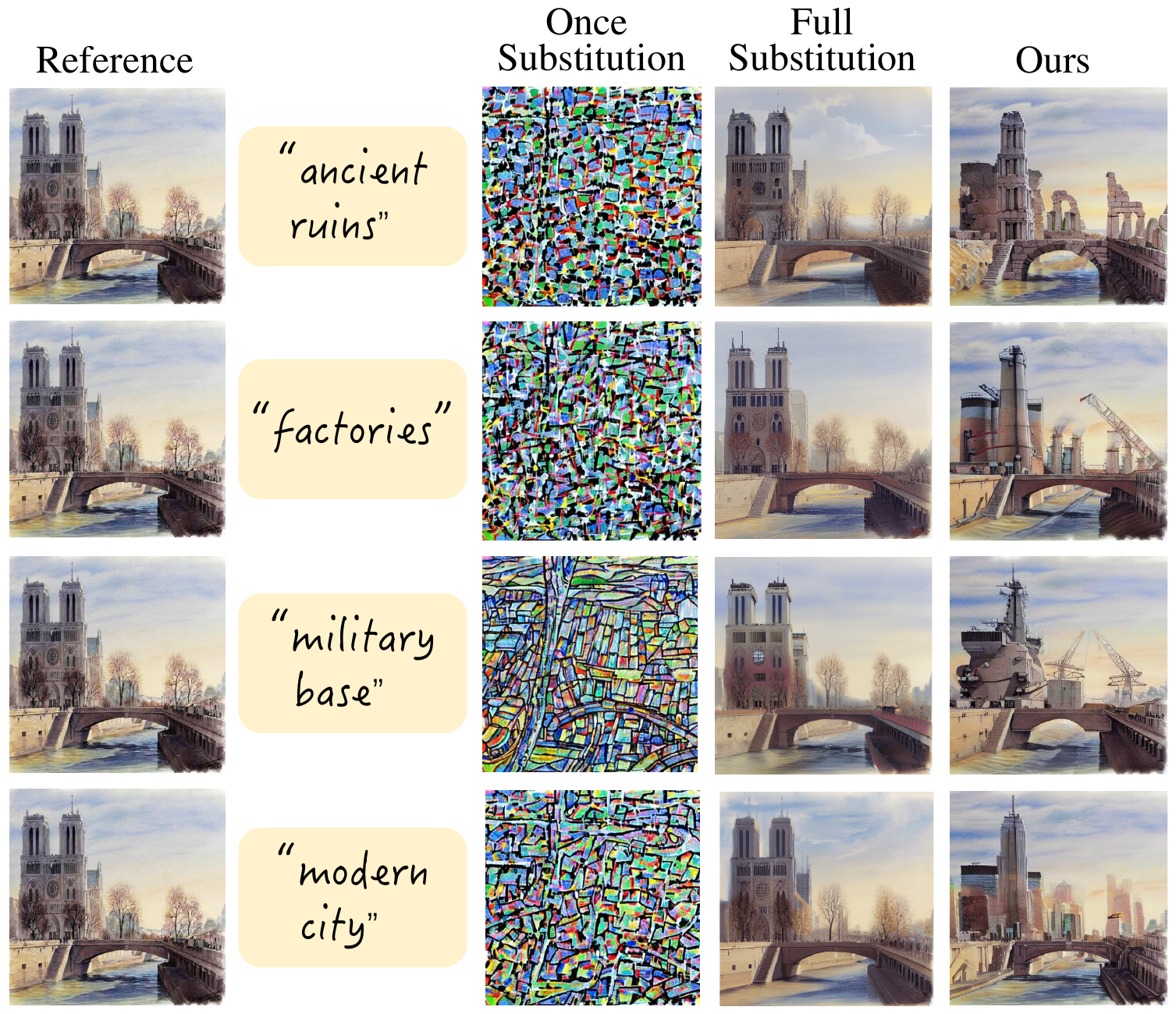}
    \caption{Ablation study w.r.t. different manners of frequency band substitution.}
    \label{fig:subtitute_ablation}
\end{figure}

The image translation results of different designs of frequency band substitution (FBS) are displayed in Fig. \ref{fig:subtitute_ablation}. It shows that Once Substitution produces severely noisy results rather than reasonable images, which indicates that step-by-step FBS along the whole calibration phase is of crucial importance for smooth and stable information fusion. Since image content is basically formed in the early stage of the diffusion sampling process, removing per-step feature calibration of FBS in the early sampling process will inevitably lead to large deviation of the sampling trajectory against the reconstruction trajectory. In this case, substituting a frequency band at an intermediate time step will cause completely incoherent 2D DCT spectrum, and thus leads to abnormal image translation results after converting the diffusion features back to the spatial domain. 

Besides, it also shows that Full Substitution fails to manipulate the reference image as per the text prompt. This is because substituting the full DCT spectrum is equivalent to complete feature replacement, which makes the sampling trajectory totally the same as the reconstruction trajectory during the calibration phase, the early part of the diffusion sampling process that dominates the forming of image content. Therefore, the generated image content is forced to be the same as the reference image after the calibration phase and is difficult to be modified noticeably during the subsequent non-calibration phase, the latter part of the diffusion sampling process that focuses on refining fine-grained image details rather than coarse-grained image content. Thus, the sampling results of Full Substitution closely resemble the reference images, lacking editability and text fidelity. 

\subsection{Quantitative Evaluations}
For quantitative method evaluation, we separately evaluate methods on the text-driven I2I translation task pursuing image appearance consistency and the task pursuing image appearance divergence. For the former task, we assess models' appearance and layout preservation ability by measuring Structure Similarity ($\uparrow$), Perceptual Similarity ($\uparrow$), and Style Distance ($\downarrow$) between the reference image and the translated image pair. For the latter task, we assess models' contour preservation and appearance alteration capabilities by measuring Structure Similarity ($\uparrow$) and Style Distance ($\uparrow$) between I2I translation pairs. For Structure Similarity measurement, we use DINO-ViT self-similarity distance \cite{tumanyan2022splicing} as the metric for Structure Distance between two images, and define Structure Similarity as 1 - Structure Distance. We use LPIPS \cite{zhang2018unreasonable} metric to measure Perceptual Similarity, and use AdaIN style loss \cite{huang2017arbitrary} to measure Style Distance between I2I pairs. Besides, CLIP Similarity ($\uparrow$) metric is used to measure semantic consistency between the target text prompt and the translated image, i.e., text fidelity of the I2I translation results. Finally, we evaluate Aesthetic Score ($\uparrow$) of the translated images via the pre-trained LAION Aesthetics Predictor V2 model.

We sample reference images from the LAION Aesthetics 6.5+ dataset for quantitative evaluation. For the above-mentioned two tasks, we separately sample 500 reference images for each task and manually design 2 editing text prompts for each reference image, resulting in 1000 evaluation samples (reference image and target text pairs) for each task. For evaluation of our method, we use low-FBS for the task pursuing appearance consistency and use high-FBS for the task pursuing appearance divergence. The average values of all the evaluation metrics are reported in Tab. \ref{tab:metrics}. Our method achieves top rankings for all the metrics in both two tasks, indicating superiority of our method in layout and appearance preservation with low-FBS, as well as simultaneous contour preservation and appearance modification with high-FBS. Moreover, the competitive results in CLIP Similarity and Aesthetic Score reflect that our method can generate I2I translation results with high text fidelity and visual quality.

\begin{table*}[htbp]\small 
\begin{center}   
\caption{Comparison of our approach with related models in method properties.}  
\label{tab: method property} 
\small
\renewcommand\arraystretch{1}
\resizebox{\textwidth}{!}{
\begin{tabular}{|c|c|c|c|c|c|c|}
\hline \textbf{Methods} & Training free & Fine-tuning free & Optimization free & Source-text free & Attention free & Backbone invariant \\
\hline 

Null-text \cite{mokady2023null} &  \Checkmark & \Checkmark & \XSolidBrush & \XSolidBrush & \XSolidBrush & \XSolidBrush \\      
PAP \cite{tumanyan2023plug} & \Checkmark & \Checkmark & \Checkmark & \XSolidBrush & \XSolidBrush & \XSolidBrush \\
Pix2Pix-zero \cite{parmar2023zero} & \Checkmark & \Checkmark & \XSolidBrush & \XSolidBrush & \XSolidBrush & \XSolidBrush \\ 
InsPix2Pix \cite{brooks2023instructpix2pix} & \XSolidBrush & \Checkmark & \Checkmark & \Checkmark & \Checkmark & \Checkmark \\
PT-inversion \cite{dong2023prompt} & \Checkmark & \Checkmark & \XSolidBrush & \XSolidBrush & \XSolidBrush & \XSolidBrush \\
StyleDiffusion \cite{li2023stylediffusion} & \Checkmark & \Checkmark & \XSolidBrush & \XSolidBrush & \XSolidBrush & \XSolidBrush \\
VQCLIP \cite{crowson2022vqgan} & \Checkmark & \Checkmark & \XSolidBrush & \Checkmark & \Checkmark & \XSolidBrush \\
DiffuseIT \cite{kwon2022diffusion} & \Checkmark & \Checkmark & \XSolidBrush & \Checkmark & \Checkmark & \XSolidBrush \\
DiffusionCLIP \cite{kim2022diffusionclip} & \Checkmark & \XSolidBrush & \Checkmark & \Checkmark & \Checkmark & \XSolidBrush \\
Design Booster \cite{sun2023design} & \XSolidBrush & \Checkmark & \Checkmark & \Checkmark & \Checkmark & \XSolidBrush \\
SINE \cite{zhang2023sine} & \Checkmark & \XSolidBrush & \XSolidBrush & \XSolidBrush & \Checkmark & \Checkmark \\
Imagic \cite{kawar2023imagic}& \Checkmark & \XSolidBrush & \XSolidBrush & \Checkmark & \Checkmark & \Checkmark \\
\textbf{FBSDiff (Ours)} & \Checkmark & \Checkmark & 
  \Checkmark & \Checkmark & \Checkmark & \Checkmark \\
\hline   
\end{tabular}}
\end{center}   
\end{table*}

We compare our FBSDiff with related text-driven I2I translation methods in method properties, results are summarized in Tab. \ref{tab: method property}. Among the compared approaches, our method is the only one that possesses all the following advantages: 
\begin{itemize}
    \item Dispense with model training;
    \item Dispense with model fine-tuning;
    \item Dispense with online-optimization at inference time;
    \item Dispense with paired source text of the reference image;
    \item Dispense with attention modulation operations inside the denoising network;
    \item Invariant to the specific architecture of the backbone diffusion model.
\end{itemize}

\begin{figure*}[t]
    \centering
    \includegraphics[width=0.86\textwidth]{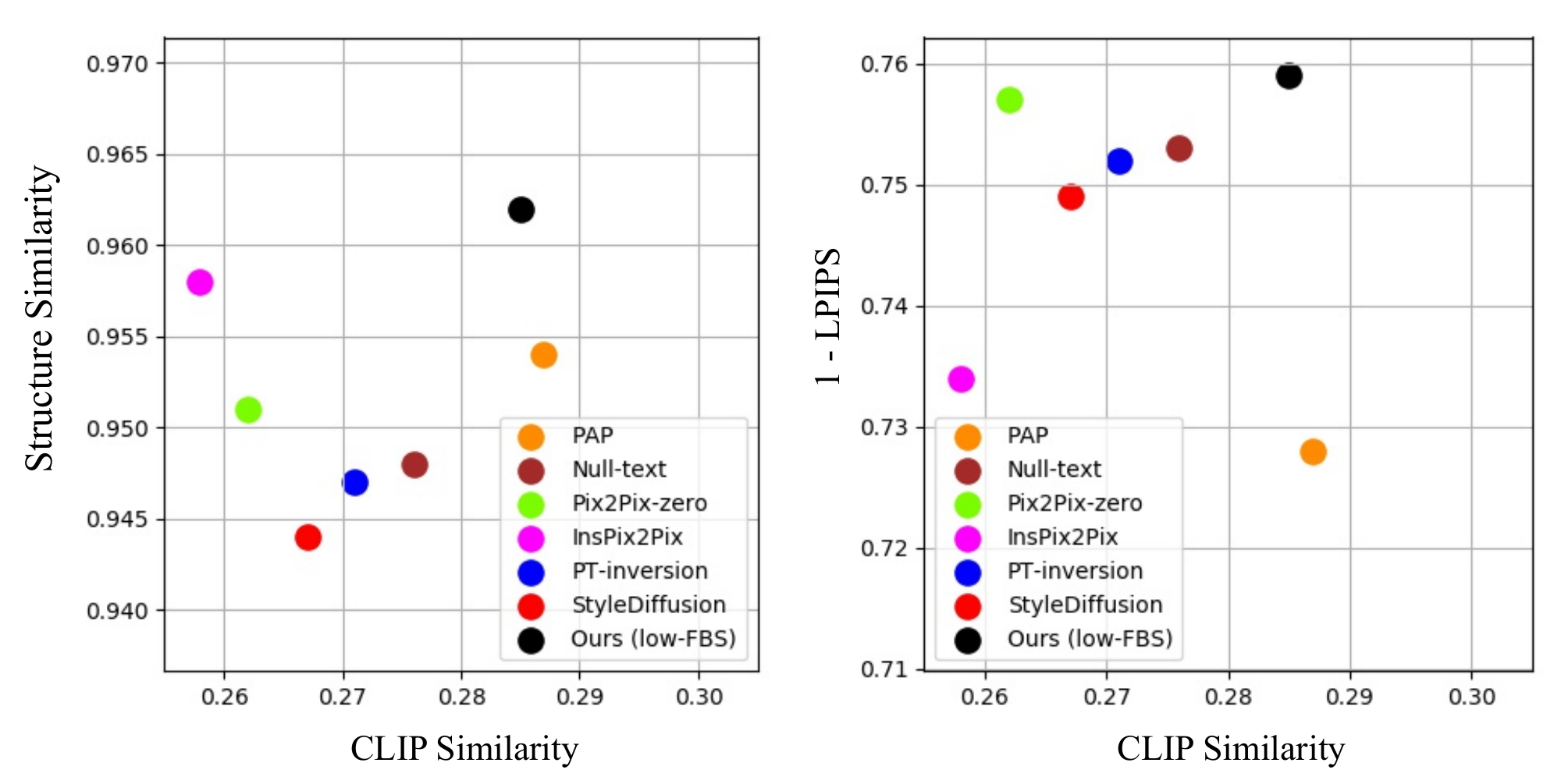}
    \caption{Visualization of the quantitative method comparison for I2I translation task \textbf{pursuing image appearance consistency}. Left: comparison in CLIP Similarity ($\uparrow$) and Structure Similarity ($\uparrow$). Right: comparison in CLIP Similarity ($\uparrow$) and (1-LPIPS) ($\uparrow$). Our method with \textbf{low-FBS} achieves the most top-right position in both two scatters, indicating the best trade-off achieved by our method (low-FBS) in I2I translation appearance consistency and text fidelity.}
    \label{fig:pursue_appearance_consistency_vis}
\end{figure*}

\begin{figure*}[t]
    \centering
    \includegraphics[width=0.86\textwidth]{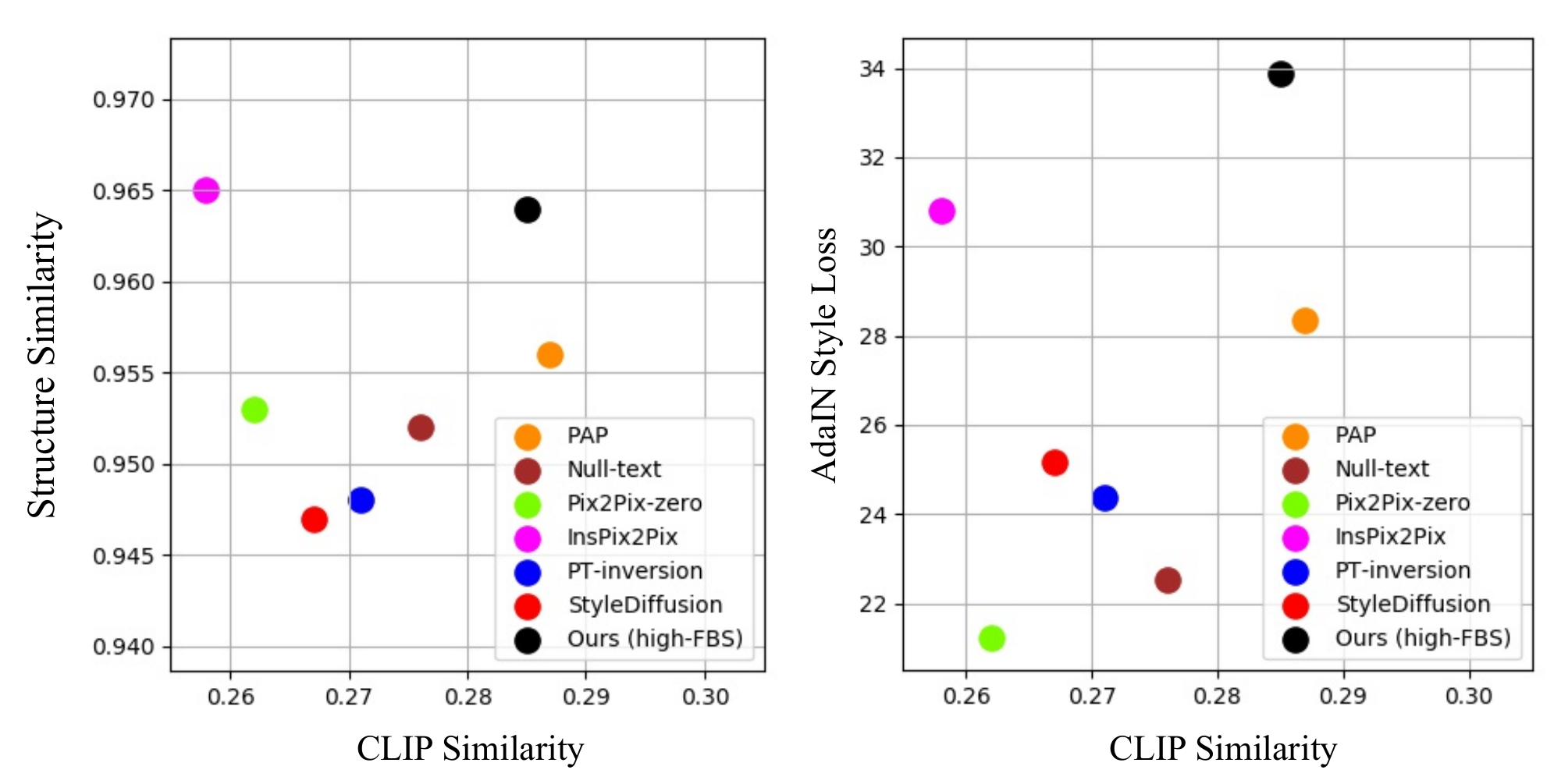}
    \caption{Visualization of method comparison for the image translation task \textbf{pursuing image appearance} divergence. Left: comparison in CLIP Similarity ($\uparrow$) and Structure Similarity ($\uparrow$). Right: comparison in CLIP Similarity ($\uparrow$) and AdaIN Style Loss ($\uparrow$). Our method with \textbf{high-FBS} achieves the most top-right position in both two scatters, indicating the best trade-off achieved by our method (high-FBS) in I2I translation appearance divergence and text fidelity.}
    \label{fig:pursue_appearance_divergence_vis}
\end{figure*}

For quantitative evaluation reported in Tab. \ref{tab:metrics}, we visualize partial results to highlight the superiority of our method over related approaches. For I2I task pursuing image appearance consistency, we display the scatter plot about Structure Similarity ($\uparrow$) and CLIP Similarity ($\uparrow$), and the scatter plot about (1-LPIPS) ($\uparrow$) and CLIP Similarity ($\uparrow$) in Fig. \ref{fig:pursue_appearance_consistency_vis}. Results show that our method with low-FBS achieves the most top-right position in both two scatter plots, indicating the best trade-off achieved by our method (low-FBS) in I2I translation appearance consistency and text fidelity. For I2I task pursuing image appearance divergence, we display the scatter plot about Structure Similarity ($\uparrow$) and CLIP Similarity ($\uparrow$), and the scatter plot about AdaIN Style Loss ($\uparrow$) and CLIP Similarity ($\uparrow$) in Fig. \ref{fig:pursue_appearance_divergence_vis}. Results also show the most top-right position achieved by our method with high-FBS in both two plots, indicating the best trade-off achieved by our method (high-FBS) in I2I translation appearance divergence and text fidelity. 

\section{Conclusion}
This paper proposes FBSDiff, a plug-and-play method adapting pre-trained T2I diffusion model to highly controllable text-driven I2I translation. At the heart of our method is decomposing different guiding factors of the reference image in the diffusion feature DCT space, and dynamically transplanting a certain DCT frequency band from diffusion features along the reconstruction trajectory into the corresponding features along the sampling trajectory, which is realized via our proposed frequency band substitution layer. Experiments demonstrate that our method allows flexible control over both guiding factors and guiding intensity of the reference image simply by tuning the type and bandwidth of the substituted frequency band, respectively. In summary, our FBSDiff provides a novel solution to text-driven I2I translation from a frequency-domain perspective, integrating advantages in versatility, high controllability, high visual quality, and plug-and-play efficiency.

\appendices
\section{Diffusion Model Background}
The Denoising Diffusion Probabilistic Model (DDPM) is a latent variable model that comprises a forward noising diffusion process and a reverse denoising diffusion process. Starting with a given data distribution $x_{0}\sim q(x_{0})$, the forward diffusion process employs a T-step Markov chain to repeatedly add Gaussian noise to the original data $x_{0}$ according to $q(x_{t}|x_{t-1})$ defined as follows:
\begin{equation}
    q(x_{t}|x_{t-1}):=\mathcal{N}(x_{t}; \sqrt{\alpha_{t}}x_{t-1}, (1-\alpha_{t})\mathrm{I}),
\end{equation}
where $\alpha_{t} \in (0, 1)$, and $\alpha_{t} \geq \alpha_{t+1}$. Using the notation $\bar{\alpha}_{t}:=\prod_{i=1}^{t}\alpha_{i}$, we can derive the marginal distribution $q(x_{t}|x_{0})$ as follows:
\begin{equation}
    q(x_{t}|x_{0}):=\mathcal{N}(x_{t}; \sqrt{\bar{\alpha}_{t}}x_{0}, (1-\bar{\alpha}_{t})\mathrm{I}),
    \label{xt|x0}
\end{equation}

where $\sqrt{\bar{\alpha}_{T}}$ approaches to $0$. With the above forward noising diffusion process, the source data distribution will be transformed into an isotropic Gaussian distribution. 

The reverse denoising diffusion process conversely converts the isotropic Gaussian distribution to the data distribution by gradually estimating and sampling from the posterior distribution $q(x_{t-1}|x_{t})$. However, $q(x_{t-1}|x_{t})$ is difficult to estimate while $q(x_{t-1}|x_{t}, x_{0})$ is tractable with some algebraic manipulation:
\begin{equation}
    q(x_{t-1}|x_{t},x_{0}):=\mathcal{N}(x_{t-1}; \tilde{\mu}_{t}(x_{t}, x_{0}), \tilde{\beta}_{t}\mathrm{I}),
\end{equation}
\begin{equation}
    \tilde{\mu}_{t}(x_{t}, x_{0}):=\frac{\sqrt{\bar{\alpha}_{t-1}}\beta_{t}}{1-\bar{\alpha}_{t}}x_{0}+\frac{\sqrt{\alpha}_{t}(1-\bar{\alpha}_{t-1})}{1-\bar{\alpha}_{t}}x_{t},
\end{equation}
\begin{equation}
    \tilde{\beta}_{t}:=\frac{1-\bar{\alpha}_{t-1}}{1-\bar{\alpha}_{t}}\beta_{t},
\end{equation}
where $\beta_{t}:=1-\alpha_{t}$. Though no $x_{0}$ is available at inference time, its approximate value can be estimated according to Eq. \ref{xt|x0}:
\begin{equation}
    y_{\theta}(x_{t}):=\frac{1}{\sqrt{\bar{\alpha}_{t}}}(x_{t}-\sqrt{1-\bar{\alpha}_{t}}\epsilon_{\theta}(x_{t})),
    \label{predicted_x0}
\end{equation}
where $\epsilon_{\theta}(x_{t})$ is the prediction of the Gaussian noise sampled at time step $t$ estimated by the denoising network $\epsilon_{\theta}$, $y_{\theta}(x_{t})$ is the calculated approximation of $x_{0}$. 

For image-to-image translation or text-to-image generation, additional condition (could be an image or a text) is required for noise prediction. In these cases, Eq. \ref{predicted_x0} can be updated as follows:
\begin{equation}
    y_{\theta}(x_{t}, c):=\frac{1}{\sqrt{\bar{\alpha}_{t}}}(x_{t}-\sqrt{1-\bar{\alpha}_{t}}\epsilon_{\theta}(x_{t}, c)),
\end{equation}
where $c$ denotes the additional condition that is involved in the noise prediction and the reverse denoising process.

\section{DCT and IDCT Details}
We perform 2D-DCT to project diffusion features $\textbf{z}$ into the 2D DCT space, obtaining its frequency-domain counterpart $\textbf{f}$ (Eq. \ref{eq:dct}). Conversely, we employ 2D-IDCT to transform diffusion features from the DCT domain back into the spatial domain (Eq. \ref{eq:idct}). The specific form of 2D-DCT and 2D-IDCT are respectively given by Eq. \ref{eq:dct_detail} and Eq. \ref{eq:idct_detal}, in which $f^{(n)}$ and $z^{(n)}$ denote the $n^{th}$ channel of $\textbf{f}$ and $\textbf{z}$ respectively; $i,j$ and $u,v$ are two-dimensional coordinate indices of the spatial domain and DCT frequency domain respectively; $h$ and $w$ denote the height and width of the latent diffusion features; $m(0)=\frac{1}{\sqrt{2}}$, $m(\gamma)=1$ for all $\gamma>0$. It is worth mentioning that though the 2D-DCT and 2D-IDCT are performed on each individual channel of diffusion features (per-channel transformation), our PyTorch implementation with efficient GPU parallel computing capability enables to transform all channels simultaneously, and thus brings negligible additional time overhead during the sampling process. 

\begin{equation}
    \textbf{f}=DCT(\textbf{z}),
    \label{eq:dct}
\end{equation}

\begin{equation}
\begin{aligned}
f^{(n)}_{u,v}=&\frac{2}{\sqrt{hw}}m(u)m(v)\sum\nolimits_{i=0}^{h-1}\sum\nolimits_{j=0}^{w-1}[z^{(n)}_{i,j} \\
    &\cos(\frac{(2i+1)u\pi}{2h})\cos(\frac{(2j+1)v\pi}{2w})],
    \label{eq:dct_detail}
\end{aligned}
\end{equation}

\begin{equation}
    \textbf{z}=IDCT(\textbf{f}),
    \label{eq:idct}
\end{equation}

\begin{equation}
\begin{aligned}
    z^{(n)}_{i,j}=&\frac{2}{\sqrt{hw}}\sum\nolimits_{u=0}^{h-1}\sum\nolimits_{v=0}^{w-1}[m(u)m(v)f^{(n)}_{u,v} \\
    &\cos(\frac{(2i+1)u\pi}{2h})\cos(\frac{(2j+1)v\pi}{2w})].
    \label{eq:idct_detal}
\end{aligned}
\end{equation}

\begin{figure*}[htbp]
    \centering
    \includegraphics[width=\textwidth]{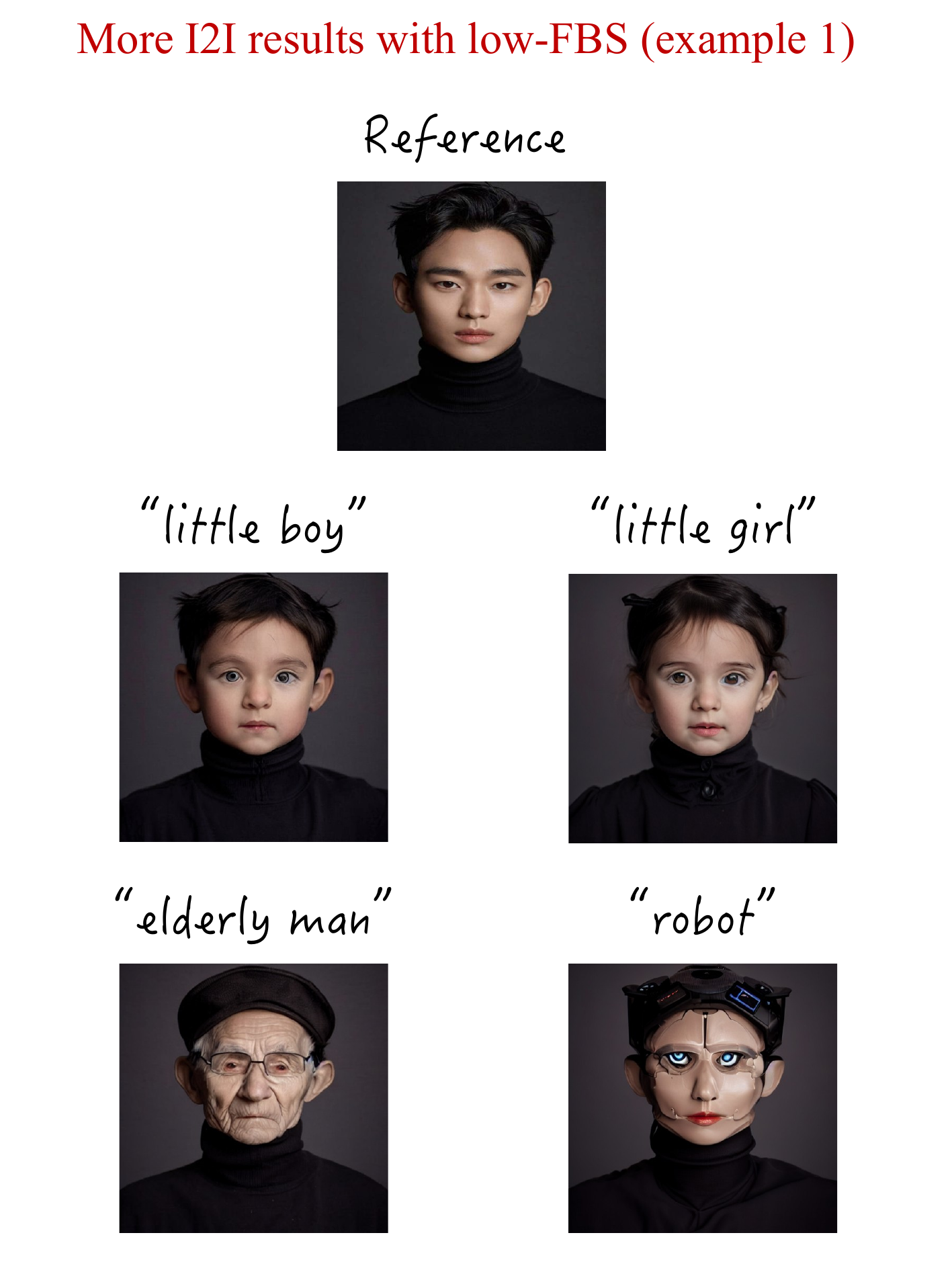}
    \caption{More text-driven I2I results of our method with low-FBS for image appearance and layout control.}
    \label{fig:low_FBS_1}
\end{figure*}

\begin{figure*}[htbp]
    \centering
    \includegraphics[width=\textwidth]{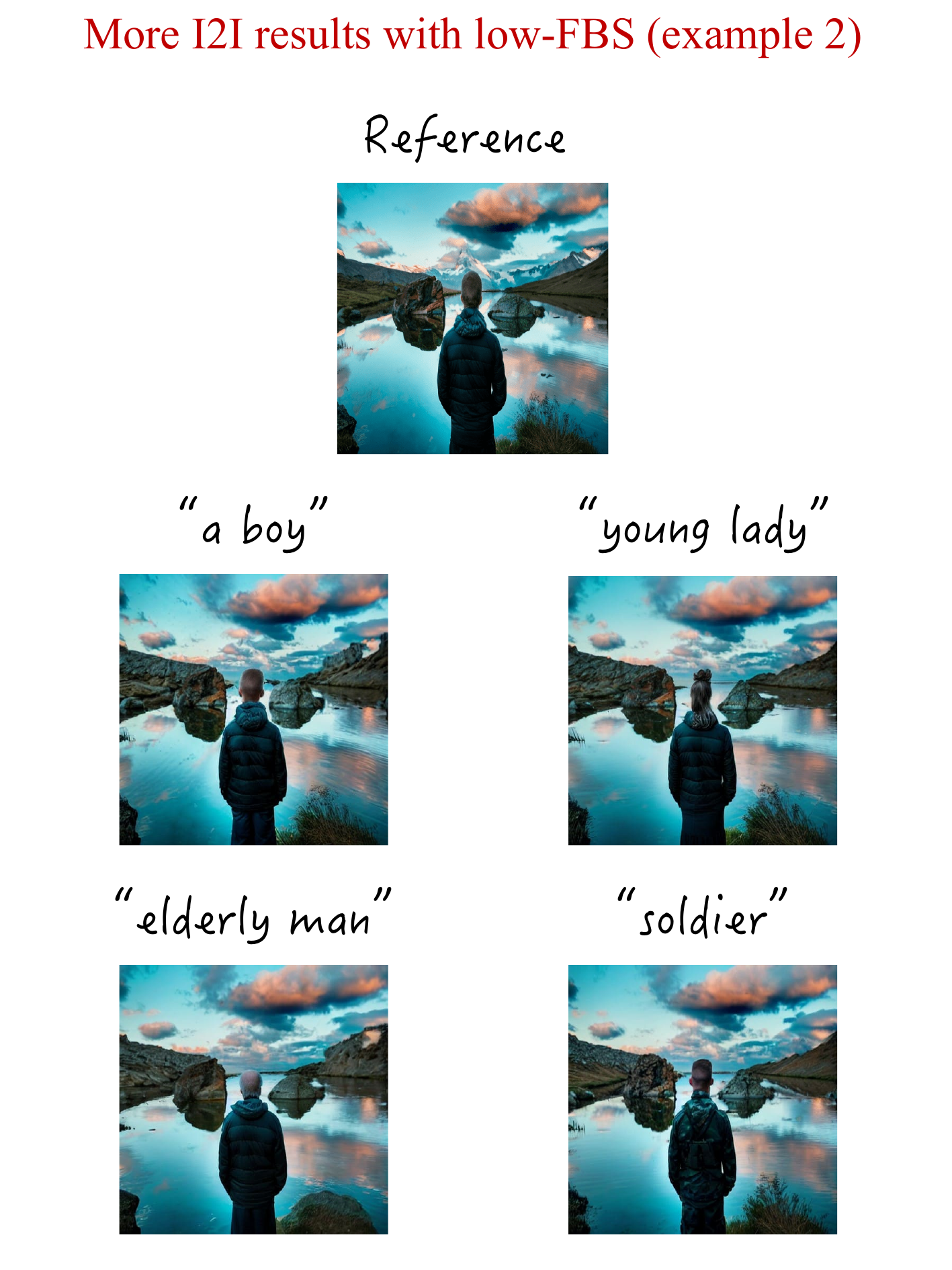}
    \caption{More text-driven I2I results of our method with low-FBS for image appearance and layout control.}
    \label{fig:low_FBS_2}
\end{figure*}

\begin{figure*}[htbp]
    \centering
    \includegraphics[width=\textwidth]{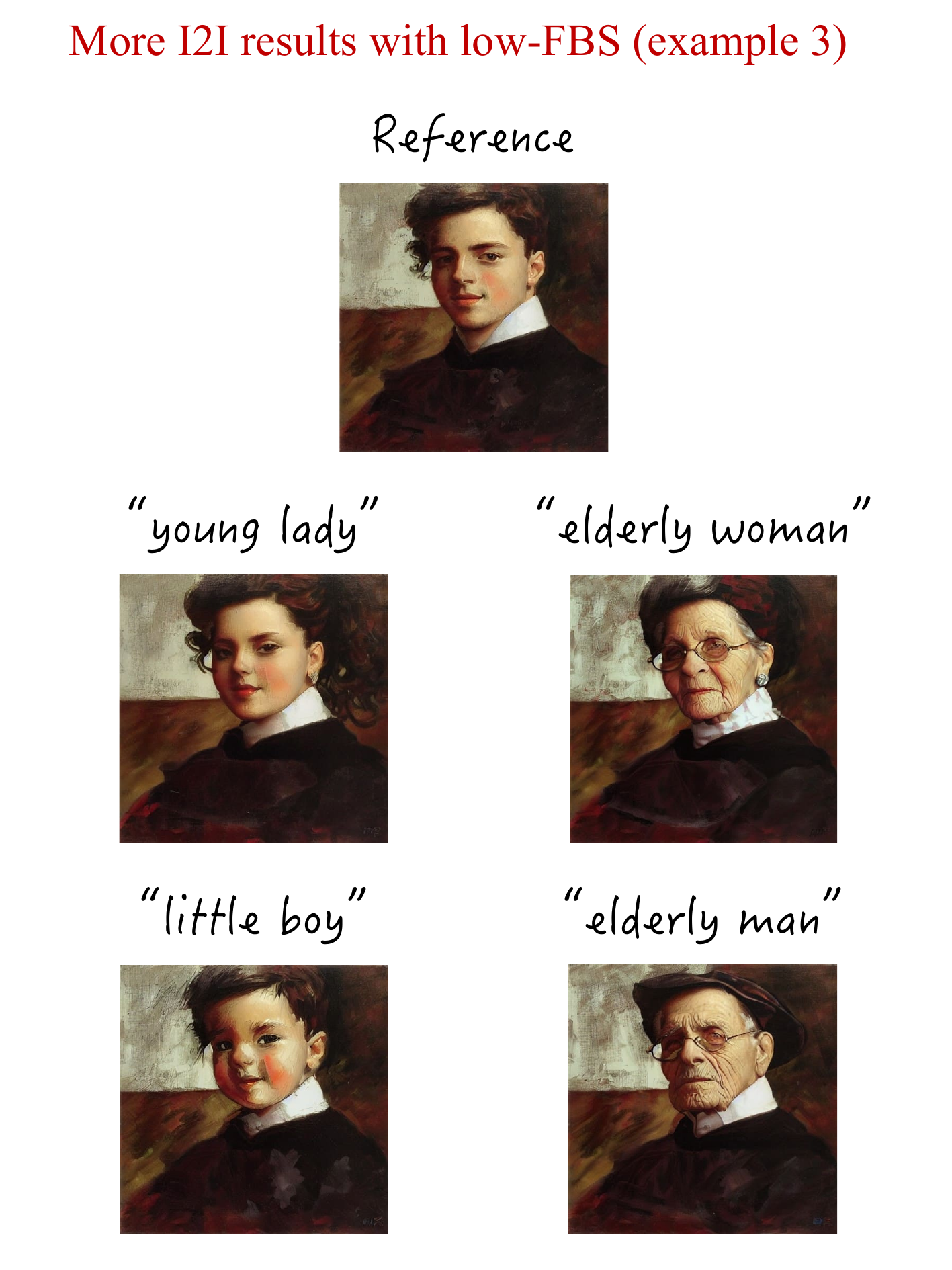}
    \caption{More text-driven I2I results of our method with low-FBS for image appearance and layout control.}
    \label{fig:low_FBS_3}
\end{figure*}

\begin{figure*}[htbp]
    \centering
    \includegraphics[width=\textwidth]{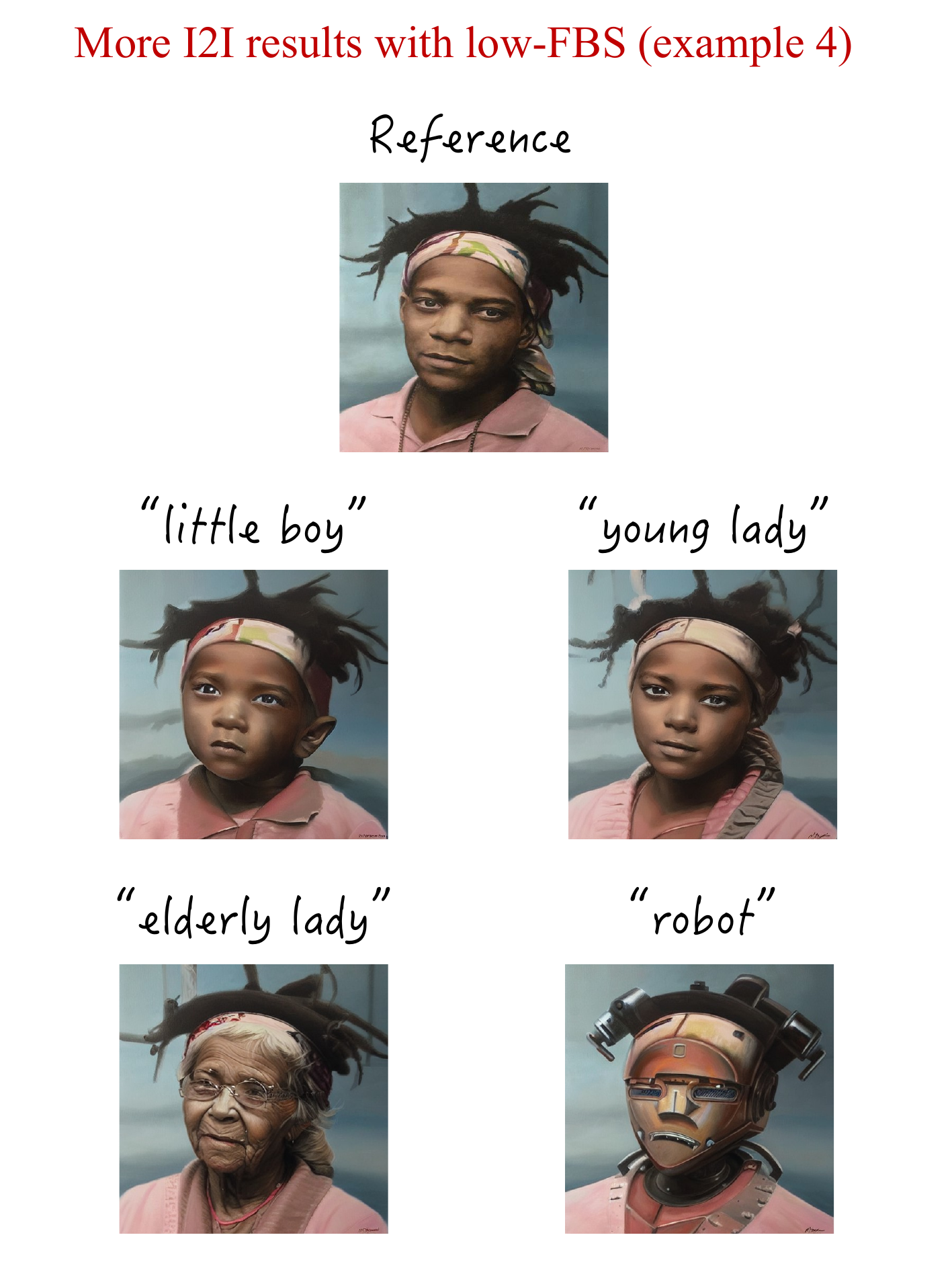}
    \caption{More text-driven I2I results of our method with low-FBS for image appearance and layout control.}
    \label{fig:low_FBS_4}
\end{figure*}

\begin{figure*}[htbp]
    \centering
    \includegraphics[width=\textwidth]{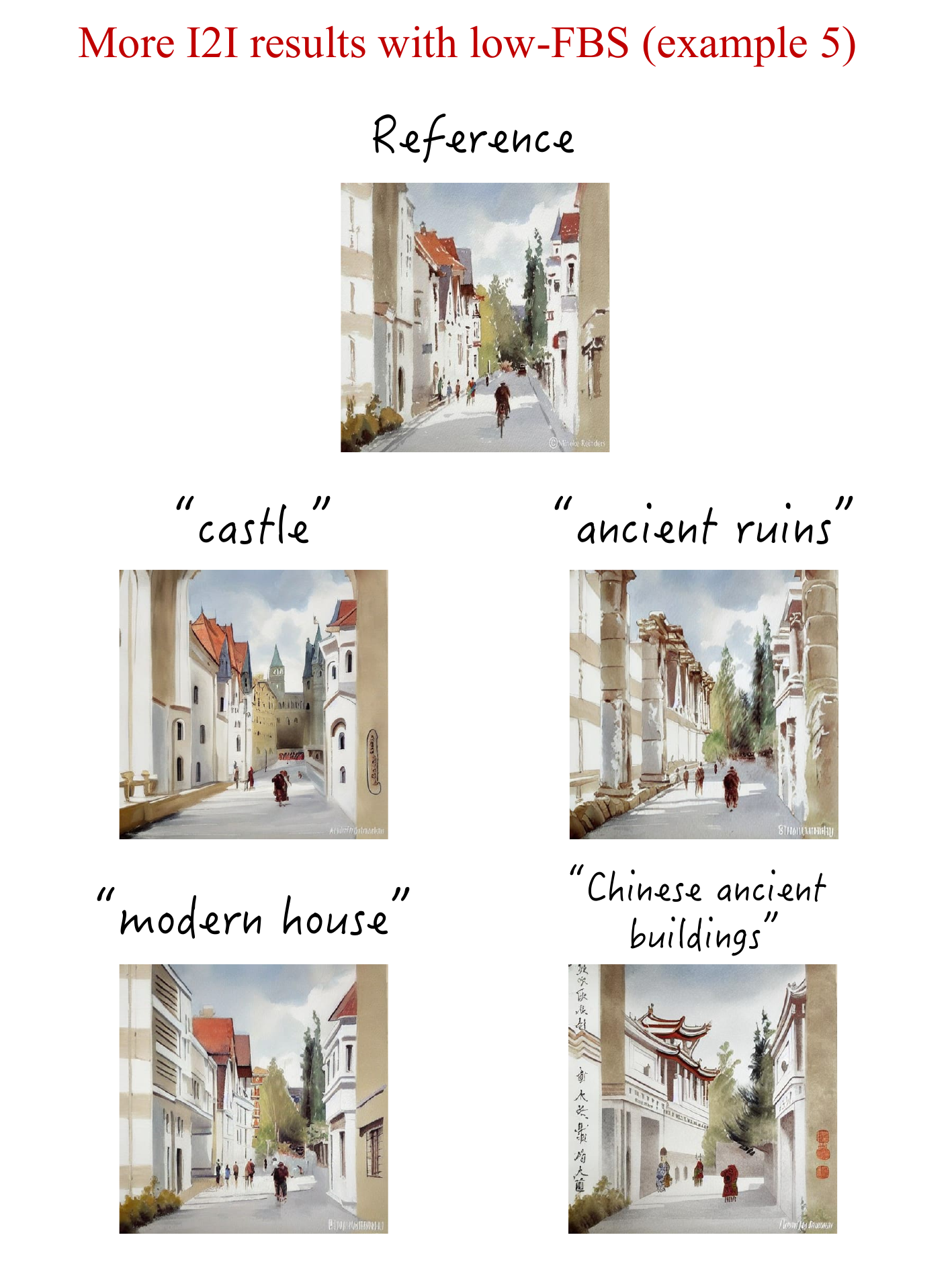}
    \caption{More text-driven I2I results of our method with low-FBS for image appearance and layout control.}
    \label{fig:low_FBS_5}
\end{figure*}

\begin{figure*}[htbp]
    \centering
    \includegraphics[width=\textwidth]{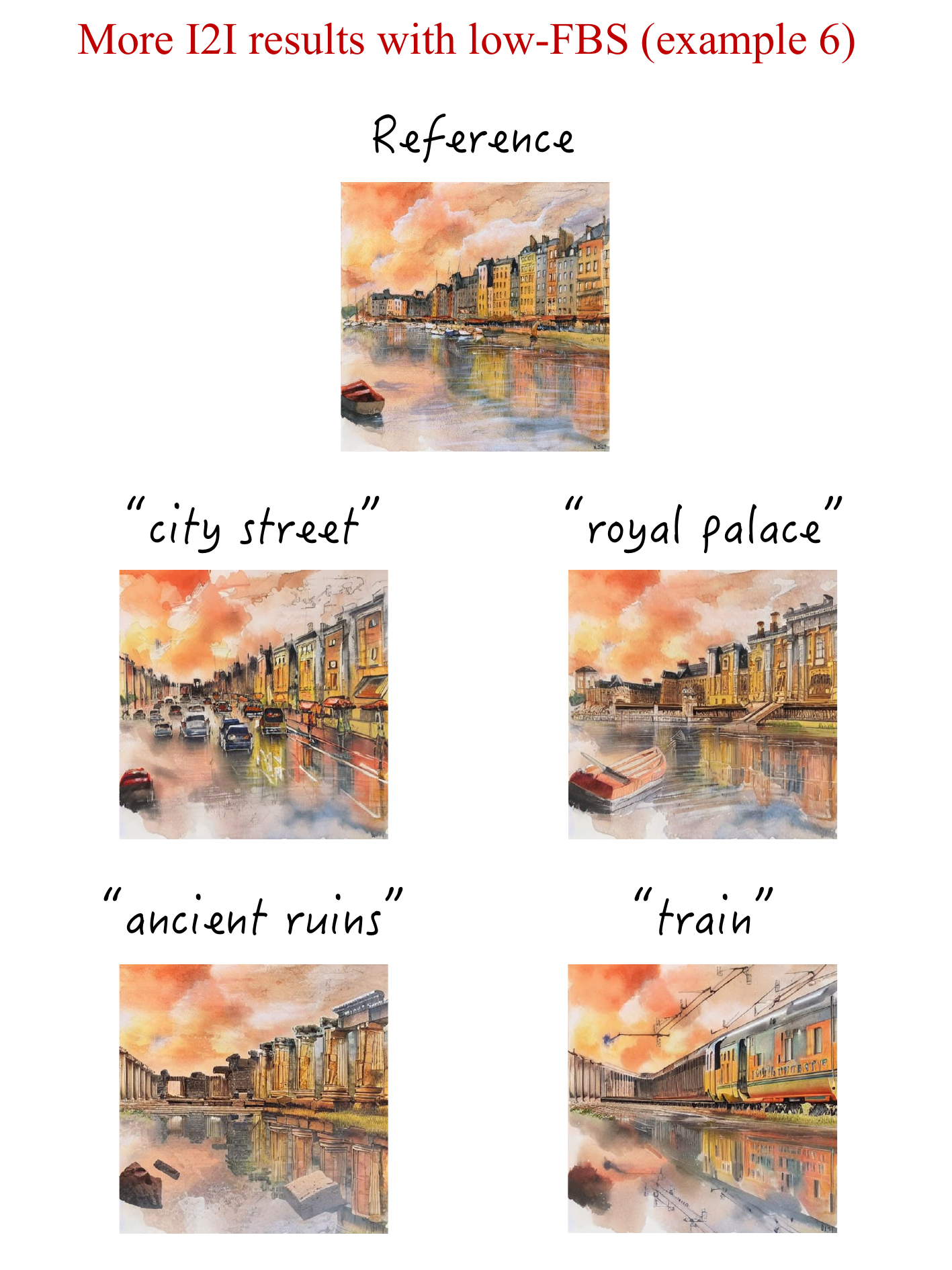}
    \caption{More text-driven I2I results of our method with low-FBS for image appearance and layout control.}
    \label{fig:low_FBS_6}
\end{figure*}

\begin{figure*}[htbp]
    \centering
    \includegraphics[width=\textwidth]{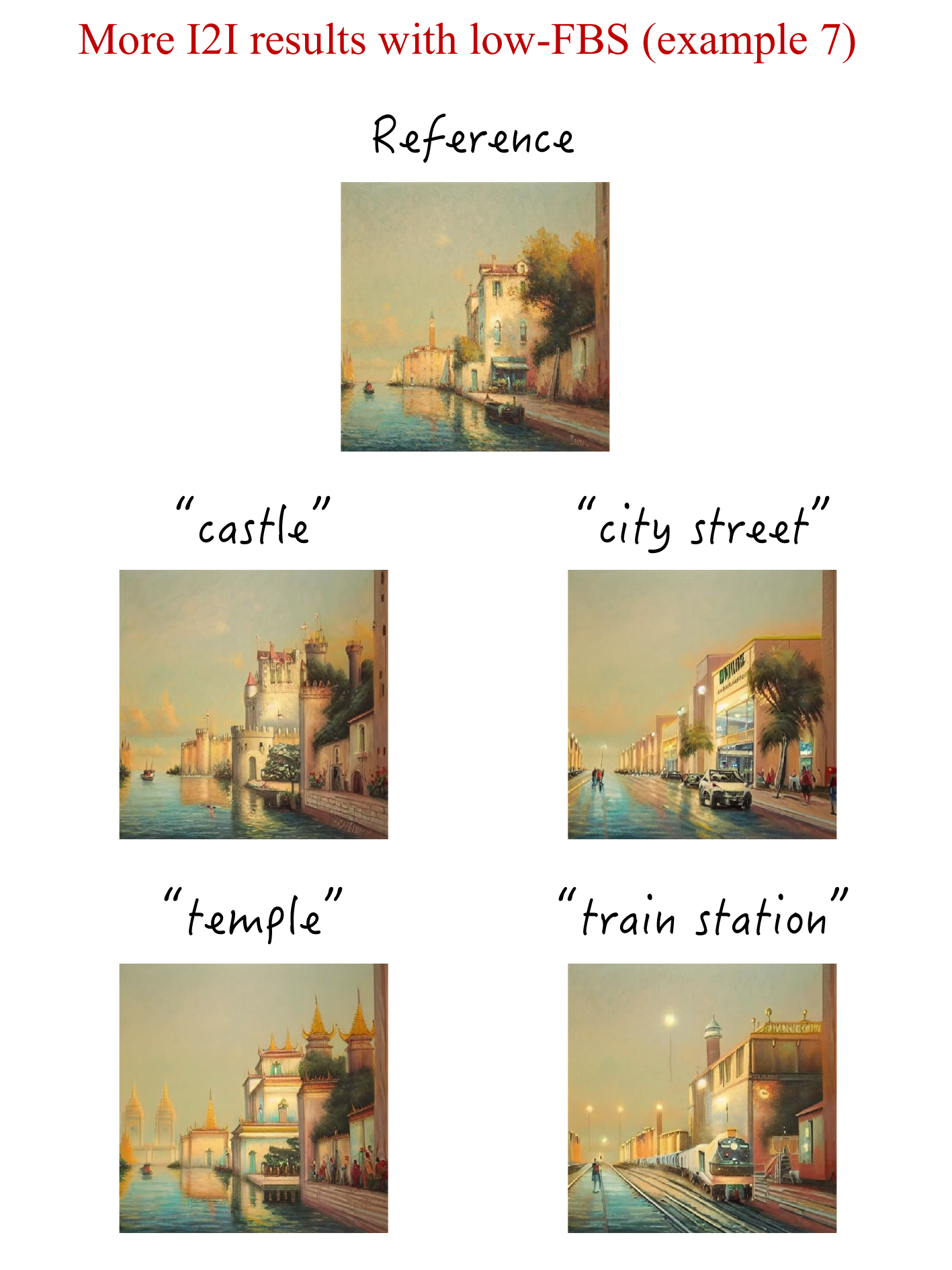}
    \caption{More text-driven I2I results of our method with low-FBS for image appearance and layout control.}
    \label{fig:low_FBS_7}
\end{figure*}

\begin{figure*}[htbp]
    \centering
    \includegraphics[width=\textwidth]{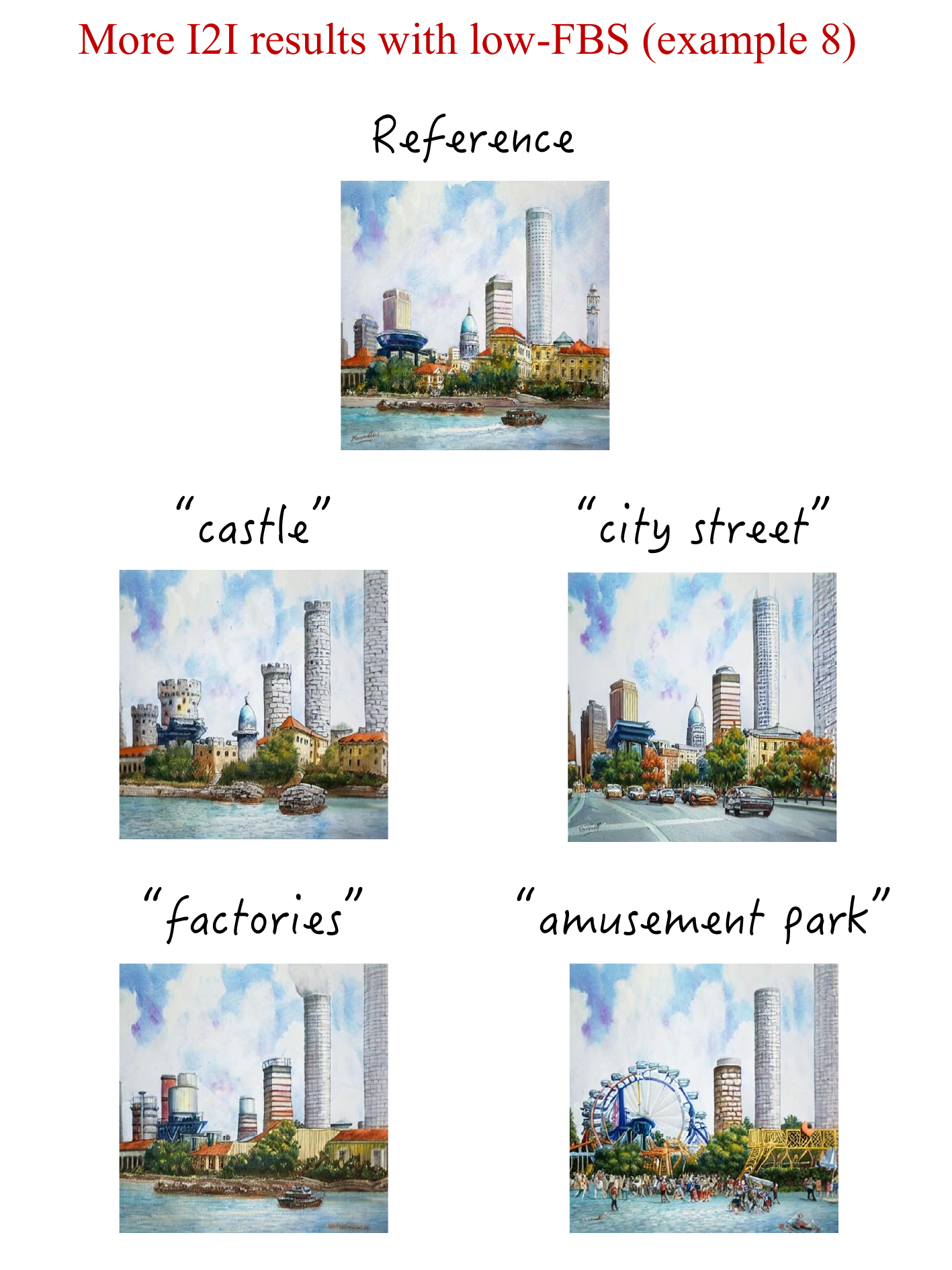}
    \caption{More text-driven I2I results of our method with low-FBS for image appearance and layout control.}
    \label{fig:low_FBS_8}
\end{figure*}

\begin{figure*}[htbp]
    \centering
    \includegraphics[width=\textwidth]{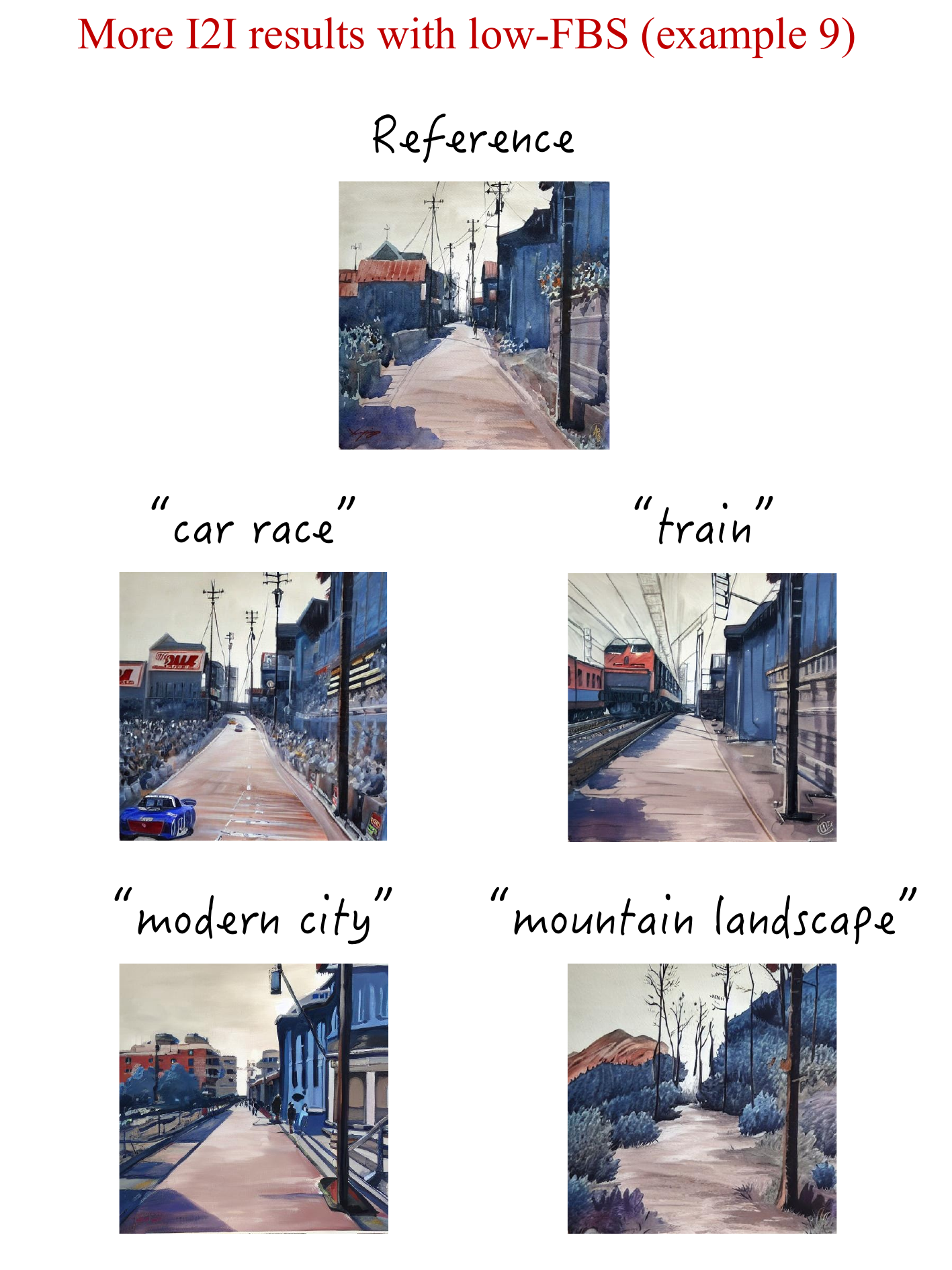}
    \caption{More text-driven I2I results of our method with low-FBS for image appearance and layout control.}
    \label{fig:low_FBS_9}
\end{figure*}

\begin{figure*}[htbp]
    \centering
    \includegraphics[width=\textwidth]{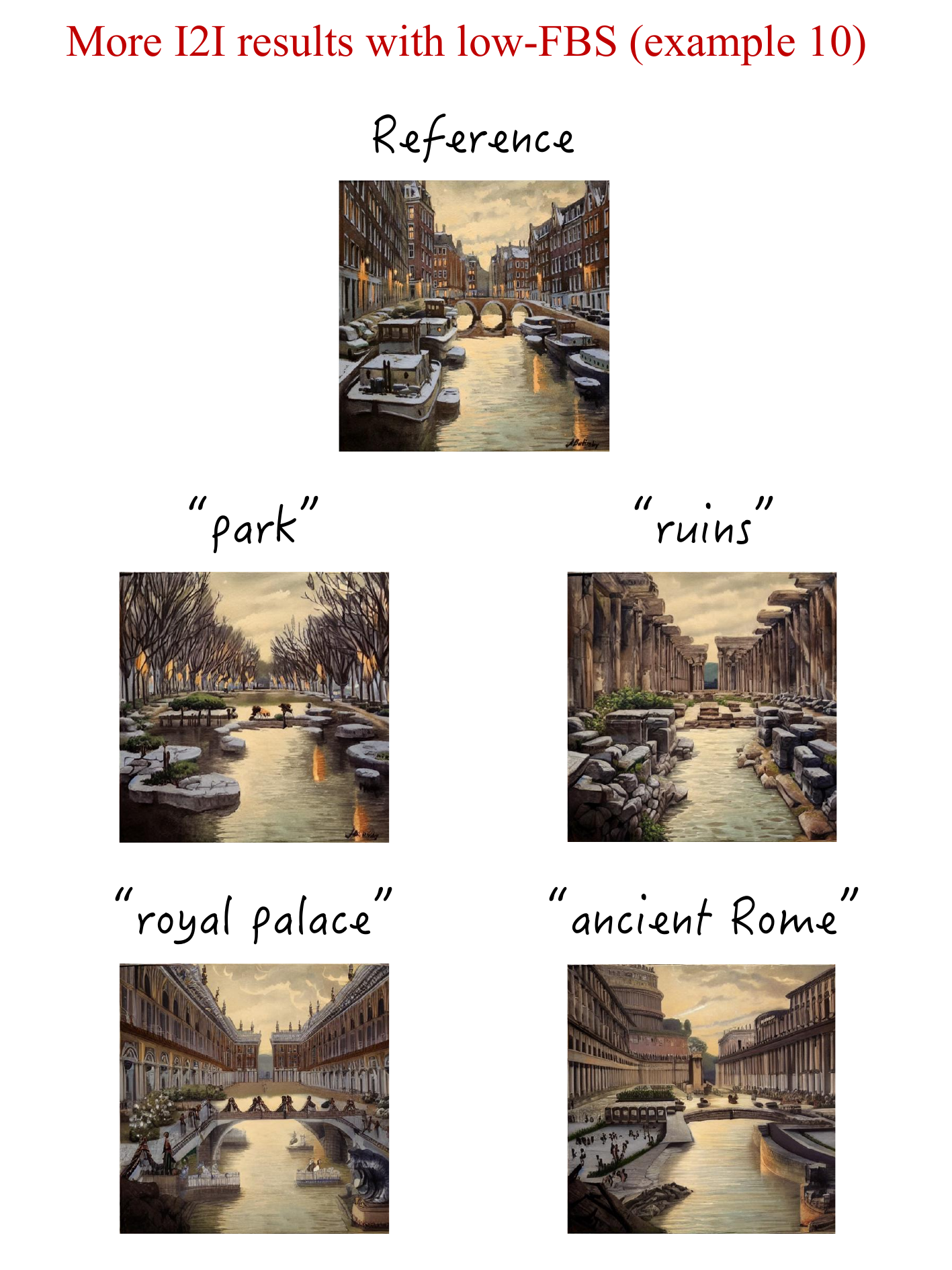}
    \caption{More text-driven I2I results of our method with low-FBS for image appearance and layout control.}
    \label{fig:low_FBS_10}
\end{figure*}

\begin{figure*}[htbp]
    \centering
    \includegraphics[width=\textwidth]{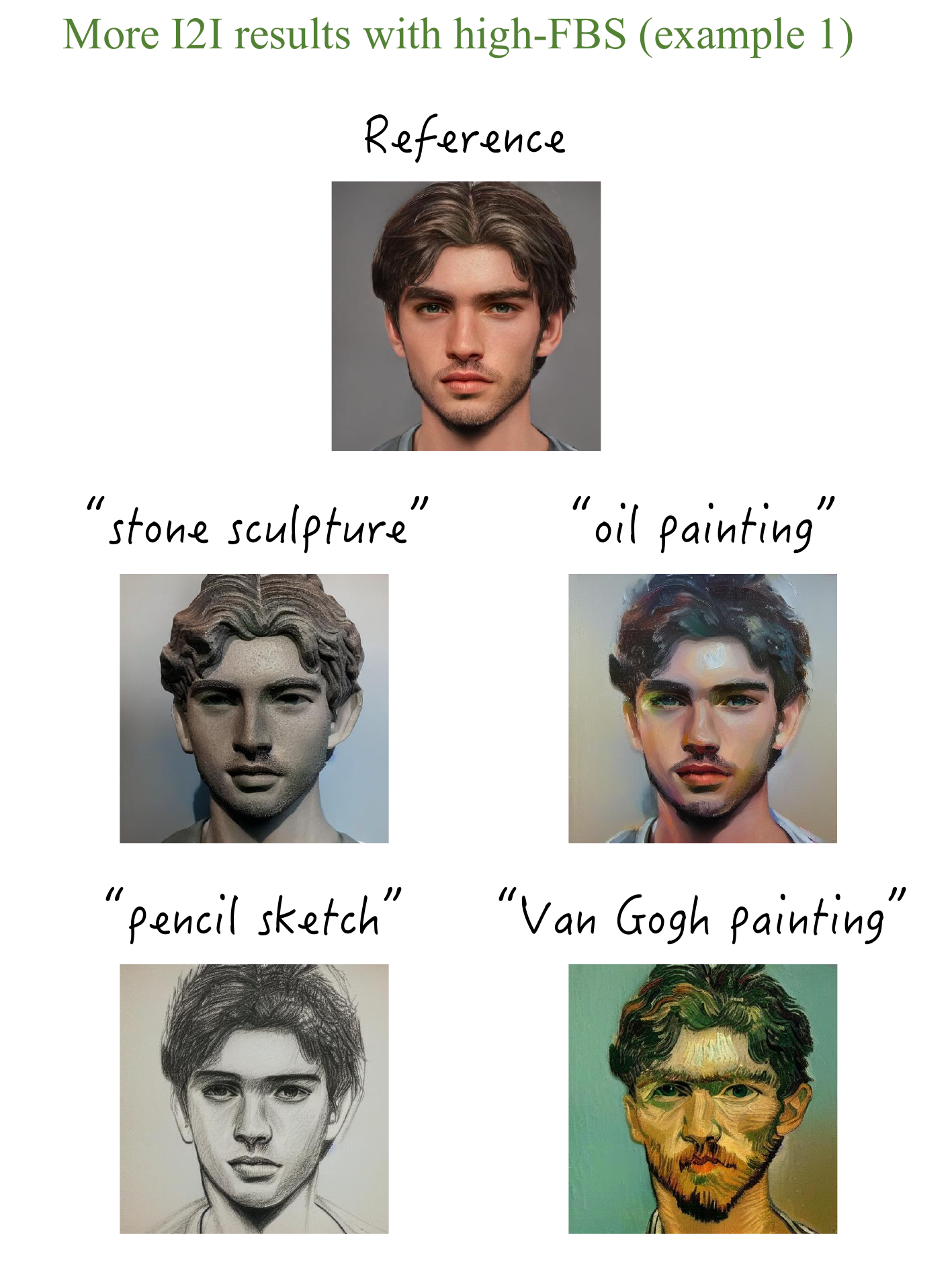}
    \caption{More text-driven I2I results of our method with high-FBS for image contour control.}
    \label{fig:high_FBS_1}
\end{figure*}

\begin{figure*}[htbp]
    \centering
    \includegraphics[width=\textwidth]{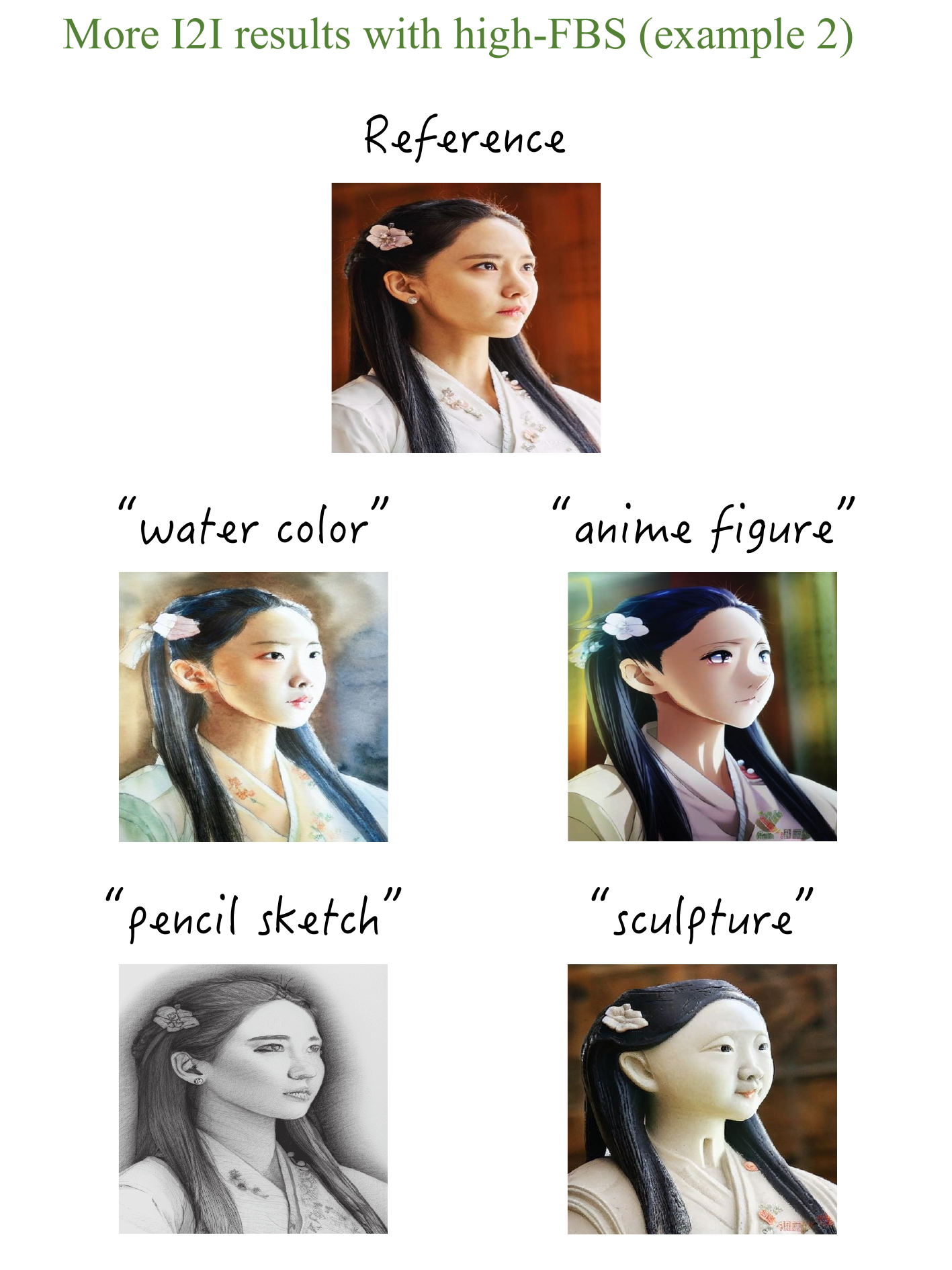}
    \caption{More text-driven I2I results of our method with high-FBS for image contour control.}
    \label{fig:high_FBS_2}
\end{figure*}

\begin{figure*}[htbp]
    \centering
    \includegraphics[width=\textwidth]{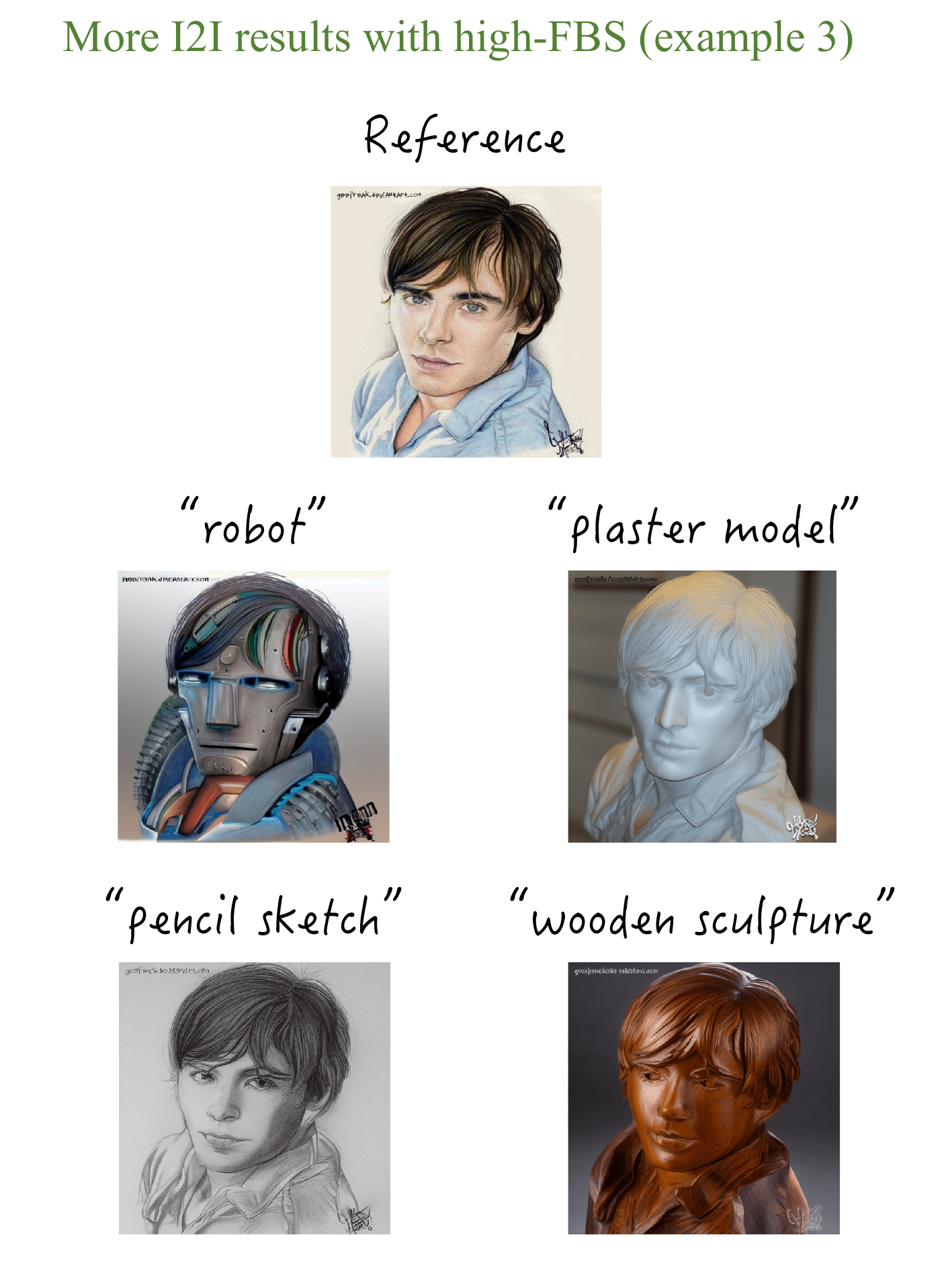}
    \caption{More text-driven I2I results of our method with high-FBS for image contour control.}
    \label{fig:high_FBS_3}
\end{figure*}

\begin{figure*}[htbp]
    \centering
    \includegraphics[width=\textwidth]{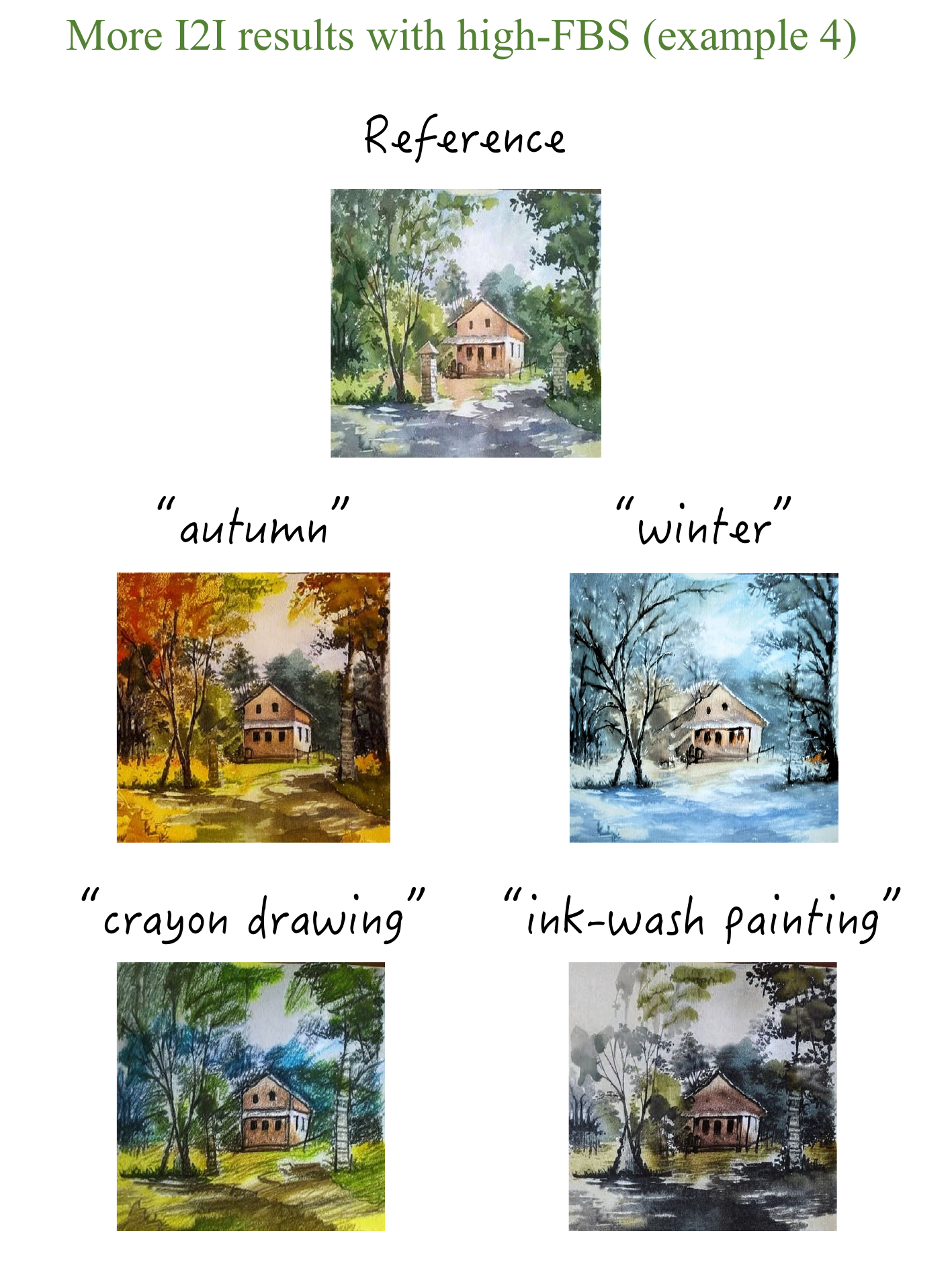}
    \caption{More text-driven I2I results of our method with high-FBS for image contour control.}
    \label{fig:high_FBS_4}
\end{figure*}

\begin{figure*}[htbp]
    \centering
    \includegraphics[width=\textwidth]{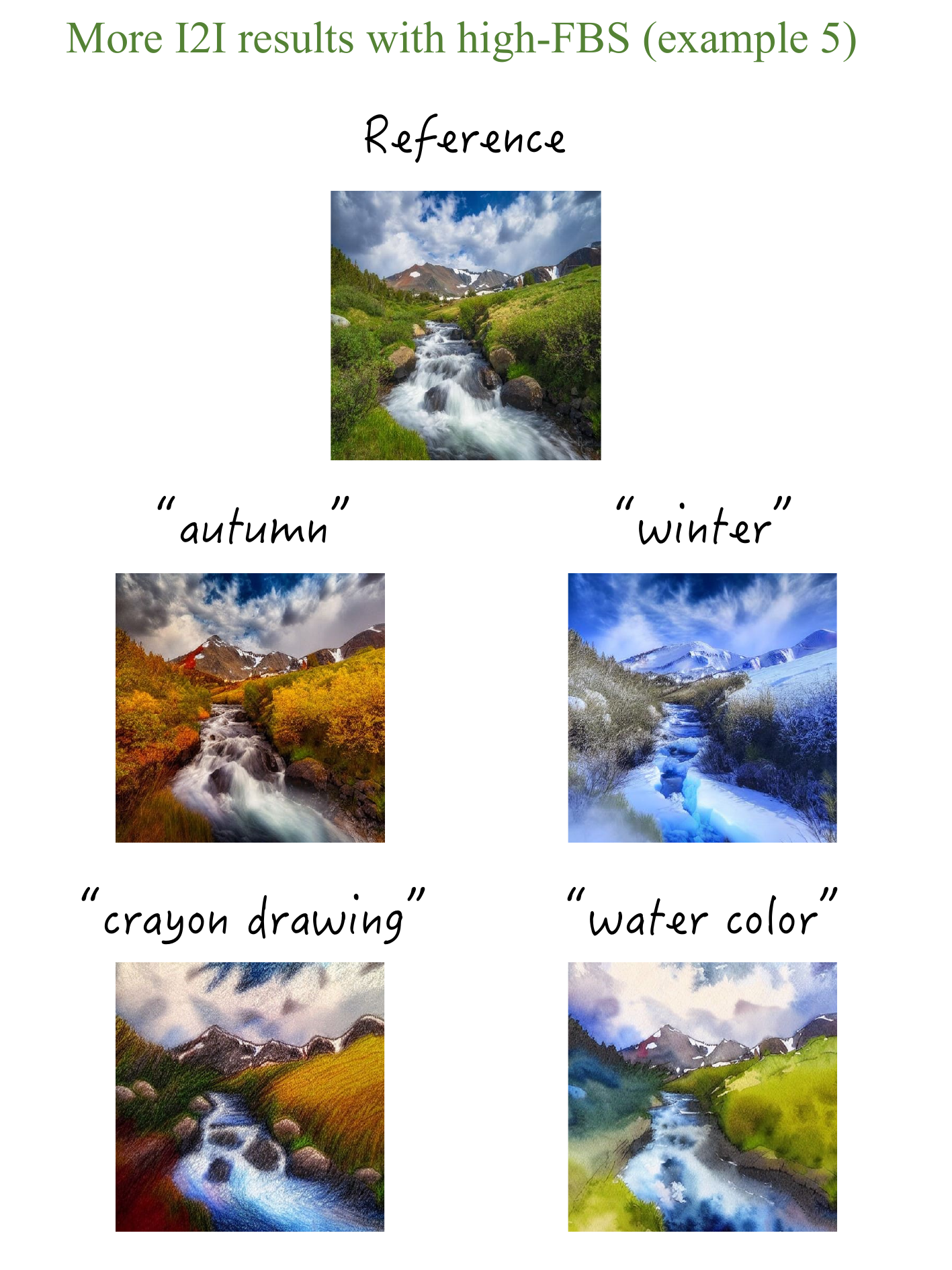}
    \caption{More text-driven I2I results of our method with high-FBS for image contour control.}
    \label{fig:high_FBS_5}
\end{figure*}

\begin{figure*}[htbp]
    \centering
    \includegraphics[width=\textwidth]{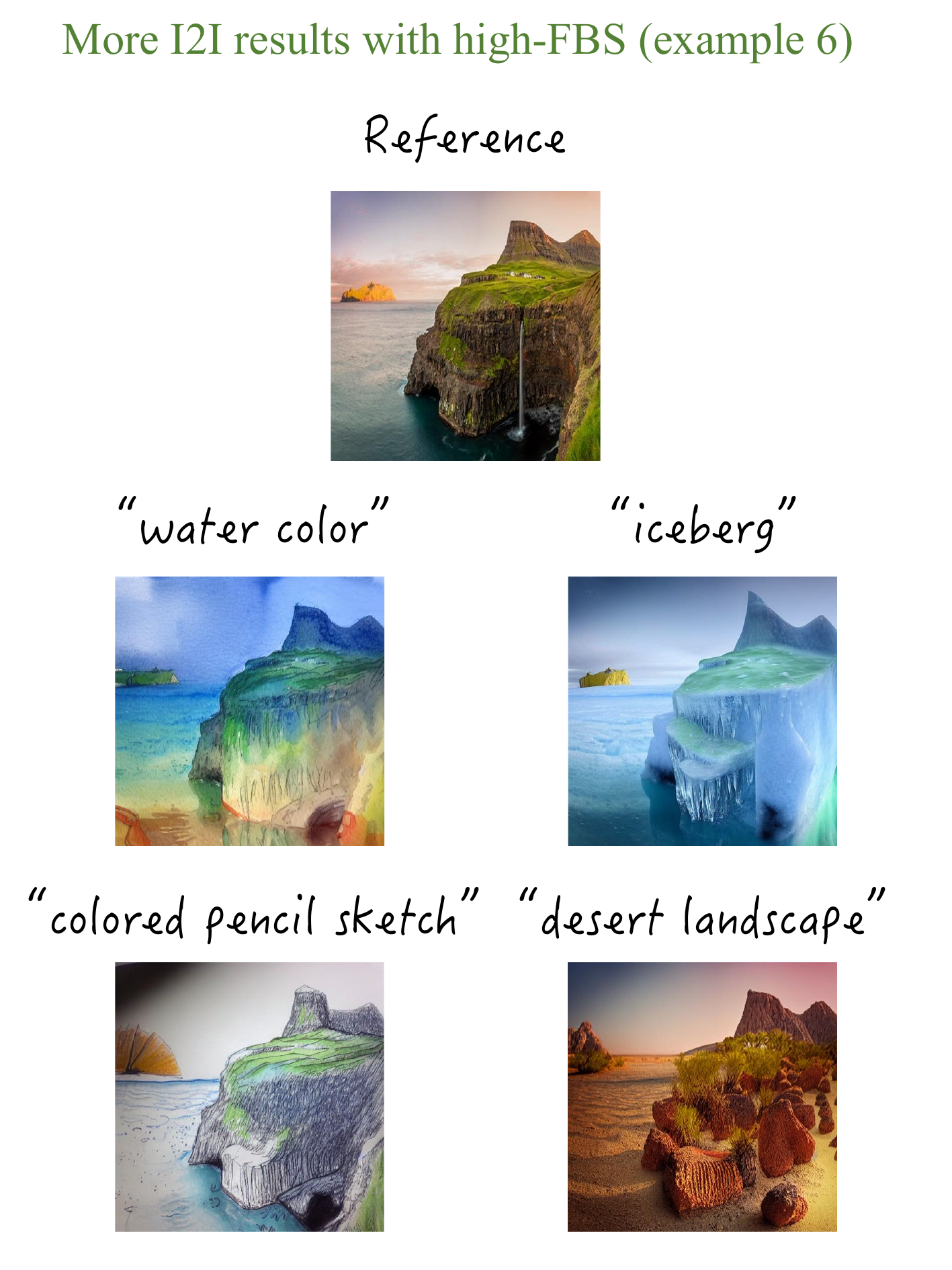}
    \caption{More text-driven I2I results of our method with high-FBS for image contour control.}
    \label{fig:high_FBS_6}
\end{figure*}

\begin{figure*}[htbp]
    \centering
    \includegraphics[width=\textwidth]{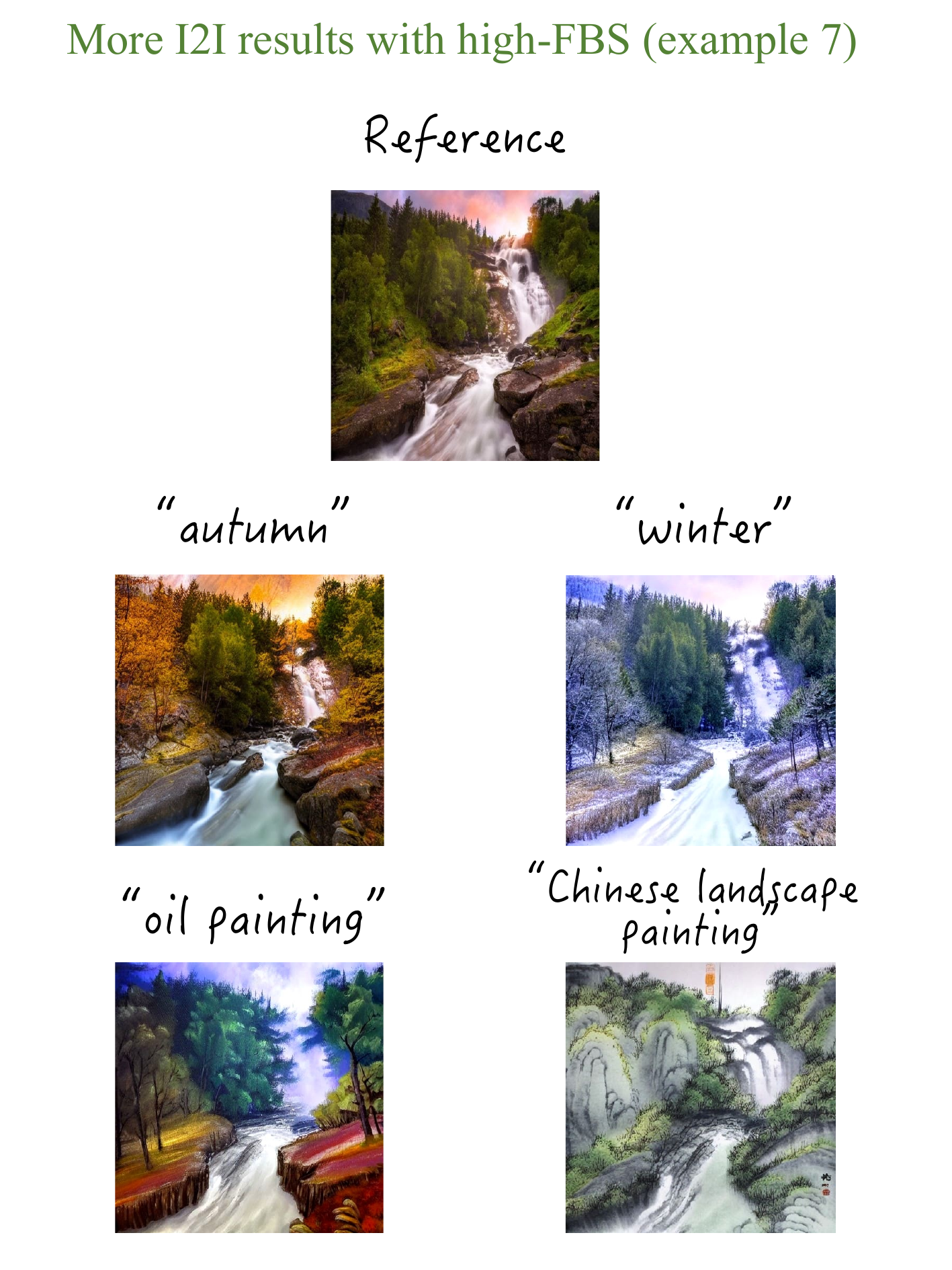}
    \caption{More text-driven I2I results of our method with high-FBS for image contour control.}
    \label{fig:high_FBS_7}
\end{figure*}

\begin{figure*}[htbp]
    \centering
    \includegraphics[width=\textwidth]{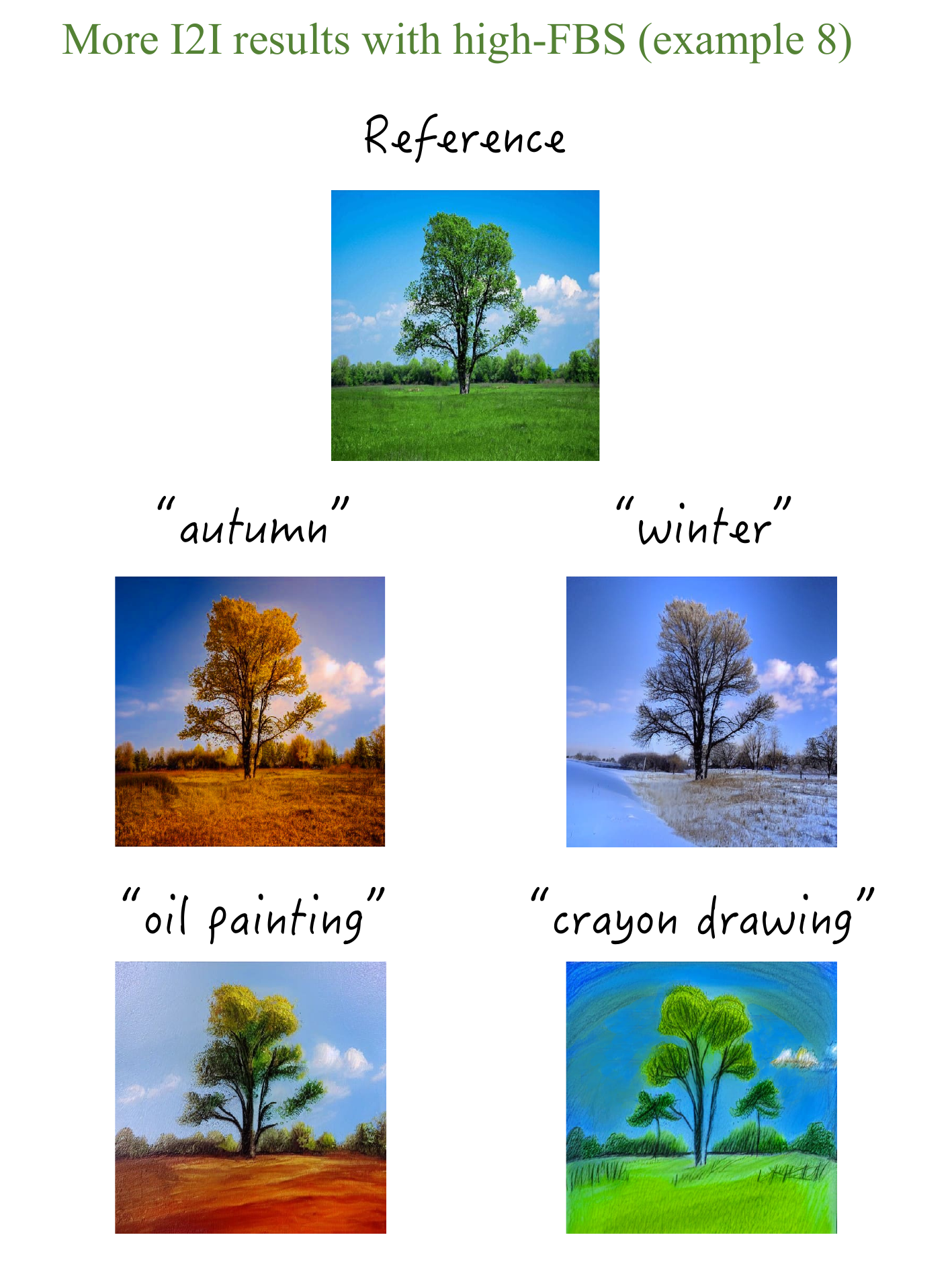}
    \caption{More text-driven I2I results of our method with high-FBS for image contour control.}
    \label{fig:high_FBS_8}
\end{figure*}

\begin{figure*}[htbp]
    \centering
    \includegraphics[width=\textwidth]{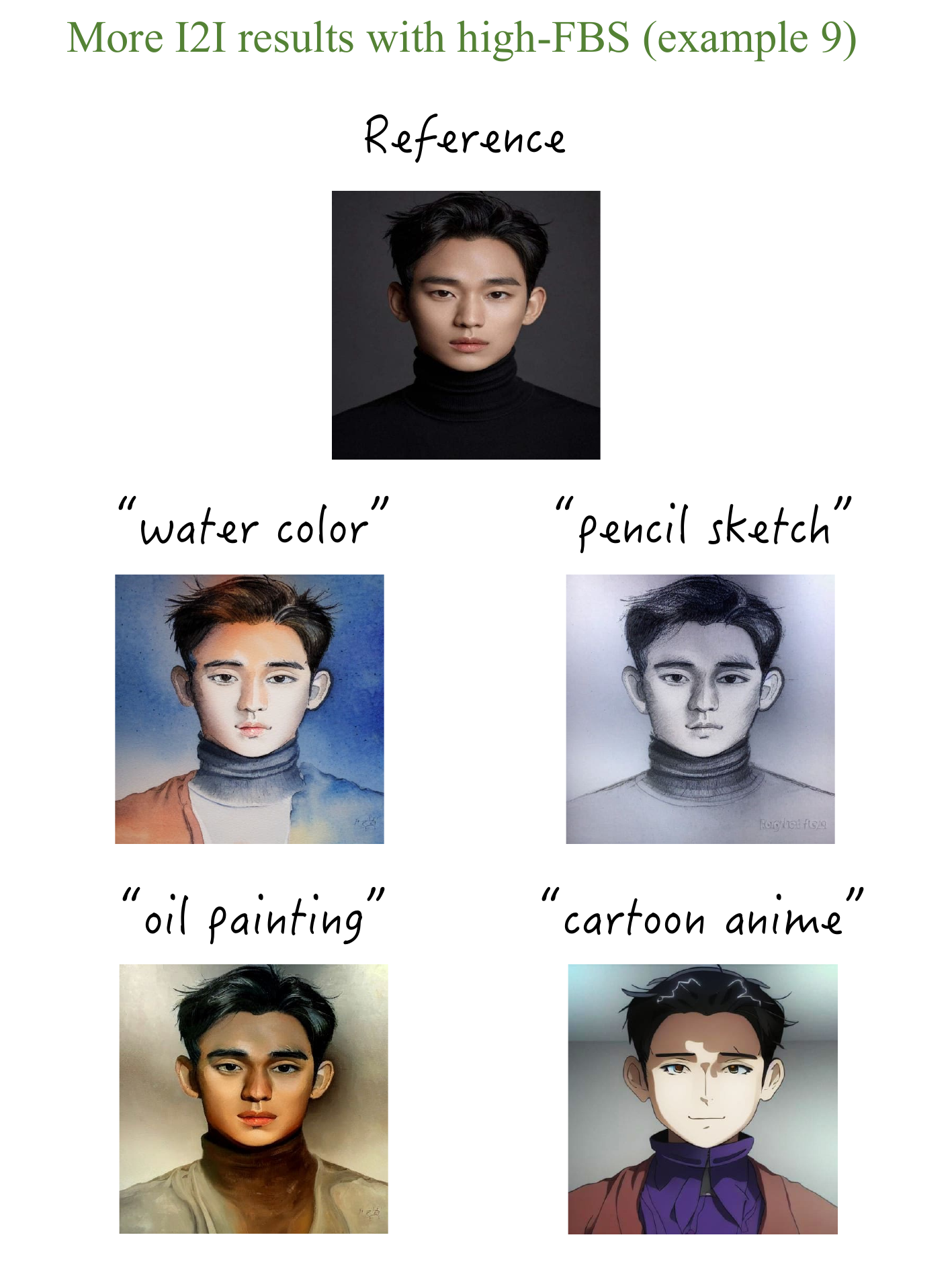}
    \caption{More text-driven I2I results of our method with high-FBS for image contour control.}
    \label{fig:high_FBS_9}
\end{figure*}

\begin{figure*}[htbp]
    \centering
    \includegraphics[width=\textwidth]{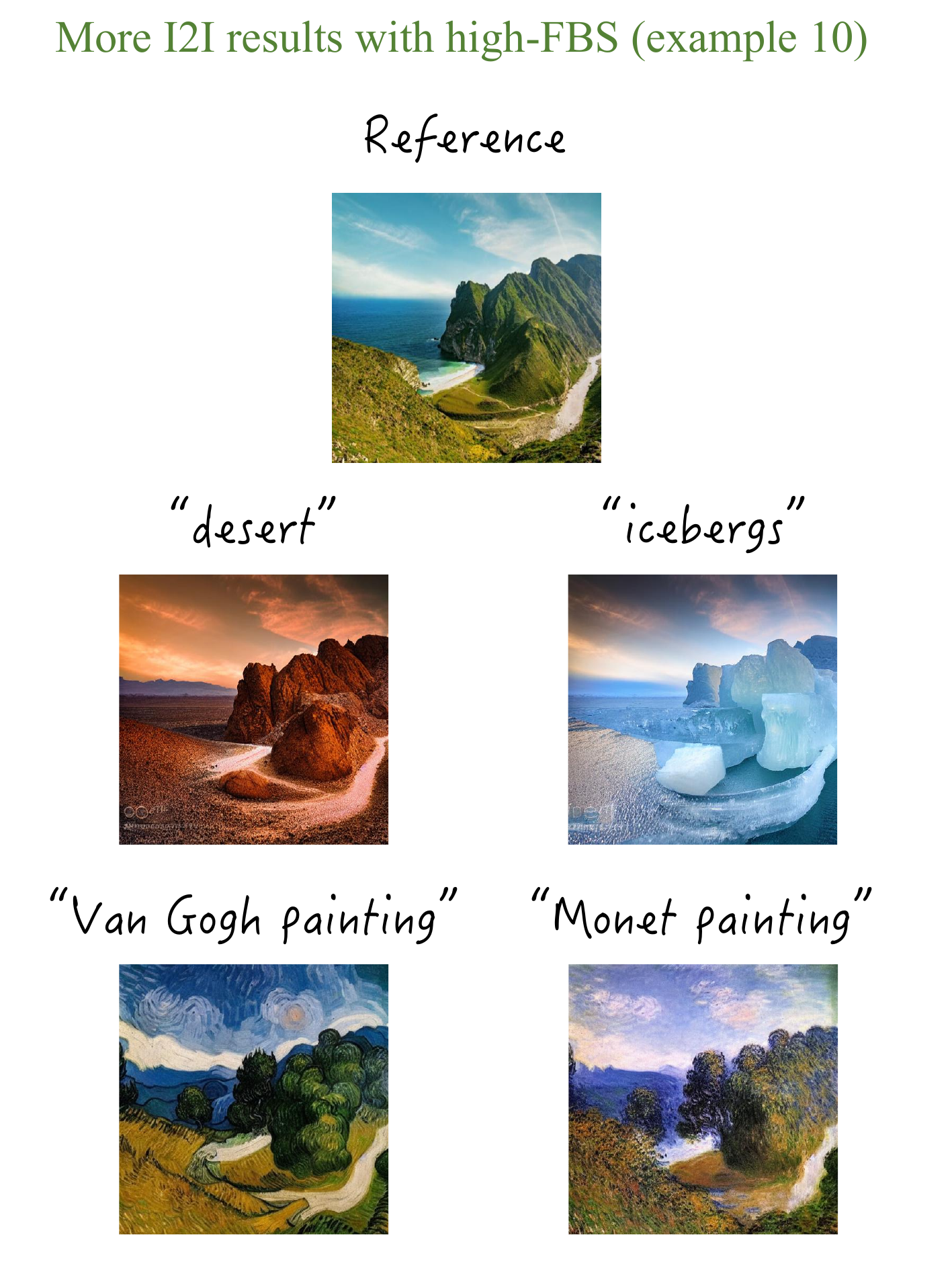}
    \caption{More text-driven I2I results of our method with high-FBS for image contour control.}
    \label{fig:high_FBS_10}
\end{figure*}

\begin{figure*}[htbp]
    \centering
    \includegraphics[width=\textwidth]{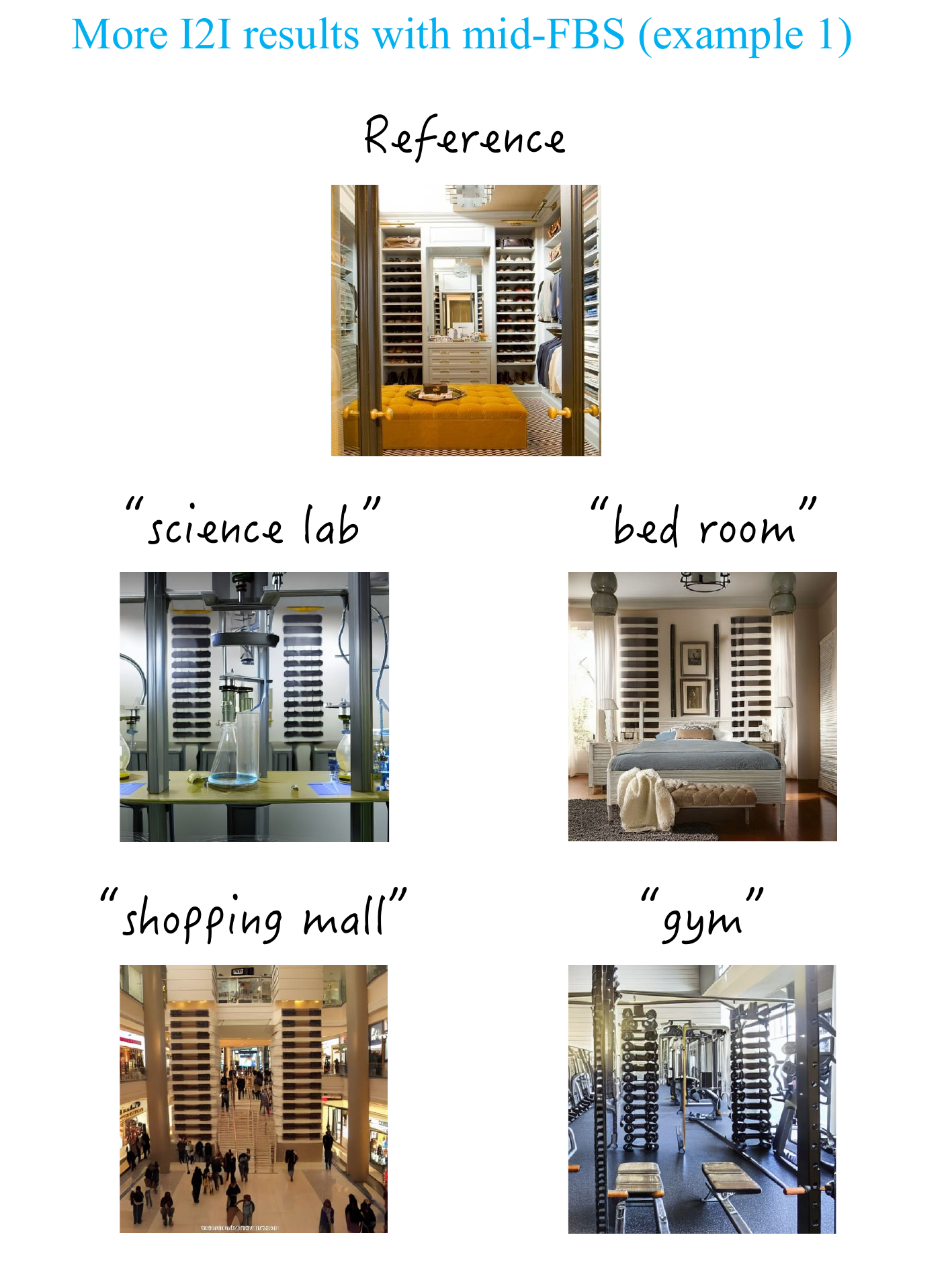}
    \caption{More text-driven I2I results of our method with mid-FBS for image layout control.}
    \label{fig:mid_FBS_1}
\end{figure*}

\begin{figure*}[htbp]
    \centering
    \includegraphics[width=\textwidth]{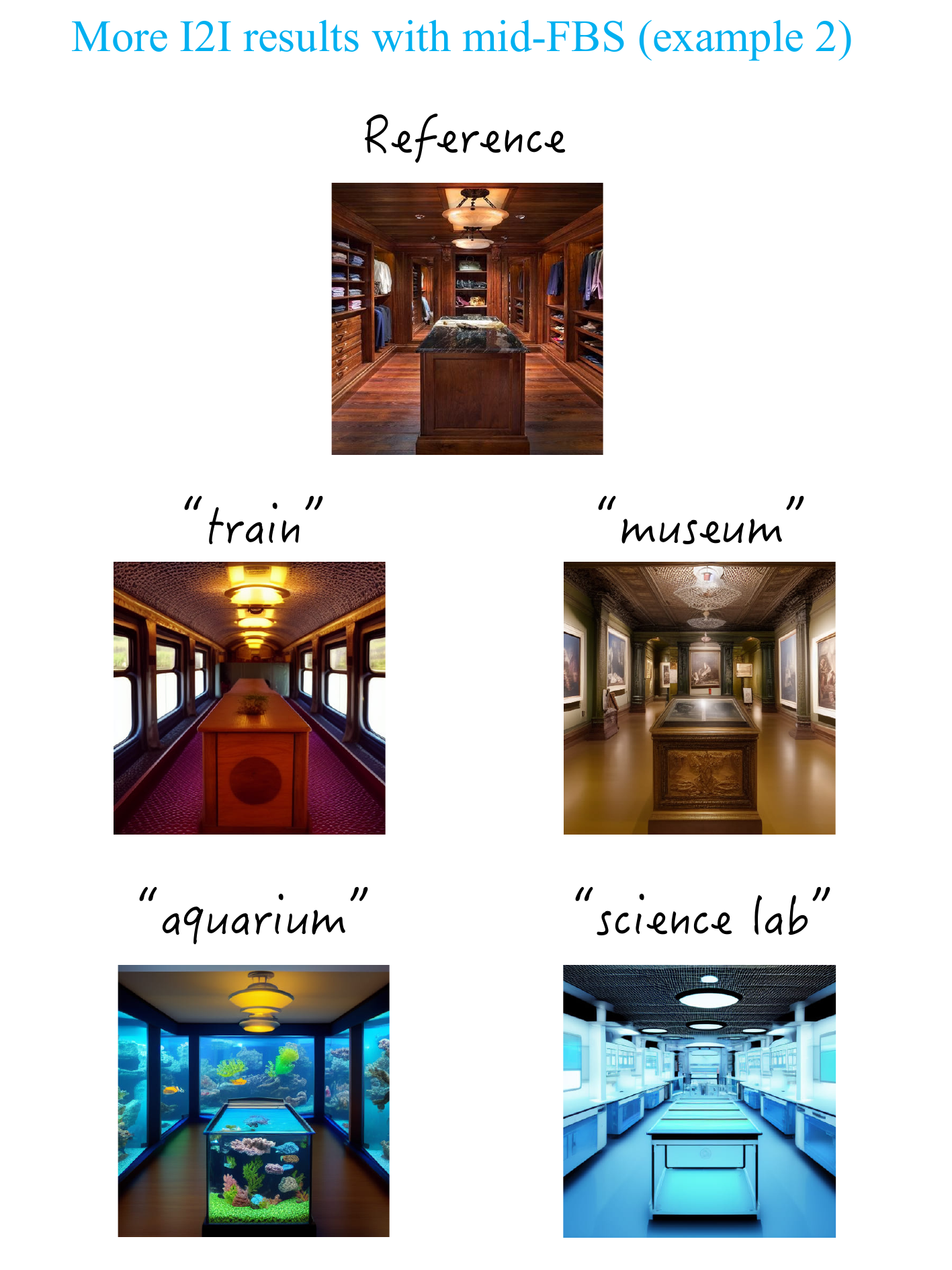}
    \caption{More text-driven I2I results of our method with mid-FBS for image layout control.}
    \label{fig:mid_FBS_2}
\end{figure*}

\begin{figure*}[htbp]
    \centering
    \includegraphics[width=\textwidth]{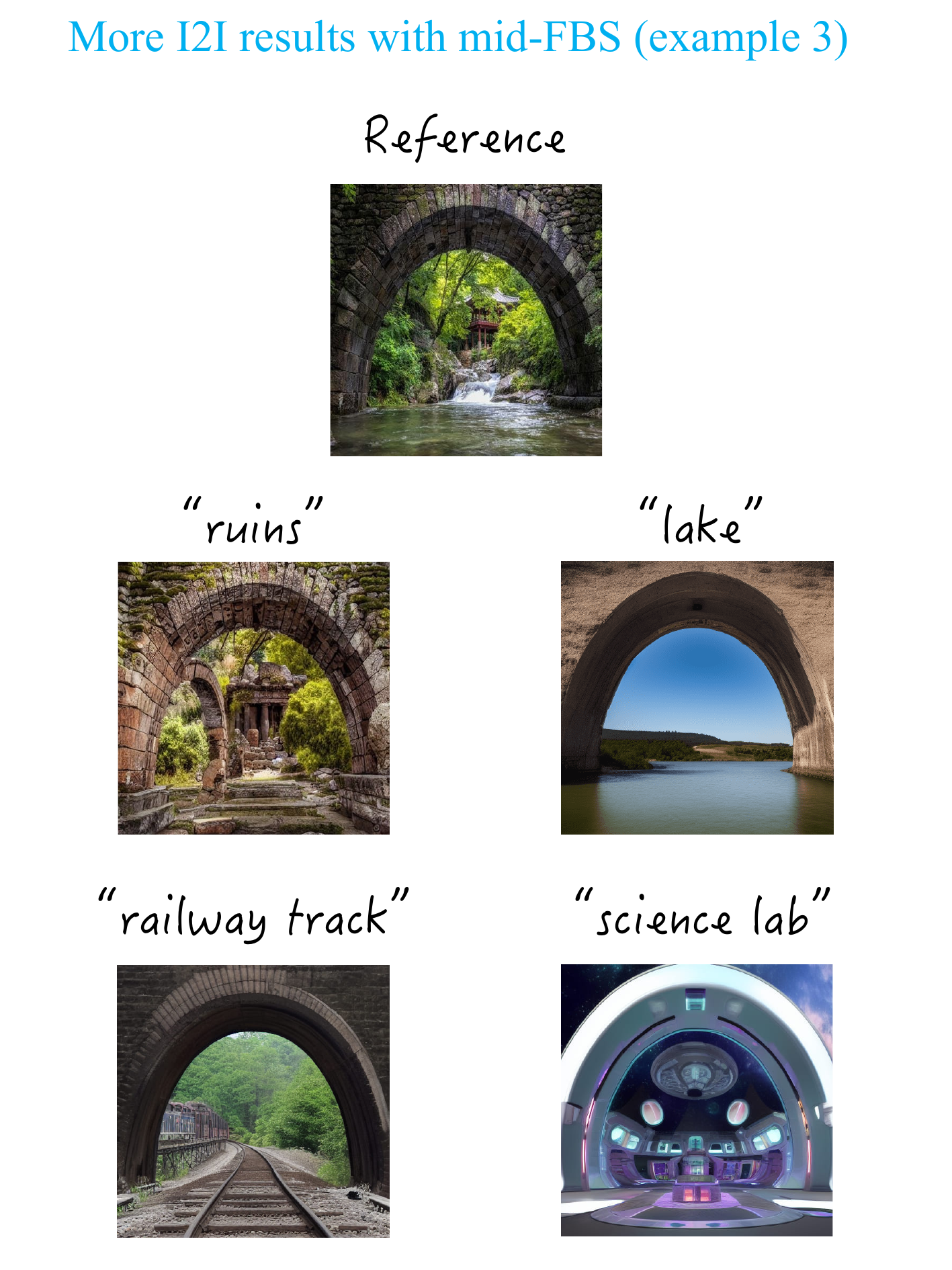}
    \caption{More text-driven I2I results of our method with mid-FBS for image layout control.}
    \label{fig:mid_FBS_3}
\end{figure*}

\begin{figure*}[htbp]
    \centering
    \includegraphics[width=\textwidth]{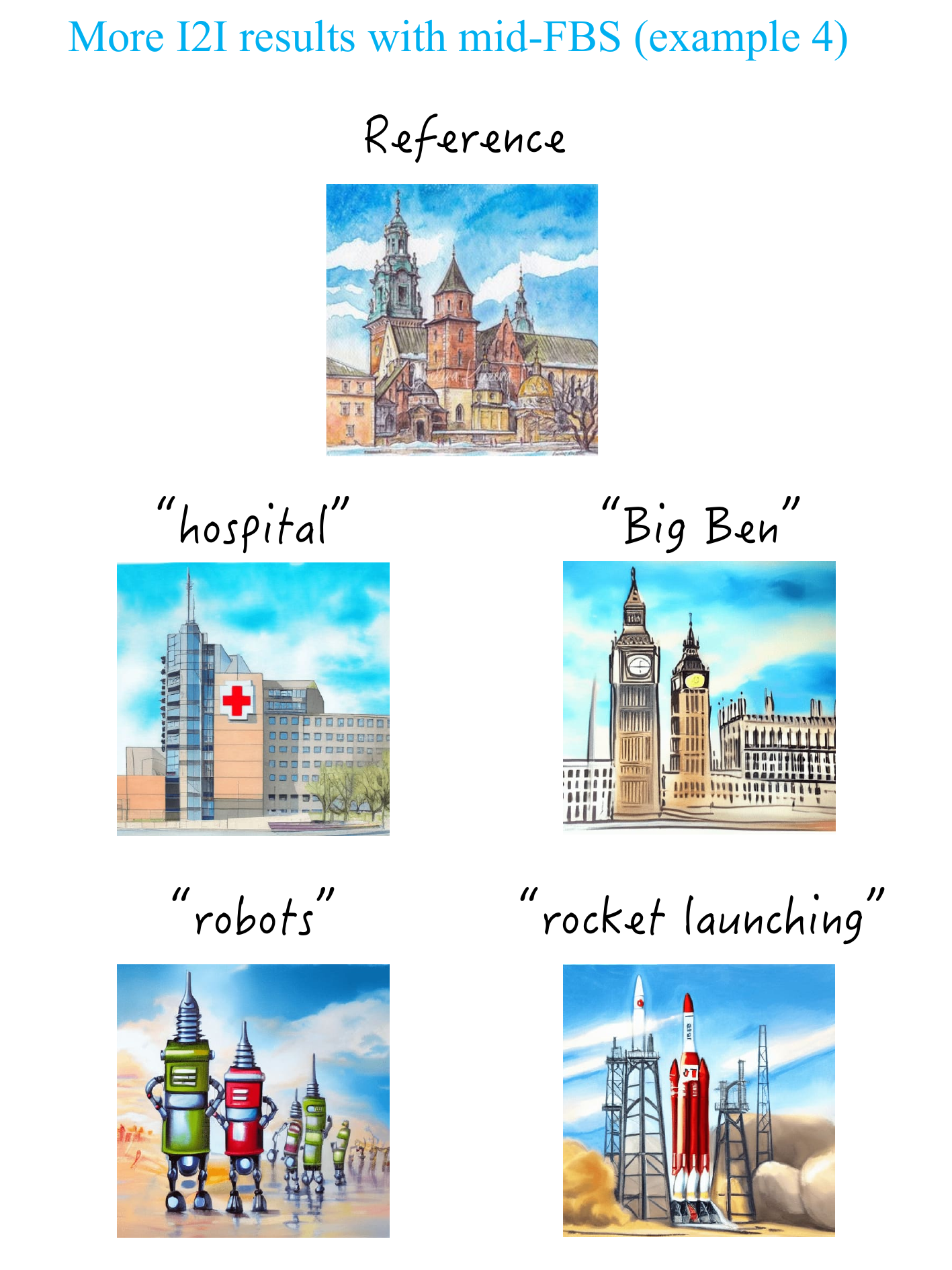}
    \caption{More text-driven I2I results of our method with mid-FBS for image layout control.}
    \label{fig:mid_FBS_4}
\end{figure*}

\begin{figure*}[htbp]
    \centering
    \includegraphics[width=\textwidth]{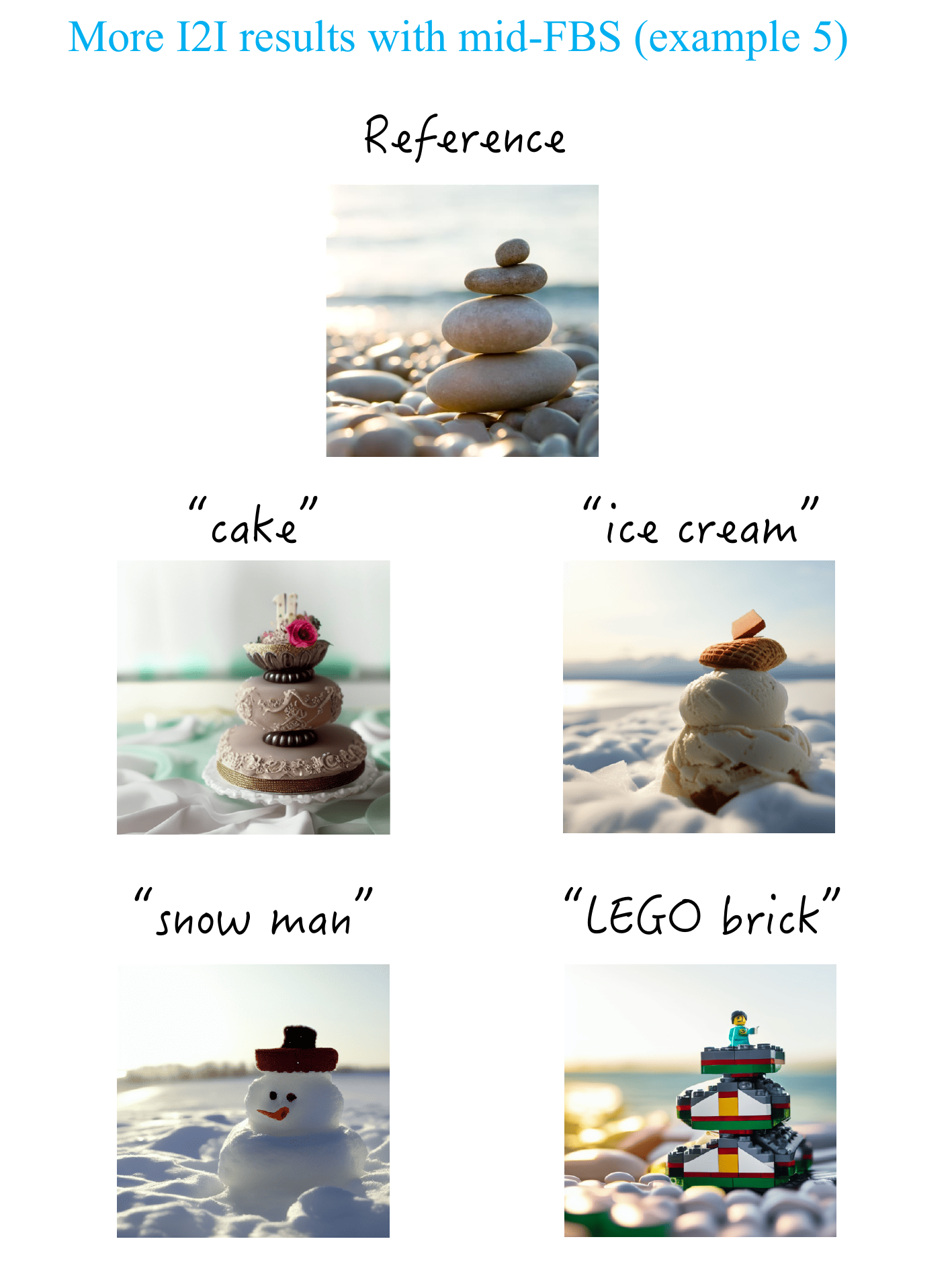}
    \caption{More text-driven I2I results of our method with mid-FBS for image layout control.}
    \label{fig:mid_FBS_5}
\end{figure*}

\bibliographystyle{IEEEtran}
\bibliography{ref}

@String{Computer = "{IEEE} Computer" }

@String{Springer = "Springer-Verlag" }

@inproceedings{radford2021learning,
  title={Learning transferable visual models from natural language supervision},
  author={Radford, Alec and Kim, Jong Wook and Hallacy, Chris and Ramesh, Aditya and Goh, Gabriel and Agarwal, Sandhini and Sastry, Girish and Askell, Amanda and Mishkin, Pamela and Clark, Jack and others},
  booktitle={International Conference on Machine Learning},
  pages={8748--8763},
  year={2021},
  organization={PMLR}
}

@article{ho2020denoising,
  title={Denoising diffusion probabilistic models},
  author={Ho, Jonathan and Jain, Ajay and Abbeel, Pieter},
  journal={Advances in Neural Information Processing Systems},
  volume={33},
  pages={6840--6851},
  year={2020}
}

@inproceedings{crowson2022vqgan,
  title={Vqgan-clip: Open domain image generation and editing with natural language guidance},
  author={Crowson, Katherine and Biderman, Stella and Kornis, Daniel and Stander, Dashiell and Hallahan, Eric and Castricato, Louis and Raff, Edward},
  booktitle={European Conference on Computer Vision},
  pages={88--105},
  year={2022},
  organization={Springer}
}

@inproceedings{esser2021taming,
  title={Taming transformers for high-resolution image synthesis},
  author={Esser, Patrick and Rombach, Robin and Ommer, Bjorn},
  booktitle={Proceedings of the IEEE/CVF Conference on Computer Vision and Pattern Recognition},
  pages={12873--12883},
  year={2021}
}

@inproceedings{kim2022diffusionclip,
  title={Diffusionclip: Text-guided diffusion models for robust image manipulation},
  author={Kim, Gwanghyun and Kwon, Taesung and Ye, Jong Chul},
  booktitle={Proceedings of the IEEE/CVF Conference on Computer Vision and Pattern Recognition},
  pages={2426--2435},
  year={2022}
}

@inproceedings{kwon2022diffusion,
  title={Diffusion-based Image Translation using disentangled style and content representation},
  author={Kwon, Gihyun and Ye, Jong Chul},
  booktitle={The Eleventh International Conference on Learning Representations},
  year={2022}
}

@inproceedings{tumanyan2022splicing,
  title={Splicing vit features for semantic appearance transfer},
  author={Tumanyan, Narek and Bar-Tal, Omer and Bagon, Shai and Dekel, Tali},
  booktitle={Proceedings of the IEEE/CVF Conference on Computer Vision and Pattern Recognition},
  pages={10748--10757},
  year={2022}
}

@article{chung2022improving,
  title={Improving diffusion models for inverse problems using manifold constraints},
  author={Chung, Hyungjin and Sim, Byeongsu and Ryu, Dohoon and Ye, Jong Chul},
  journal={Advances in Neural Information Processing Systems},
  volume={35},
  pages={25683--25696},
  year={2022}
}

@inproceedings{brooks2023instructpix2pix,
  title={Instructpix2pix: Learning to follow image editing instructions},
  author={Brooks, Tim and Holynski, Aleksander and Efros, Alexei A},
  booktitle={Proceedings of the IEEE/CVF Conference on Computer Vision and Pattern Recognition},
  pages={18392--18402},
  year={2023}
}

@article{brown2020language,
  title={Language models are few-shot learners},
  author={Brown, Tom and Mann, Benjamin and Ryder, Nick and Subbiah, Melanie and Kaplan, Jared D and Dhariwal, Prafulla and Neelakantan, Arvind and Shyam, Pranav and Sastry, Girish and Askell, Amanda and others},
  journal={Advances in Neural Information Processing Systems},
  volume={33},
  pages={1877--1901},
  year={2020}
}

@inproceedings{rombach2022high,
  title={High-resolution image synthesis with latent diffusion models},
  author={Rombach, Robin and Blattmann, Andreas and Lorenz, Dominik and Esser, Patrick and Ommer, Bj{\"o}rn},
  booktitle={Proceedings of the IEEE/CVF Conference on Computer Vision and Pattern Recognition},
  pages={10684--10695},
  year={2022}
}

@article{sun2023design,
  title={Design Booster: A Text-Guided Diffusion Model for Image Translation with Spatial Layout Preservation},
  author={Sun, Shiqi and Fang, Shancheng and He, Qian and Liu, Wei},
  journal={arXiv preprint arXiv:2302.02284},
  year={2023}
}

@inproceedings{zhang2023sine,
  title={Sine: Single image editing with text-to-image diffusion models},
  author={Zhang, Zhixing and Han, Ligong and Ghosh, Arnab and Metaxas, Dimitris N and Ren, Jian},
  booktitle={Proceedings of the IEEE/CVF Conference on Computer Vision and Pattern Recognition},
  pages={6027--6037},
  year={2023}
}

@inproceedings{kawar2023imagic,
  title={Imagic: Text-based real image editing with diffusion models},
  author={Kawar, Bahjat and Zada, Shiran and Lang, Oran and Tov, Omer and Chang, Huiwen and Dekel, Tali and Mosseri, Inbar and Irani, Michal},
  booktitle={Proceedings of the IEEE/CVF Conference on Computer Vision and Pattern Recognition},
  pages={6007--6017},
  year={2023}
}

@article{ho2022classifier,
  title={Classifier-free diffusion guidance},
  author={Ho, Jonathan and Salimans, Tim},
  journal={arXiv preprint arXiv:2207.12598},
  year={2022}
}

@inproceedings{mokady2023null,
  title={Null-text inversion for editing real images using guided diffusion models},
  author={Mokady, Ron and Hertz, Amir and Aberman, Kfir and Pritch, Yael and Cohen-Or, Daniel},
  booktitle={Proceedings of the IEEE/CVF Conference on Computer Vision and Pattern Recognition},
  pages={6038--6047},
  year={2023}
}

@article{vaswani2017attention,
  title={Attention is all you need},
  author={Vaswani, Ashish and Shazeer, Noam and Parmar, Niki and Uszkoreit, Jakob and Jones, Llion and Gomez, Aidan N and Kaiser, {\L}ukasz and Polosukhin, Illia},
  journal={Advances in neural information processing systems},
  volume={30},
  year={2017}
}

@inproceedings{gao2024frequency,
  title={Frequency-Controlled Diffusion Model for Versatile Text-Guided Image-to-Image Translation},
  author={Gao, Xiang and Xu, Zhengbo and Zhao, Junhan and Liu, Jiaying},
  booktitle={Proceedings of the AAAI Conference on Artificial Intelligence},
  volume={38},
  number={3},
  pages={1824--1832},
  year={2024}
}

@inproceedings{parmar2023zero,
  title={Zero-shot image-to-image translation},
  author={Parmar, Gaurav and Kumar Singh, Krishna and Zhang, Richard and Li, Yijun and Lu, Jingwan and Zhu, Jun-Yan},
  booktitle={ACM SIGGRAPH 2023 Conference Proceedings},
  pages={1--11},
  year={2023}
}

@inproceedings{dong2023prompt,
  title={Prompt tuning inversion for text-driven image editing using diffusion models},
  author={Dong, Wenkai and Xue, Song and Duan, Xiaoyue and Han, Shumin},
  booktitle={Proceedings of the IEEE/CVF International Conference on Computer Vision},
  pages={7430--7440},
  year={2023}
}

@article{li2023stylediffusion,
  title={Stylediffusion: Prompt-embedding inversion for text-based editing},
  author={Li, Senmao and van de Weijer, Joost and Hu, Taihang and Khan, Fahad Shahbaz and Hou, Qibin and Wang, Yaxing and Yang, Jian},
  journal={arXiv preprint arXiv:2303.15649},
  year={2023}
}

@article{hertz2022prompt,
  title={Prompt-to-prompt image editing with cross attention control},
  author={Hertz, Amir and Mokady, Ron and Tenenbaum, Jay and Aberman, Kfir and Pritch, Yael and Cohen-Or, Daniel},
  journal={arXiv preprint arXiv:2208.01626},
  year={2022}
}

@inproceedings{tumanyan2023plug,
  title={Plug-and-play diffusion features for text-driven image-to-image translation},
  author={Tumanyan, Narek and Geyer, Michal and Bagon, Shai and Dekel, Tali},
  booktitle={Proceedings of the IEEE/CVF Conference on Computer Vision and Pattern Recognition},
  pages={1921--1930},
  year={2023}
}

@article{dhariwal2021diffusion,
  title={Diffusion models beat gans on image synthesis},
  author={Dhariwal, Prafulla and Nichol, Alexander},
  journal={Advances in Neural Information Processing Systems},
  volume={34},
  pages={8780--8794},
  year={2021}
}

@article{song2020denoising,
  title={Denoising diffusion implicit models},
  author={Song, Jiaming and Meng, Chenlin and Ermon, Stefano},
  journal={arXiv preprint arXiv:2010.02502},
  year={2020}
}

@article{lu2022dpm,
  title={Dpm-solver: A fast ode solver for diffusion probabilistic model sampling in around 10 steps},
  author={Lu, Cheng and Zhou, Yuhao and Bao, Fan and Chen, Jianfei and Li, Chongxuan and Zhu, Jun},
  journal={Advances in Neural Information Processing Systems},
  volume={35},
  pages={5775--5787},
  year={2022}
}

@inproceedings{saharia2022palette,
  title={Palette: Image-to-image diffusion models},
  author={Saharia, Chitwan and Chan, William and Chang, Huiwen and Lee, Chris and Ho, Jonathan and Salimans, Tim and Fleet, David and Norouzi, Mohammad},
  booktitle={ACM SIGGRAPH 2022 conference proceedings},
  pages={1--10},
  year={2022}
}

@inproceedings{zhang2018unreasonable,
  title={The unreasonable effectiveness of deep features as a perceptual metric},
  author={Zhang, Richard and Isola, Phillip and Efros, Alexei A and Shechtman, Eli and Wang, Oliver},
  booktitle={Proceedings of the IEEE conference on computer vision and pattern recognition},
  pages={586--595},
  year={2018}
}

@inproceedings{huang2017arbitrary,
  title={Arbitrary style transfer in real-time with adaptive instance normalization},
  author={Huang, Xun and Belongie, Serge},
  booktitle={Proceedings of the IEEE international conference on computer vision},
  pages={1501--1510},
  year={2017}
}

@inproceedings{nichol2022glide,
  title={GLIDE: Towards Photorealistic Image Generation and Editing with Text-Guided Diffusion Models},
  author={Nichol, Alexander Quinn and Dhariwal, Prafulla and Ramesh, Aditya and Shyam, Pranav and Mishkin, Pamela and Mcgrew, Bob and Sutskever, Ilya and Chen, Mark},
  booktitle={International Conference on Machine Learning},
  pages={16784--16804},
  year={2022},
  organization={PMLR}
}

@article{ramesh2022hierarchical,
  title={Hierarchical text-conditional image generation with clip latents},
  author={Ramesh, Aditya and Dhariwal, Prafulla and Nichol, Alex and Chu, Casey and Chen, Mark},
  journal={arXiv preprint arXiv:2204.06125},
  volume={1},
  number={2},
  pages={3},
  year={2022}
}

@article{saharia2022photorealistic,
  title={Photorealistic text-to-image diffusion models with deep language understanding},
  author={Saharia, Chitwan and Chan, William and Saxena, Saurabh and Li, Lala and Whang, Jay and Denton, Emily L and Ghasemipour, Kamyar and Gontijo Lopes, Raphael and Karagol Ayan, Burcu and Salimans, Tim and others},
  journal={Advances in Neural Information Processing Systems},
  volume={35},
  pages={36479--36494},
  year={2022}
}

@inproceedings{zhang2023adding,
  title={Adding conditional control to text-to-image diffusion models},
  author={Zhang, Lvmin and Rao, Anyi and Agrawala, Maneesh},
  booktitle={Proceedings of the IEEE/CVF International Conference on Computer Vision},
  pages={3836--3847},
  year={2023}
}

@inproceedings{mou2024t2i,
  title={T2i-adapter: Learning adapters to dig out more controllable ability for text-to-image diffusion models},
  author={Mou, Chong and Wang, Xintao and Xie, Liangbin and Wu, Yanze and Zhang, Jian and Qi, Zhongang and Shan, Ying},
  booktitle={Proceedings of the AAAI Conference on Artificial Intelligence},
  volume={38},
  number={5},
  pages={4296--4304},
  year={2024}
}

@article{podell2023sdxl,
  title={Sdxl: Improving latent diffusion models for high-resolution image synthesis},
  author={Podell, Dustin and English, Zion and Lacey, Kyle and Blattmann, Andreas and Dockhorn, Tim and M{\"u}ller, Jonas and Penna, Joe and Rombach, Robin},
  journal={arXiv preprint arXiv:2307.01952},
  year={2023}
}

@inproceedings{peebles2023scalable,
  title={Scalable diffusion models with transformers},
  author={Peebles, William and Xie, Saining},
  booktitle={Proceedings of the IEEE/CVF International Conference on Computer Vision},
  pages={4195--4205},
  year={2023}
}

@article{saharia2022image,
  title={Image super-resolution via iterative refinement},
  author={Saharia, Chitwan and Ho, Jonathan and Chan, William and Salimans, Tim and Fleet, David J and Norouzi, Mohammad},
  journal={IEEE Transactions on Pattern Analysis and Machine Intelligence},
  volume={45},
  number={4},
  pages={4713--4726},
  year={2022},
  publisher={IEEE}
}

@inproceedings{lugmayrinpainting,
  title={Inpainting using denoising diffusion probabilistic models},
  author={Lugmayr, Andreas and Danelljan, Martin and Romero, Andres and Yu, Fisher and Timofte, Radu and Van Gool, L Repaint},
  booktitle={Proceedings of the IEEE/CVF Conference on Computer Vision and Pattern Recognition},
  pages={11461--11471}
}

@article{liang2024control,
  title={Control Color: Multimodal Diffusion-based Interactive Image Colorization},
  author={Liang, Zhexin and Li, Zhaochen and Zhou, Shangchen and Li, Chongyi and Loy, Chen Change},
  journal={arXiv preprint arXiv:2402.10855},
  year={2024}
}

@article{tan2023diffss,
  title={Diffss: Diffusion model for few-shot semantic segmentation},
  author={Tan, Weimin and Chen, Siyuan and Yan, Bo},
  journal={arXiv preprint arXiv:2307.00773},
  year={2023}
}

@inproceedings{luo2021diffusion,
  title={Diffusion probabilistic models for 3d point cloud generation},
  author={Luo, Shitong and Hu, Wei},
  booktitle={Proceedings of the IEEE/CVF Conference on Computer Vision and Pattern Recognition},
  pages={2837--2845},
  year={2021}
}

@inproceedings{yu2023video,
  title={Video probabilistic diffusion models in projected latent space},
  author={Yu, Sihyun and Sohn, Kihyuk and Kim, Subin and Shin, Jinwoo},
  booktitle={Proceedings of the IEEE/CVF Conference on Computer Vision and Pattern Recognition},
  pages={18456--18466},
  year={2023}
}

@inproceedings{anciukevivcius2023renderdiffusion,
  title={Renderdiffusion: Image diffusion for 3d reconstruction, inpainting and generation},
  author={Anciukevi{\v{c}}ius, Titas and Xu, Zexiang and Fisher, Matthew and Henderson, Paul and Bilen, Hakan and Mitra, Niloy J and Guerrero, Paul},
  booktitle={Proceedings of the IEEE/CVF Conference on Computer Vision and Pattern Recognition},
  pages={12608--12618},
  year={2023}
}

@inproceedings{ghosh2016deep,
  title={Deep feature extraction in the DCT domain},
  author={Ghosh, Arthita and Chellappa, Rama},
  booktitle={International Conference on Pattern Recognition},
  pages={3536--3541},
  year={2016}
}

@inproceedings{xie2021learning,
  title={Learning frequency-aware dynamic network for efficient super-resolution},
  author={Xie, Wenbin and Song, Dehua and Xu, Chang and Xu, Chunjing and Zhang, Hui and Wang, Yunhe},
  booktitle={Proceedings of the IEEE/CVF International Conference on Computer Vision},
  pages={4308--4317},
  year={2021}
}

@inproceedings{cai2021frequency,
  title={Frequency domain image translation: More photo-realistic, better identity-preserving},
  author={Cai, Mu and Zhang, Hong and Huang, Huijuan and Geng, Qichuan and Li, Yixuan and Huang, Gao},
  booktitle={Proceedings of the IEEE/CVF International Conference on Computer Vision},
  pages={13930--13940},
  year={2021}
}

@article{si2023freeu,
  title={Freeu: Free lunch in diffusion u-net},
  author={Si, Chenyang and Huang, Ziqi and Jiang, Yuming and Liu, Ziwei},
  journal={arXiv preprint arXiv:2309.11497},
  year={2023}
}

@article{choi2021ilvr,
  title={Ilvr: Conditioning method for denoising diffusion probabilistic models},
  author={Choi, Jooyoung and Kim, Sungwon and Jeong, Yonghyun and Gwon, Youngjune and Yoon, Sungroh},
  journal={arXiv preprint arXiv:2108.02938},
  year={2021}
}

\end{document}